%% file: bonato_report.tex
\documentclass{article}

\usepackage{RR}
\usepackage{hyperref}
\usepackage{latexsym}
\usepackage{amssymb}
\usepackage{t1enc}
\usepackage{graphicx}
\usepackage{amssymb}
\usepackage{amsmath}
\usepackage{cite}

\RRdate{Juin 2006}
\RRauthor{Roberto Bonato}
\authorhead{Bonato} 
\RRtitle{Une \'etude sur l'apprenabilit\'e \\ des Grammaires de Lambek Rigides}
\RRetitle{A Study on Learnability for Rigid Lambek Grammars} 

\titlehead{Learnability for Rigid Lambek Grammars} 

\RRresume{On pr\'esente les notions basiques du paradigme d'{\it apprenabilit\'e \`a la limite}
pour une classe de grammaires formelles defini par Gold en 1967, comme possible formalisation
du processus cognitif qui permet l'apprentissage d'une langue naturelle \`a partir d'exemples
d'\'enonc\'es bien form\'es. Ensuite, nous presentons les grammaires de Lambek, un formalisme issu
des grammaires cat\'egorielles que, bien que encore insuffisant \`a rendre compte de
nombre de phenomenes linguistiques, a des qualit\'es int\'er\'essantes par rapport \`a l'interface
syntaxe-s\'emantique. Enfin, nous pr\'esentons un r\'esultat d'apprenabilit\'e pour les grammaires
de Lambek Rigides dans le mod\`ele d'apprentissage de Gold \`a partir d'exemples structur\'es.}

\RRabstract{We present basic notions of Gold's \emph{learnability in
the limit} paradigm, first presented in 1967, a formalization of the
cognitive process by which a native speaker gets to grasp the
underlying grammar of his/her own native language by being exposed
to well formed sentences generated by that grammar. Then we present
Lambek grammars, a formalism issued from categorial grammars which,
although not as expressive as needed for a full formalization of
natural languages, is particularly suited to easily implement a
natural interface between syntax and semantics. In hte last part of
this work, we present a learnability result for Rigid Lambek
grammars from structured examples.}

\RRmotcle{Th\'eorie formelle de l'apprentissage, apprentissage automatique,
calcul de Lambek, linguistique computationnelle, grammaires formelles}

\RRkeyword{Formal Learning Theory, machine learning, Lambek calculus,
computational linguistics, formal grammars}

\RRprojet{SIGNES}

\RRtheme{\THSym}

\URFuturs

\begin{document}
    \makeRR

    \newtheorem{definition}{Definition}[section]
    \newtheorem{lemma}[definition]{Lemma}
    \newtheorem{example}[definition]{Example}
    \newtheorem{theorem}[definition]{Theorem}
    \newtheorem{condition}[definition]{Condition}
    \newtheorem{proposition}[definition]{Proposition}
    \newtheorem{corollary}[definition]{Corollary}

    \newcommand{\samples}{{\cal S}}
    \newcommand{\naturals}{\mathbb N}
    \newcommand{\naming}{{\rm L}}
    \newcommand{\seq}[1]{\langle #1 \rangle}
    \newcommand{\set}[1]{\{ #1 \}}
    \newcommand{\pl}{{\rm PL}}

    \include{nproof}
    \include{introduction_report}

    \include{learning_report}
    \include{lambek_report}
    \include{rigidlambek_report}
        \include{learnlambek_report}

    \include{conclusion_report}

    \bibliographystyle{alpha}
    \bibliography{biblio_report}
    \tableofcontents

\end{document}

%% file: nproof.tex
\message{<Paul Taylor's Proof Trees, 17 August 1990>}

\def\introrule{{\cal I}}\def\elimrule{{\cal E}}
\def\andintro{\using{\land}\introrule\justifies}
\def\impelim{\using{\Rightarrow}\elimrule\justifies}
\def\allintro{\using{\forall}\introrule\justifies}
\def\allelim{\using{\forall}\elimrule\justifies}
\def\falseelim{\using{\bot}\elimrule\justifies}
\def\existsintro{\using{\exists}\introrule\justifies}

\def\andelim#1{\using{\land}#1\elimrule\justifies}
\def\orintro#1{\using{\lor}#1\introrule\justifies}

\def\impintro#1{\using{\Rightarrow}\introrule_{#1}\justifies}
\def\orelim#1{\using{\lor}\elimrule_{#1}\justifies}
\def\existselim#1{\using{\exists}\elimrule_{#1}\justifies}


\newdimen\proofrulebreadth \proofrulebreadth=.05em
\newdimen\proofdotseparation \proofdotseparation=1.25ex
\newdimen\proofrulebaseline \proofrulebaseline=2ex
\newcount\proofdotnumber \proofdotnumber=3
\let\then\relax
\def\hfi{\hskip0pt plus.0001fil}
\mathchardef\squigto="3A3B
%
\newif\ifinsideprooftree\insideprooftreefalse
\newif\ifonleftofproofrule\onleftofproofrulefalse
\newif\ifproofdots\proofdotsfalse
\newif\ifdoubleproof\doubleprooffalse
\let\wereinproofbit\relax
%
\newdimen\shortenproofleft
\newdimen\shortenproofright
\newdimen\proofbelowshift
\newbox\proofabove
\newbox\proofbelow
\newbox\proofrulename
%
\def\shiftproofbelow{\let\next\relax\afterassignment\setshiftproofbelow\dimen0 }
\def\shiftproofbelowneg{\def\next{\multiply\dimen0 by-1 }%
\afterassignment\setshiftproofbelow\dimen0 }
\def\setshiftproofbelow{\next\proofbelowshift=\dimen0 }
\def\setproofrulebreadth{\proofrulebreadth}

\def\prooftree{
%
\ifnum	\lastpenalty=1
\then	\unpenalty
\else	\onleftofproofrulefalse
\fi
%
\ifonleftofproofrule
\else	\ifinsideprooftree
	\then	\hskip.5em plus1fil
	\fi
\fi
%
\bgroup
\setbox\proofbelow=\hbox{}\setbox\proofrulename=\hbox{}%
\let\justifies\proofover\let\leadsto\proofoverdots\let\Justifies\proofoverdbl
\let\using\proofusing\let\[\prooftree
\ifinsideprooftree\let\]\endprooftree\fi
\proofdotsfalse\doubleprooffalse
\let\thickness\setproofrulebreadth
\let\shiftright\shiftproofbelow \let\shift\shiftproofbelow
\let\shiftleft\shiftproofbelowneg
\let\ifwasinsideprooftree\ifinsideprooftree
\insideprooftreetrue
%
\setbox\proofabove=\hbox\bgroup$\displaystyle 
\let\wereinproofbit\prooftree
%
\shortenproofleft=0pt \shortenproofright=0pt \proofbelowshift=0pt
%
\onleftofproofruletrue\penalty1
}

\def\eproofbit{
%
\ifx	\wereinproofbit\prooftree
\then	\ifcase	\lastpenalty
	\then	\shortenproofright=0pt	
	\or	\unpenalty\hfil		
	\or	\unpenalty\unskip	
	\else	\shortenproofright=0pt	
	\fi
\fi
%
\global\dimen0=\shortenproofleft
\global\dimen1=\shortenproofright
\global\dimen2=\proofrulebreadth
\global\dimen3=\proofbelowshift
\global\dimen4=\proofdotseparation
\global\count10=\proofdotnumber
%
$\egroup  
%
\shortenproofleft=\dimen0
\shortenproofright=\dimen1
\proofrulebreadth=\dimen2
\proofbelowshift=\dimen3
\proofdotseparation=\dimen4
\proofdotnumber=\count10
}

\def\proofover{
\eproofbit 
\setbox\proofbelow=\hbox\bgroup 
\let\wereinproofbit\proofover
$\displaystyle
}%
%
\def\proofoverdbl{
\eproofbit 
\doubleprooftrue
\setbox\proofbelow=\hbox\bgroup 
\let\wereinproofbit\proofoverdbl
$\displaystyle
}%
%
\def\proofoverdots{
\eproofbit 
\proofdotstrue
\setbox\proofbelow=\hbox\bgroup 
\let\wereinproofbit\proofoverdots
$\displaystyle
}%
%
\def\proofusing{
\eproofbit 
\setbox\proofrulename=\hbox\bgroup 
\let\wereinproofbit\proofusing
\kern0.3em$
}

\def\endprooftree{
\eproofbit 
  \dimen5 =0pt
%
\dimen0=\wd\proofabove \advance\dimen0-\shortenproofleft
\advance\dimen0-\shortenproofright
%
\dimen1=.5\dimen0 \advance\dimen1-.5\wd\proofbelow
\dimen4=\dimen1
\advance\dimen1\proofbelowshift \advance\dimen4-\proofbelowshift
%
\ifdim	\dimen1<0pt
\then	\advance\shortenproofleft\dimen1
	\advance\dimen0-\dimen1
	\dimen1=0pt
	\ifdim  \shortenproofleft<0pt
        \then   \setbox\proofabove=\hbox{%
			\kern-\shortenproofleft\unhbox\proofabove}%
                \shortenproofleft=0pt
        \fi
\fi
%
\ifdim	\dimen4<0pt
\then	\advance\shortenproofright\dimen4
	\advance\dimen0-\dimen4
	\dimen4=0pt
\fi
%
\ifdim	\shortenproofright<\wd\proofrulename
\then	\shortenproofright=\wd\proofrulename
\fi
%
\dimen2=\shortenproofleft \advance\dimen2 by\dimen1
\dimen3=\shortenproofright\advance\dimen3 by\dimen4
%
\ifproofdots
\then
	\dimen6=\shortenproofleft \advance\dimen6 .5\dimen0
	\setbox1=\vbox to\proofdotseparation{\vss\hbox{$\cdot$}\vss}%
	\setbox0=\hbox{%
		\advance\dimen6-.5\wd1
		\kern\dimen6
		$\vcenter to\proofdotnumber\proofdotseparation
			{\leaders\box1\vfill}$%
		\unhbox\proofrulename}%
\else	\dimen6=\fontdimen22\the\textfont2 
	\dimen7=\dimen6
	\advance\dimen6by.5\proofrulebreadth
	\advance\dimen7by-.5\proofrulebreadth
	\setbox0=\hbox{%
		\kern\shortenproofleft
		\ifdoubleproof
		\then	\hbox to\dimen0{%
			$\mathsurround0pt\mathord=\mkern-6mu%
			\cleaders\hbox{$\mkern-2mu=\mkern-2mu$}\hfill
			\mkern-6mu\mathord=$}%
		\else	\vrule height\dimen6 depth-\dimen7 width\dimen0
		\fi
		\unhbox\proofrulename}%
	\ht0=\dimen6 \dp0=-\dimen7
\fi
%
\let\doll\relax
\ifwasinsideprooftree
\then	\let\VBOX\vbox
\else	\ifmmode\else$\let\doll=$\fi
	\let\VBOX\vcenter
\fi
\VBOX	{\baselineskip\proofrulebaseline \lineskip.2ex
	\expandafter\lineskiplimit\ifproofdots0ex\else-0.6ex\fi
	\hbox	spread\dimen5	{\hfi\unhbox\proofabove\hfi}%
	\hbox{\box0}%
	\hbox	{\kern\dimen2 \box\proofbelow}}\doll%
%
\global\dimen2=\dimen2
\global\dimen3=\dimen3
\egroup 
\ifonleftofproofrule
\then	\shortenproofleft=\dimen2
\fi
\shortenproofright=\dimen3
%
\onleftofproofrulefalse
\ifinsideprooftree
\then	\hskip.5em plus 1fil \penalty2
\fi
}


%% file: introduction_report.tex
\section{Introduction}

\begin{quote}
    {\it How comes it that human beings, whose contacts with the world
    are brief and personal and limited, are nevertheless able to
    know as much as they do know?}\newline \newline
    Sir Bertrand Russell (citato da Noam Chomsky in \cite{chomsky75}).
\end{quote}

Formal Learning Theory was first defined in an article by E. M. Gold
in 1967 (see \cite{gold67}) as a first effort to provide 
a rigurous formalization of grammatical inference,
that is the process by which a learner,
presented with a certain given subset of well-formed sentences
of a given language, gets to infer the grammar that generates it. 
The typical example of such a process is given by a child
whi gets to master, in a completely spontaneous way and on the
basis of the relatively small amount of information provided
by sentences uttered in its cultural environment, the higly complex
and subtle rules of her mother tongue, to the point that she can
utter correct and {\it original} sentences before her third year
of life. In \cite{owjm97} such a formal framework is used in
the broder context of the mathematical formalization of any 
kind of inductive reasoning. In this case the learner is
``the scientist'' who, on the basis of finite amount of
empirical evidences provided by natural phenomena,
formulates scientific hypotheses would could intensionally
accunt for them.

After an initial skepticism about the grammars that could be
actually learnt in Gold's paradigm (a skepticism shared and
in a way enouraged by Gold himself, who proves the non-learnability
in its model of the four classes of grammars of Chomsky's 
hierarchy), recently there has been a renewal of interest
toward this computational model of learning. One of the most
recent results is Shinohara's (see \cite{shinohara90a}), who
proves that as soon as we bound the number of rules
in a context-sensitive grammar, it becomes learnable in Gold's
paradigm.\newline

Lambek Grammars have recently known a renewed interest
as a mathematical tool for the description of certain linguistics
phenomena, after having being long neglected after their
first definition in \cite{lambek58}. Van Benthem was among the
first who stressed the singular correspondence between Montague
Semantics (see \cite{montague73}) and the notion of structure
associated to a sentence of a Lambek grammar. In particular,
a recent work by Hans-Jorg Tiede (see \cite{tiede99}) has made
clearer the notion of structure of a sentence in a Lambek grammar,
in contrast with a previsous definition given by  
Buszkowski (see \cite{buszkowski86}). In doing so, he gets to prove
a meaningful result about Lambek Grammars, that is that the class 
of tree languages generated by Lambek grammars strictly contains
the class of tree languages generated by context-free grammars.
\newline

Section 2 introduces the basic notions of Learning Theory by Gold
and provides a short review of most important known fact and
results about it. 
Section 3 is a short introduction fo Lambek Grammars: we give
their definition and we present the features which make them
attractive from a computational linguistics point of view.
Section 4 briefly presents the class of {\it rigid} Lambek 
Grammars, which is the object of our lerning algorithm, along
with some basic properties and open questions.
In Section 5 we present a learning algorithm for rigid 
Lambek grammars from a {\it structured input}: the algorithm
takes as its input a finite set of what has been defined in 
chapter 3 as {\it proof tree structures}. It is proved
convergence for the algorithm and so the lernability 
for the class of rigid Lambek grammars.

%% file: learning_report.tex
\section{Grammatical Inference}

    \subsection{Child's First Language Acquisition}
    One of the most challenging goals for modern cognitive
    sciences is providing a sound theory accounting for the process by which
    any human being gets to master the highly complex and articulated
    grammatical structure of her mother tongue in a relatively small 
    amount of time. Between the age of 3 and 5 we witness in children a {\it linguistic 
    explosion}, at the end of which we can say that the child masters all the
    grammatical rules of her mother tongue, and subsequent learning is not but
    lexicon acquisition.
    Moreover, cognitive psychologists agree (see \cite{osherson95}) in stating
    that the learning process is almost completely based on {\it positive} evidence
    provided by the cultural environment wherein the child is grown up: 
    that is, correct statements belonging to her mother tongue. 
    {\it Negative} evidence (any information or feedback given to the 
    child to identify not-well-formed sentences), is almost completely 
    absent and, in any case, doesn't seem to play any significant role
    in the process of learning (see \cite{pinker94}). 
    Simply stated, the child acquires a language due to
    the exposition to correct sentences coming from her linguistic environment 
    and not to the negative feedback she gets when she utters a wrong sentence.

    Providing a formal framework wherein to inscribe such an astounding ability
    to extract highly articulated knowledge (i.e. the grammar of a human language) 
    from a relatively small amount of ``raw'' data (i.e. the statements of 
    the language the child is exposed to during her early childhood) was one of the
    major forces that led to the the definition of a formal learning theory as
    the one we are going to describe in the following sections.

    \subsection{Gold's Model}
    \label{GoldModelSec}
    The process of a child's first language acquisition
    can be seen as an instance of the more general
    problem of {\it grammatical inference}.
    In particular we will restrict our attention to the process of
    inference {\it from positive data only}. Simply stated, it's the
    process by which a learner can acquire the
    whole grammatical structure of a formal language
    on the basis of well-formed sentences belonging to the target language.

    In 1967 Gold defined (see \cite{gold67}) the formal model for 
    the process of grammatical inference from positive data that will be 
    adopted in the present work. In Gold's model, grammatical inference 
    is conceived as an {\it infinite} process during which a {\it learner} 
    is presented with an infinite stream of sentences $s_0, s_1, \ldots, 
    s_n \ldots$, belonging to language which has to be learnt, one sentence at a time.
    Each time the learner is presented with a new sentence $s_i$, she formulates 
    a new hypothesis $G_i$ on the nature of the underlying grammar that could 
    generate the language the sentences she has seen so far belong to: 
    since she is exposed to an infinite number of sentences, she will
    conjecture an infinite number of (not necessarily different) grammars
    $G_0, G_1, \ldots, G_n \ldots$.

    \begin{displaymath}
        \underbrace{\underbrace{\underbrace
		{\underbrace{s_0}_{G_0}, s_1}_{G_1}, \ldots, s_n}_
            	{{G_n}_{\ddots}}, \ldots}_{G}
    \end{displaymath}

    Two basic assumptions are made about the stream of sentences 
    she is presented with: (i) only grammatical sentences (i.e. belonging to 
    the target language) appear in the stream, coherently with
    our commitment to the process of grammar induction {\it from positive data only};
    (ii) every possible sentence of the language must appear in the stream 
    (which must be therefore an {\it enumeration} of the elements of the language).

    The learning process is considered {\it successful}
    when, from a given point onward, the grammar conjectured by the learner 
    doesn't change anymore and it coincides with the grammar that 
    actually generates the target language. 
    It is important to stress the fact that one can never know at any 
    finite stage whether the learning has been successful or not due 
    to the infinite nature of the learning process itself:
    at each finite stage, no one can predict whether next sentence
    will change or not the current hypothesis. The goal of the theory lies in
    devising a successful {\it strategy} for making guesses, that is, one 
    which can be proved to {\it converge} to the correct grammar after a finite
    (but unknown) amount of time (or positive evidence, which is the same in
    our model).
    Gold called this criterion of successful learning {\it identification in the
    limit}\index{identification in the limit}.

    According to this criterion, a class of grammars is
    said to be {\it learnable} when, {\it for any
    language} generated by a grammar belonging to the class, 
    and {\it for any enumeration} of its sentences, there
    is a learner that successfully identifies the correct 
    grammar that generates the language.
    A good deal of current research on formal learning theory is devoted to
    identifying non-trivial classes of languages which are
    learnable in Gold's model or useful criterions to deduce
    (un)learnability for a class of languages on the basis
    of some structural property of the language.

    As it will be made clear in the following sections, accepting this
    criterion for successful learning means that we are
    not interested in {\it when} the learning has taken place: in
    fact there's no effective way to decide if it has or not at any finite stage.
    Our aim is to devise effective procedures such that, {\it if applied to
    the infinite input stream of sentences}, are guaranteed to
    converge to the grammar we are looking for, if it exists.

\section{Basic Notions}
    We present here a short review of (Formal) Learning Theory
    as described in \cite{kanazawa98},
    whence we take the principal definitions and notation
    conventions.

    \subsection{Grammar Systems}
    The fist step in the formalization of the learning process
    is the formal definition of both the ``cultural environment''
    wherein this process takes place and the ``positive evidences'' the
    learner is exposed to. To do this, we introduce the notion of
    {\it grammar system}\index{grammar system}.

    \begin{definition}[Grammar System]
        A grammar system is a triple $\seq{\Omega, \samples, \naming}$, where
        \begin{itemize}
            \item $\Omega$\index{$\Omega$} is a certain recursive 
                set of finitary objects on which mechanical
                computations can be carried out;
            \item $\samples$ is a certain recursive subset of $\Sigma^*$, 
                where $\Sigma$ is a given finite alphabet;
            \item $\naming$ is a function that maps elements of $\Omega$ 
                to subsets of $\samples$,
                i.e. $\naming:\Omega \rightarrow~{\wp}(\samples)$.
        \end{itemize}
    \end{definition}

    We can think of $\Omega$ as the ``hypothesis space'',
    whence the learner takes her grammatical conjectures,
    according to the positive examples she has been exposed to
    up to a certain finite stage of the learning process.
    Elements of $\Omega$ are called {\it grammars}\index{grammars}.

    Positive examples presented to the learner belong to the set
    $\samples$ (often we simply have $\samples = \Sigma^*$); its
    elements are called {\it sentences}\index{sentence}, while its subsets are
    called {\it languages}\index{language}. As it will be made
    clear in the following sections, the nature of elements in
    $\samples$ strongly influences the process of learning:
    intuitively, we can guess that the more information they bear,
    the easier the learning process is, if it is possible at all.

    The function $\naming$ maps each grammar $G$ belonging to
    $\Omega$ into a subset of $\samples$ which is designated as the
    {\it language generated by G}. That's why we often refer to
    $\naming$ as the {\it naming function}\index{naming function}.
    The question of whether $s \in {\naming}(G)$ holds
    between any $s \in \samples$ and $G \in \Omega$ is addressed to
    as the {\it universal membership problem}\index{universal
    membership problem}.

    \begin{figure}[htbp]
            \begin{center}
                \includegraphics[height=7cm]{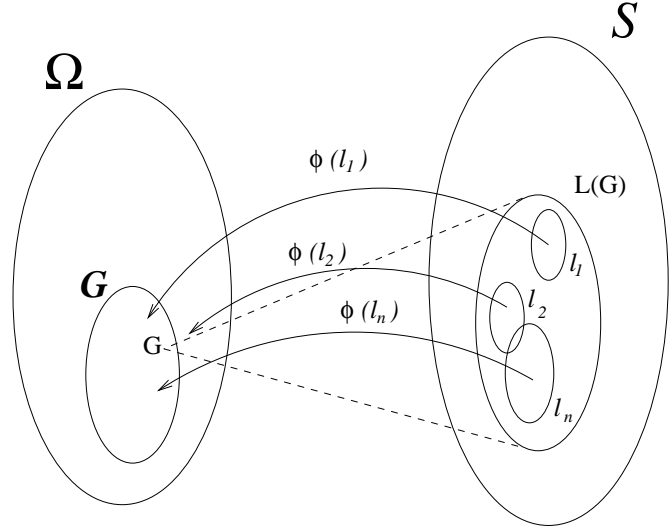}
            \end{center}
            \caption{Grammatical Inference}
        \end{figure}

    \begin{example}
        Let $\Sigma$ be any finite alphabet and let $DFA$ 
        be the set of {\it deterministic finite automata}
        whose input alphabet is $\Sigma$.
        For every $M \in DFA$, let $\naming(M)$ be the set of 
        strings over $\Sigma$ accepted by $M$. Then
        $\seq{DFA, \Sigma^*, \naming}$ is a grammar system.
    \end{example}

    \begin{example}
        Let $\Sigma$ be any finite alphabet and let $RegExpr$ 
        be the set of {\it regular expressions} over $\Sigma$.
        For every $r \in RegExpr$, let $\naming(r)$ be the regular 
        language represented by $r$. Then
        $\seq{RegExpr, \Sigma^*, \naming}$ is a grammar system.
    \end{example}

    \begin{example}[Angluin, 1980]
        \label{angluinex}
        Let $\Sigma$ any finite alphabet, and let $Var$
        be a countably infinite set of variables,
        disjoint from $\Sigma$. A {\it pattern} over $\Sigma$ 
        is any element of $(\Sigma \cup Var)^+$: let $Pat$ be 
        the set of patterns over $\Sigma$. For
        every $p \in Pat$, let $\naming(p)$ be the set of 
        strings that can be obtained from $p$
        by uniformly replacing each variable $x$ occurring in $p$ by some string
        $w \in \Sigma^+$. The triple $\seq{Pat, \Sigma^+, \naming}$ is a grammar
        system.
    \end{example}

    \subsection{Learning Functions, Convergence, Learnability}
    Once formally defined both the set of possible ``guesses''
    the learner can make and the set of the positive examples
    she is exposed to, we need a formal notion for the mechanism
    by which the learner formulates hypotheses, on the basis finite 
    sets of well-formed sentences of a given language, about the 
    grammar that generates them.

    \begin{definition}[Learning Function]
        Let $\seq{\Omega, \samples, \naming}$ be a grammar system.
        A {\it learning function} is a partial function that maps finite sets
        of sentences to grammars,
        \begin{displaymath}
            \varphi: \bigcup_{k \geq 1} \samples^k \rightharpoonup \Omega
        \end{displaymath}
        where $\samples^k$ denotes the set of k-ary sequences of sentences.
    \end{definition}
    A learning function can be seen as a formal
    model of the cognitive process by which a learner conjectures
    that a given finite set of sentences belongs to the language generated
    by a certain grammar. Since it's partial, possibly the learner
    cannot infer any grammar from the stream of sentences she has seen so far.\newline

    According to the informal model outlined in section
    \ref{GoldModelSec}, in a successful learning process, we
    require the guesses made by the learner to remain the same
    from a certain point onward in the infinite process of
    learning. That is to say, there must be a finite stage (even if we
    don't know which one) after which the grammar inferred on the
    basis of all the positive examples the learner has seen so far
    is always the same. This informal idea can be made precise by
    introducing the notion of {\it convergence}\index{convergence}
    for a learning function:
    \begin{definition}[Convergence]
        Let $\seq{\Omega, \samples, \naming}$ be a grammar system, $\varphi$ a learning function,
        \begin{displaymath}
            \seq{s_i}_{i \in \naturals} = \seq{s_0, s_1, \ldots}
        \end{displaymath}
        an infinite sequence of sentences belonging to $\samples$, and let
        \begin{displaymath}
            G_i = \varphi (\seq{s_0, \ldots, s_i})
        \end{displaymath}
        for any $i\in \naturals$ such that $\varphi$ is defined on the finite sequence
        $\seq{s_0, \ldots, s_i}$. $\varphi$ is said {\rm to converge to $G$ on
        $\seq{s_i}_{i \in \naturals}$} if there exists $n \in \naturals$ such that,
        for each $i \geq n$, $G_i$ is defined and $G_i = G$
        (equivalently, if $G_i = G$ for all but finitely many
        $i\in \naturals$).
    \end{definition}

    As we've already pointed out, one can never say exactly
    {\it if} and {\it when} convergence of a learning
    function to a certain grammar has taken place: this is
    due to the {\it infinite} nature of the process by
    which a learner gets to learn a given language
    in Gold's model. At any
    finite stage of the learning process there's no way to
    know whether the next sentence
    the learner will see causes the current hypothesis to 
    change or not.\newline
    
    We will say that a class of grammars is {\it
    learnable} when for each language generated by its grammars
    there exists a learning function which converges to the 
    correct underlying grammar on the basis of
    any enumeration of the sentences of the language. Formally:
    \begin{definition}[Learning ${\cal G}$]
        Let $\seq{\Omega, \samples, \naming}$ be a grammar system,
        and ${\cal G} \subseteq \Omega$ a given set of grammars.
        The learning function $\varphi$ is said
        {\rm to learn} ${\cal G}$ if the following condition holds:
        \begin{itemize}
            \item for every language 
                $L \in \naming({\cal G})= \set{\naming(G) \ | \ G \in {\cal G}}$,
            \item and for every infinite sequence 
                $\seq{s_i}_{i \in \naturals}$ that enumerates $L$
                (i.e., $\{s_i\  |\  i \in \naturals \} = L$)
        \end{itemize}
        there exists a $G \in \cal G$ such that $\naming(G)=L$,
        such that $\varphi$ converges to $G$ on $\seq{s_i}_{i \in \naturals}$.
    \end{definition}
    So we will say that a given learning function {\it converges}
    to a single grammar, but that it {\it learns}
    a class of grammars. The learning for a single grammar,
    indeed, could be trivially implemented by
    a learning function that, for any given sequence
    of sentences as input, always returns that grammar.

    \begin{definition}[Learnability of a Class of Grammars\index{learnability}]
        A class ${\cal G}$ of grammars is called {\rm learnable}
        if and only if there exists a learning function that
        learns ${\cal G}$. It is called {\rm effectively learnable} if and
        only if there is a {\rm computable} learning function that learns ${\cal G}$.
    \end{definition}
    Obviously effective learnability implies learnability.

    \begin{example}
        Let $\seq{\Omega, \samples, \naming}$ be any grammar system and let
        ${\cal G}=\set{G_0, G_1, G_2} \subseteq \Omega$ and suppose there are elements
        $w_1, w_2 \in \samples$ such that $w_1 \in \naming(G_1)-\naming(G_0)$ and
        $w_2 \in \naming(G_2)-(\naming(G_1) \cup \naming(G_0))$.
        Then it's easy to verify that the following learning function learns ${\cal G}$:
        \begin{displaymath}
            \varphi(\seq{s_0, \ldots, s_i}) =
            \left \{
            \begin{array}{rl}
                G_2 & \mbox{ if } w_2 \in \set{s_0, \ldots, s_i}, \\
                G_1 & \mbox{ if } w_1 \in \set{s_0, \ldots, s_i} \ {\rm and }\
                    w_2 \not \in \set{s_0, \ldots, s_i}, \\
                G_0 & \mbox{ otherwise.}
            \end{array}
            \right.
        \end{displaymath}
    \end{example}

    \begin{example}
        Let's consider the grammar system $\seq{CFG, \Sigma^*, \naming}$
        of context-free grammars over the alphabet $\Sigma$.
        Let ${\cal G}$ be the subclass of $CFG$ consisting of grammars
        whose rules are all of the form
        \begin{displaymath}
            S \rightarrow w,
        \end{displaymath}
        where $w \in \Sigma^*$. We can easily see that
        $\naming({\cal G})$ is exactly the class of
        finite languages over $\Sigma$. Let's define the learning function $\varphi$ as
        \begin{displaymath}
            \varphi(\seq{s_0, \ldots, s_i}) = \seq{\Sigma, \set{S}, S, P},
        \end{displaymath}
        where
        \begin{displaymath}
            P = \set{S \rightarrow s_0, \ldots, S \rightarrow s_i}.
        \end{displaymath}
        Then $\varphi$ learns ${\cal G}$.
    \end{example}

\subsection{Structural Conditions for (Un)Learnability}
    One of the first important results in learnability
    theory presented in \cite{gold67} was a sufficient
    condition to deduce the unlearnability
    of a class ${\cal G}$ of grammars on the basis of
    some formal properties of the class of
    languages ${\cal L}=\naming({\cal G})$ (see theorem
    \ref{goldth}). We present here some structural conditions sufficient to deduce
    (un)learnability for a class of grammars. Such results are
    useful to get a deeper understanding to the general problem of
    learnability for a class of grammars.

    \subsubsection{Existence of a Limit Point}
        Let's define the notion of {\it limit point}\index{limit
        point} for a class of languages:
        \begin{definition}[Limit Point]
            A class ${\cal L}$ of languages {\rm has a limit point}
            if there exists an infinite sequence
            $\seq{L_n}_{n \in \naturals}$ of languages in
            ${\cal L}$ such that
            \begin{displaymath}
                L_0 \subset L_1 \subset \cdots \subset L_n \subset \cdots
            \end{displaymath}
            and there exists another language $L \in \cal L$ such that
            \begin{displaymath}
                L = \bigcup_{n \in \naturals} L_n
            \end{displaymath}
            The language $L$ is called {\rm limit point} of $\cal L$.
        \end{definition}
        \begin{figure}[htbp]
            \begin{center}
                \includegraphics[height=4cm]{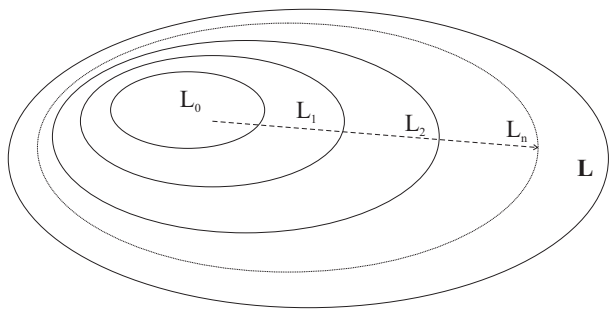}
            \end{center}
            \caption{A limit point for a class of languages.}
        \end{figure}\newpage

        \begin{lemma}[Blum and Blum's locking sequence lemma, 1975]\ \newline
            Suppose that a learning function $\varphi$ converges
            on every infinite sequence that enumerates a language $L$.
            Then there is a finite sequence $\seq{w_0, \ldots, w_l}$
            (called a {\rm locking sequence} for $\varphi$ and L)
            with the following properties:
            \renewcommand{\labelenumi}{{\rm (\roman{enumi})}}
            \begin{enumerate}
                \item $\set{w_0, \ldots, w_l} \subseteq L$,
                \item for every finite sequence $\seq{v_0, \ldots, v_m}$,
                    if $\set{v_0, \ldots, v_m} \subseteq
                    L$, then $\varphi(\seq{w_0, \ldots, w_l})=
                    \varphi(\seq{w_0, \ldots, w_l, v_0, \ldots, v_m})$.
            \end{enumerate}
        \end{lemma}
        Intuitively enough, the previous lemma (see \cite{bb75})
        states that if a learning function converges, then there 
        must exist a {\it finite} subsequence of input sentences 
        that ``locks'' the guess made by the learner on the grammar 
        the learning function converges to:
        that is to say, the learning function returns always the same
        grammar for any input stream of sentences containing that
        finite sequence.

        The locking sequence lemma proves one of the first unlearnability 
        criterions in Gold's learnability framework:
        \begin{theorem}
            If $\naming({\cal G})$ has a limit point, then ${\cal G}$ is not learnable.
        \end{theorem}

        An easy consequence of the previous theorem is the following
        \begin{theorem}[Gold, 1967]
            \label{goldth}
            For any grammar system, a class ${\cal G}$ of grammars
            is not learnable if $\naming({\cal G})$ contains all
            finite languages and at least one infinite language.
        \end{theorem}
        {\it Proof sketch.} Let $L_1 \subset L_2 \subset \ldots$
        be a sequence of finite languages and let
        $L_{\infty} = \bigcup_{i=1}^{\infty}L_i$.
        Suppose there were a learning function $\varphi$
        that learns the class
        $\{L\ |\ L \mbox{\ is finite}\} \cup \{L_{\infty}\}$.
        Then $\varphi$ must identify
        any finite language in a finite amount of time.
        But then we can build an infinite sequence
        of sentences that forces $\varphi$ to make an infinite
        number of mistakes: we first present
        $\varphi$ with enough examples from $L_1$ to make it guess $L_1$;
        then with enough examples from $L_2$ to make it guess $L_2$, and so on. Note that
        all our examples belong to $L_{\infty}$.

        \subsubsection{(In)Finite Elasticity}
        As we've seen in the previous section, the existence
        of a limit point for a class of languages implies the existence of an
        ``infinite ascending chain'' of languages like the one
        described by the following, weaker condition:
        \begin{definition}[Infinite Elasticity\index{infinite elasticity}]
            \label{elinf}
            A class ${\cal L}$ of languages is said {\rm to have
            infinite elasticity} if there exists an
            infinite sequence $\seq{s_n}_{n \in \naturals}$ of sentences and
            an infinite sequence $\seq{L_n}_{n \in \naturals}$ of languages
            such that for every $n \in \naturals$,
            \begin{displaymath}
                s_n \notin L_n,
            \end{displaymath}
            and
            \begin{displaymath}
                \set{s_0,\ldots,s_n} \subseteq L_{n+1}.
            \end{displaymath}
        \end{definition}
        The following definition, although trivial, identifies
        an extremely useful criterion to deduce learnability
        for a class of grammars:
        \begin{definition}[Finite Elasticity\index{infinite elasticity}]
            A class ${\cal L}$ of languages is said to have
            {\rm finite elasticity} if it doesn't have infinite elasticity.
        \end{definition}

        Dana Angluin proposed in \cite{angluin80} a characterization
        of the notion of learnability in a ``restrictive setting''
        which is of paramount importance in formal learning theory.
        Such restrictions are about the membership problem and
        the recursive enumerability for the class of grammars
        whose learnability is at issue.
        Let $\seq{\Omega, \samples, \naming}$ be a grammar system
        and ${\cal G} \subseteq \Omega$ a class of grammars, let's define:
        \begin{condition}
            \label{membcond}
            There is an algorithm that, given $s \in \samples$ and $G \in {\cal G}$,
            determines whether $s \in \naming(G)$.
        \end{condition}
        \begin{condition}
            \label{recond}
            ${\cal G}$ is a recursively enumerable class of grammars.
        \end{condition}

        Condition \ref{membcond} is usually referred to as {\it decidability
        for the universal membership problem}, and condition \ref{recond} as
        the {\it recursive enumerability} condition. 
        Such restrictions are not unusual in concrete situations
        where learnability is at issue,
        so they don't significantly affect the usefulness of the following
        characterization of the notion learnability under such restrictive 
        conditions.

        \begin{theorem}[Angluin 1980]
            \label{angluinth}
            Let $\seq{\Omega, \samples, \naming}$ be a grammar
            system for which both conditions \ref{membcond} and
            \ref{recond} hold, and let ${\cal G}$ be a recursively enumerable
            subset of $\Omega$.
            Then ${\cal G}$ is learnable if and only if there
            exists a computable partial function
            $\psi: \Omega \times \naturals \rightharpoonup \samples$ such that:
            \renewcommand{\labelenumi}{{\rm (\roman{enumi})}}
            \begin{enumerate}
                \item for all $n \in {\naturals}$, $\psi(G,n)$ is defined if and only if
                    $G \in {\cal G}$ and $\naming(G) \not=
                    \emptyset$;
                \item for all $G \in {\cal G}, T_G = \set{\psi(G,n)\ |\ n \in \naturals}$
                    is a finite subset of $\naming(G)$;
                \item for all $G, G' \in {\cal G}$, if $T_G \subseteq \naming(G')$, then
                    $\naming(G') \not\subset \naming(G)$.
            \end{enumerate}
        \end{theorem}
        Note: From this point onward, unless otherwise stated,
        we will restrict our attention to classes of grammars that fulfill both
        condition \ref{membcond} and condition \ref{recond}.\newline

        Angluin's theorem introduces the notion of $T_G$ as the
        {\it tell-tale set} for a given language.
        Learnability in the restricted environment is
        characterized by the existence of a mechanism
        (the function $\psi$) to enumerate all the sentences belonging
        to such a {\it finite} subset of the target language.
        Even more, a tell-tale set for a given
        grammar $G$ is such that if it is included in the
        language generated by another grammar $G^\prime$, then
        \begin{itemize}
            \item either $\naming(G)$ is included in $\naming(G^\prime)$,
            \item or $\naming(G^\prime)$ contains other sentences
            as well as those belonging to $\naming(G)$.
        \end{itemize}
        Otherwise stated, it is never the case that
        $T_G \subseteq \naming(G^\prime) \subseteq \naming(G)$.
        The point of the tell-tale subset is that once the
        strings of that subset have appeared among
        the sample strings, we need not fear overgeneralization
        in guessing a grammar $G$.
        This is because the true answer, even if it is not
        $\naming(G)$, cannot be a proper subset of $\naming(G)$.
        This means that a learner who has seen only the sentences
        belonging to the tell-tale set
        for a given grammar $G$, is justified in conjecturing $G$
        as the underlying grammar, since doing so
        never results in overshooting or inconsistency.
        \begin{figure}[htbp]
            \begin{center}
                \includegraphics[height=5cm]{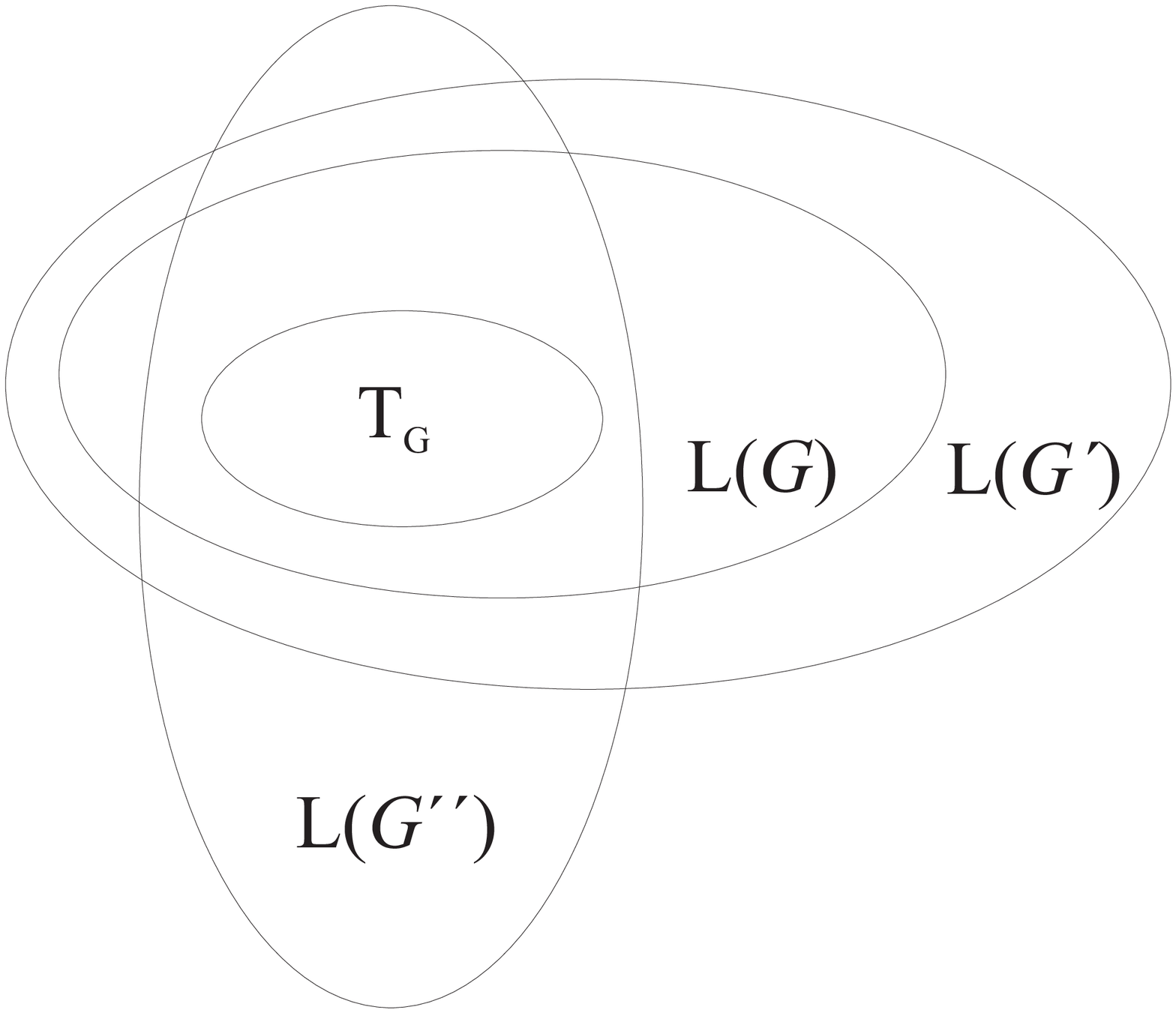}
            \end{center}
            \caption{A tell-tale set for $\naming(G)$.}
        \end{figure}\newline

        As a consequence of Angluin's theorem, Wright proved
        in \cite{wright89} the following
        \begin{theorem}[Wright, 1989]
            \label{wrightth}
            Let $\seq{\Omega, \samples, \naming}$ and ${\cal G}$
            be as in theorem \ref{angluinth}.
            If $\naming(\cal G)$ has finite elasticity, then ${\cal G}$ is learnable.
        \end{theorem}
        In such a restricted framework, therefore, the task
        of proving learnability for a certain class of grammars
        can be reduced to the usually simpler task of
        proving its finite elasticity.

        Due to Wright's theorem we can establish the following useful implications
        \begin{eqnarray*}
            \naming({\cal G}) \mbox{\ has\ finite\ elasticity} 
		& \stackrel{\dagger}{\Rightarrow} &{\cal G} \mbox{\ is\ learnable} \\
            \naming({\cal G}) \mbox{\ has\ a\ limit\ point} 
		&\Rightarrow & {\cal G} \mbox{\ is\ {\it un}learnable}\\
            {\cal G} \mbox{\ is\ {\it un}learnable} &\stackrel{\dagger}{\Rightarrow}
		&  \naming({\cal G}) \mbox{\ has\ {\it in}finite\ elasticity}
        \end{eqnarray*}
        The implications indicated by $\stackrel{\dagger}{\Rightarrow}$
        depend on the decidability of universal membership and
        recursive enumerability of the class of grammars at issue,
        as defined in conditions \ref{membcond} and \ref{recond}.

        \subsubsection{Kanazawa's Theorem}
        The following theorem (see \cite{kanazawa98}),
        which is a generalization of a
        previous theorem by Wright, provides a sufficient
        condition for a class of grammars
        to have finite elasticity, and therefore to be learnable.
        A relation $R \subseteq \Sigma^* \times \Upsilon^*$ 
        is said to be {\it finite-valued} if and only if for every 
	$s \in \Sigma^*$, the set $\{u \in \Upsilon^*\ |\ sRu\}$ is finite.
        \begin{theorem}
            \label{finelth}
            Let ${\cal M}$ be a class of languages over $\Upsilon$ 
            that has finite elasticity, and
            let $R \subseteq \Sigma^* \times \Upsilon^*$ 
	    be a finite-valued relation. Then
            ${\cal L} = \set{R^{-1}[M]\ |\ M\in{\cal M}}$ also has finite elasticity.
        \end{theorem}

        This theorem is a powerful tool to prove finite
        elasticity (and therefore learnability) for classes of
        grammars. Once we prove the finite elasticity for a
        certain class of grammars in the ``straight'' way, we can
        get a proof for finite elasticity of other classes
        of grammars, due to the relatively loose requirements of
        the theorem. All we have to do is to devise a ``smart''
        finite-valued relation between the first class and a new
        class of grammars such that the anti-image of the latter
        under this relation is the class for which we want to prove
        finite elasticity.

        \begin{figure}[htbp]
            \begin{center}
                \includegraphics[height=4.5cm]{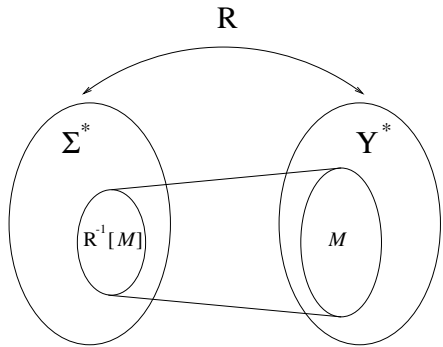}
            \end{center}
            \caption{Kanazawa's theorem.}
        \end{figure}
	\newpage

    \subsection{Constraints on Learning Functions}
    In the definition of learnability nothing is
    said about the behaviour of learning functions apart
    from convergence to a correct grammar.
    Further constraints can be imposed: one can choose
    a certain learning strategy. Intuitively, a strategy
    refers to a policy, or preference, for choosing hypotheses.
    Formally, a strategy can be
    analyzed as merely picking a subset of possible learning
    functions. Strategies can be grouped by numerous properties.
    We choose to group them by restrictiveness, defined as follows:
    \begin{definition}[Restrictiveness]
        If a strategy constrains the class of learnable languages it is said to be
        {\rm restrictive}.
    \end{definition}

    For example, strategies are grouped as computational
    constraints (computability, time complexity), constraints
    on potential conjectures (consistency), constraints
    on the relation between conjectures (conservatism), etc.
    Since the classes we will be discussing are all classes of
    recursive languages, ``restrictive'' will be taken to
    mean ``restrictive for classes of recursive languages''.

    \subsubsection{Non-restrictive Constraints}
    The proof of theorem \ref{angluinth} implies that
    in a grammar system where universal membership is decidable,
    a recursively enumerable class of grammars is learnable if
    and only if there is a computable learning function that
    learns it {\it order-independently}, {\it prudently}, and is 
    {\it responsive} and {\it consistent} on this class.
    \begin{definition}[Order-independent Learning]
        A learning function $\varphi$  learns ${\cal G}$ {\rm order-independently}
        if for all $L \in \naming({\cal G})$, there exists $G \in {\cal G}$
        such that $\naming(G)=L$ and for all
        infinite sequences $\seq{s_i}_{i \in \naturals}$ that enumerate $L$,
        $\varphi$ converges on $\seq{s_i}_{i \in \naturals}$ to $G$.
    \end{definition}
    Intuitively this seems a reasonable strategy. There does not seem
    to be an a priori reason why either the order of presentation should influence the
    final choice of hypothesis. On the other hand, it has already
    been proved (see \cite{jors99}) that in any grammar system, a
    class of grammars is learnable if and only if there is a
    computable learning function that learns it
    order-independently.

    \begin{definition}[Exact Learning]
        A learning function $\varphi$ learns ${\cal G}$ {\rm exactly} if for all
        ${\cal G}^\prime$ such that $\varphi$ learns ${\cal G}^\prime$,
        $\naming({\cal G}^\prime) \subseteq \naming({\cal G})$.
    \end{definition}
    In other words, the learning function will not hypothesize grammars that are outside its
    class. This is not really a constraint on learning functions,
    but on the relation between a class of
    languages and a learning function. For every learning
    function there exists a class that it learns exactly.
    The reason for this constraint is the idea that children
    only learn languages that have at least a certain minimal
    expressiveness. If we want to model language learning,
    we want learning functions to learn a chosen class exactly.
    There seems to be empirical support for this idea.
    Some of it comes from studies of children raised in pidgin
    dialects, some from studies of sensory deprived
    children (see \cite{pinker94}).

    \begin{definition}[Prudent Learning]
        A learning function $\varphi$ learns ${\cal G}$ {\rm prudently} if
        $\varphi$ learns ${\cal G}$ and ${\rm range}(\varphi) \subseteq {\cal G}$.
    \end{definition}
    Note that prudent learning implies exact learning.
    This reduces to the condition that a learning function
    should only produce a hypothesis if the learning function can back up
    its hypotheses, i.e. if the hypothesis is confirmed by the input,
    the learning function is able to identify the language.

    \begin{definition}[Responsive Learning]
        A learning function $\varphi$ is {\rm responsive on} ${\cal G}$ if
        for any $L \in  \naming({\cal G}) $ and for any finite sequence
        $\seq{s_0, \ldots, s_i}$ of elements of
        $L\ (\set{\seq{ s_0, \ldots, s_i}} \subseteq L)$,
        $\varphi(\seq{s_0, \ldots, s_i})$ is defined.
    \end{definition}
    This constraint can be regarded as the complement of prudent learning:
    if all sentences found in the input are in a language in the class of languages
    learned, the learning function should always produce a hypothesis.

    \begin{definition}[Consistent Learning]
        A learning function $\varphi$ is {\rm consistent on} ${\cal G}$
        if for any $L \in \naming(G)$ and for any finite sequence
        $\seq{s_0, \ldots, s_i}$ of elements of $L$, either
        $\varphi(\seq{s_0, \ldots, s_i})$ is undefined
        or $\set{s_0, \ldots, s_i} \subseteq \naming(\varphi(\seq{s_0, \ldots, s_i}))$.
    \end{definition}
    The idea behind this constraint is that all the data given should be explained
    by the chosen hypothesis. It should be self-evident that this is a desirable
    property. Indeed, one would almost expect it to be part of the definition of learning.
    However, learning functions that are not consistent are not necessarily trivial.
    If, for example, the input is noisy, it would not be unreasonable for a learning function to
    ignore certain data because it considers them as unreliable.
    Also, it is a well known fact that children do not learn languages consistently.

    \subsubsection{Restrictive Constraints}
    \begin{definition}[Set-Drivenness]
        A learning function $\varphi$ learns ${\cal G}$ {\rm set-driven} if
        $\varphi(\seq{s_0, \ldots, s_i})$
        is determined by $\set{s_0, \ldots, s_i}$
        or, more precisely, if the following holds: whenever
        $\set{s_0, \ldots, s_i} = \set{u_0, \ldots, u_j}$,
        $\varphi(\seq{s_0, \ldots, s_i})$
        is defined if and only if $\varphi(\seq{u_0, \ldots, u_j})$
        is defined, and if they are defined, they are equal.
    \end{definition}
    It is easy to see that set-drivenness implies order-independence.
    Set-driven learning could be very loosely described as order-independent
    learning with the addition of ignoring ``doubles'' in the input. It is obvious
    that this is a nice property for a learning function to have: one would not
    expect the choice of hypothesis to be influenced by repeated presentation of
    the same data. The assumption here is that the order of presentation
    and the number of repetitions are essentially arbitrary, i.e. they carry no information that
    is of any use to the learning function.
    One can devise situations where this is not the case.

    \begin{definition}[Conservative Learning]
        A learning function $\varphi$ is {\rm conservative} if for any finite sequence
        $\seq{s_0, \ldots, s_i}$ of sentences and for any
        sentence~$s_{i+1}$,
        whenever $\varphi(\seq{s_0, \ldots, s_i})$ is defined and
        $s_{i+1} \in \naming(\varphi(\seq{s_0, \ldots, s_i}))$,
        $\varphi(\seq{s_0, \ldots, s_i, s_{i+1}})$ is also defined and
        $\varphi(\seq{s_0, \ldots, s_i})=\varphi(\seq{s_0, \ldots, s_i, s_{i+1}})$.
    \end{definition}
    At first glance conservatism may seem a desirable property.
    Why change your hypothesis if there is no direct need for it?
    One could imagine cases, however, where it would not be unreasonable for a
    learning function to change its mind, even though the new data fits in the
    current hypothesis. Such a function could for example make reasonable
    but ``wild'' guesses which it could later retract. The function could ``note'' after
    a while that the inputs cover only a proper subset of its conjectured language.
    While such behaviour will sometimes result in temporarily overshooting,
    such a function could still be guaranteed to converge to the correct hypothesis
    in the limit. \newline

    It is a common assumption in cognitive science that human
    cognitive processes can be simulated by computer. This would lead one 
    to believe that children's learning functions are computable.
    The corresponding strategy is the set of all partial and total recursive
    functions.
    Since this is only a subset of all possible functions, the computability strategy
    is a non trivial hypothesis, but not necessarily a restrictive one.

    The computability constraint interacts with consistency (see \cite{fulk88}):
    \begin{proposition}
        There is a collection of languages that is identifiable
        by a computable learning function
        but by no consistent, computable learning function.
    \end{proposition}

    The computability constraint also interacts with conservative 
    learning (see \cite{angluin80}):
    \begin{proposition}[Angluin, 1980]
        There is a collection of languages that is identifiable by a computable
        learning function but by no conservative, computable learning function.
    \end{proposition}

    \begin{definition}[Monotonicity]
        The learning function $\varphi$ is {\rm monotone increasing} if for all finite
        sequences $\seq{s_0, \ldots, s_n}$ and $\seq{s_0, \ldots, s_{n+m}}$,
        whenever $\varphi(\seq{s_0, \ldots, s_n})$ and
        $\varphi(\seq{s_0, \ldots, s_{n+m}})$ are defined,
        \begin{displaymath}
            \naming(\varphi(\seq{s_0, \ldots, s_n}))
            \subseteq \naming(\varphi(\seq{s_0, \ldots,
            s_{n+m}})).
        \end{displaymath}
    \end{definition}
    When a learning function that is monotone increasing changes its hypothesis,
    the language associated with the previous hypothesis will be (properly) included
    in the language associated with the new hypothesis.
    There seems to be little or no empirical support for such a constraint.

    \begin{definition}[Incrementality, Kanazawa 1998]
        The learning function $\varphi$ is {\rm incremental} if there exists a
        computable function $\psi$ such that
        \begin{displaymath}
            \varphi(\seq{s_0, \ldots, s_{n+1}})
            \simeq \psi(\varphi(\seq{s_0, \ldots, s_n}), s_{n+1}).
        \end{displaymath}
    \end{definition}
    An incremental learning function does not need to store previous data.
    All it needs is current input, $s_n$, and its previous hypothesis.
    A generalized form of this constraint, called {\it memory limitation},
    limits access for a learning function to only $n$ previous elements of the input
    sequence. This seems reasonable from an empirical point of view; it seems
    improbable that children (unconsciously) store all utterances they encounter.

    Note that, on an infinite sequence enumerating language $L$ in $\naming({\cal G})$,
    a conservative learning function $\varphi$ learning ${\cal G}$ never outputs
    any grammar that generates a proper superset of~$L$.\newline

    Let $\varphi$ be a conservative and computable learning function that is responsive
    and consistent on ${\cal G}$, and learns ${\cal G}$ prudently. Then, whenever
    $\set{s_0, \ldots, s_n} \subseteq L$ for some $L \in \naming({\cal G})$,
    $\naming(\varphi(\seq{s_0, \ldots, s_n}))$ must be a minimal element
    of the set
    $\set{L \in \naming({\cal G})\ |\ \set{s_0, \ldots, s_n}\subseteq L}$.
    This implies the following condition:

    \begin{condition}
        \label{cond1}
        There is a computable partial function $\psi$ that takes any
        finite set $D$ of sentences and maps it to a grammar
        $\psi(D)\in {\cal G}$ such that
        $\naming(\psi(D))$ is a minimal element of
        $\set{L \in \naming({\cal G})\ |\ D \subseteq L}$
        whenever the latter set is non-empty.
    \end{condition}

    \begin{definition}
        \label{prev}
        Let $\psi$ a computable function satisfying condition \ref{cond1}.
        Define a learning function $\varphi$ as follows
        \begin{eqnarray*}
            &&\varphi(\seq{s_0}) \simeq \psi(\set{s_0}),\\
            &&\varphi(\seq{s_0, \ldots, s_i+1}) \simeq
                \left \{
                    \begin{array}{rl}
                        \varphi(\seq{s_0, \ldots, s_i})  & 
                         \mbox{if\ } s_{i+1}\in \naming(\varphi(\seq{s_0, \ldots, s_i})),\\
                        \psi(\set{s_0, \ldots, s_{i+1}}) & \mbox{otherwise.}
                    \end{array}
                \right.
        \end{eqnarray*}
    \end{definition}
    Under certain conditions the function just defined
    is guaranteed to learn ${\cal G}$,
    one such case is where $\naming({\cal G}$ has finite elasticity.

    \begin{proposition}
        Let ${\cal G}$ be a class of grammars such that $\naming({\cal G})$ has finite
        elasticity, and a computable function $\psi$ satisfying condition \ref{cond1} exists.
        Then the learning function $\varphi$ defined in definition \ref{prev} learns ${\cal G}$.
    \end{proposition}

\section{Is Learning Theory Powerful Enough?}
    \subsection{First Negative Results}
    \label{resultsec}
    One of the main and apparently discouraging consequences of
    the theorem \ref{goldth} proved by Gold in the original article
    wherein he laid the foundations of Formal Learning
    Theory was that none of the four classes of Chomsky's
    Hierarchy is learnable under the criterion of identification in the limit.
    Such a first negative result has been
    taken for a long time as a proof that identifying
    languages from positive data according to his
    {\it identification in the limit} criterion was too hard a task.
    Gold himself looks quite pessimistic about the future of the
    theory he has just defined along its main directions:
    \begin{quote}
        However, the results presented in the last section show that only the most
        trivial class of languages considered is learnable... \cite{gold67}
    \end{quote}

    \subsection{Angluin's Results}
    The first example of non-trivial class of learnable grammars was
    discovered by Dana Angluin (see \cite{angluin80}).
    If $Pat$ is defined like in example \ref{angluinex},
    we can prove that the class
    of all pattern languages has finite elasticity and, therefore, it is learnable.
    Furthermore, such a learnable class of grammars was also the
    first example of an interesting class of grammars that
    cross-cuts Chomsky Hierarchy, therefore showing that Chomsky's
    is not but one of many meaningful possible classifications
    for formal grammars.

    \subsection{Shinohara's Results}
    Initial pessimism about effective usefulness
    of Gold's notion of identification in the limit
    was definitely abandoned after an impressive result by
    Shinohara who proves (see \cite{shinohara90a}),
    that {\it k-rigid context sensitive grammars}
    (context-sensitive grammars over a finite alphabet $\Sigma$ with at most $k$ rules),
    have finite elasticity for any $k$. Since the universal membership problem for
    context-sensitive grammars is decidable, that class of grammars is
    learnable. This is a particular case of his more general result about
    finite elasticity for what he calls {\it monotonic formal system}.

    \subsection{Kanazawa's Results}
    Makoto Kanazawa in \cite{kanazawa98} makes another decisive
    step toward bridging the existing gap between Formal
    Learning Theory and computational linguistics. Indeed, he gets
    some important results on the learnability for some
    non-trivial
    subclasses of Classical Categorial Grammars (also known as AB
    Grammars). Analogously to what is done in \cite{shinohara90a}
    he proves that as soon as we bound the maximum number of types
    a classical categorial grammar assigns to a word, we get
    subclasses which can be effectively learnable: in particular, he proves
    effective learnability for the class of {\it k-valued
    Classical Categorial Grammars}, both from structures and from
    strings.

    In the first case, each string of the language 
    the learner is presented to comes with
    additional information about the underlying structure 
    induced by the grammar formalism that generates the language. 
    The availability of such additional information for each string
    is somewhat in contrast with Gold's model of learning and gives 
    rise to weaker results. On the other hand, psychological plausibility 
    of the process is preserved by the fact that such an underlying 
    structure can be seen as some kind of semantic information that could be
    available to the child learning the language from 
    the very early stages of her cognitive development.

    \subsection{Our Results}
    The present work pushes Kanazawa's results a little further in the
    direction of proving the effective learnability for more and
    more powerful and expressive classes of formal languages. In
    particular, we will be able to prove learnability for the
    class of Rigid Lambek Grammars (see chapter
    \ref{learnlambeksection}) and to show an effective algorithm to
    learn them on the basis of a structured input.
    Much is left to be done along this direction of research,
    since even a formal theory for Rigid Lambek Grammars 
    is still under-developed. However,
    our results confirm once again that initial pessimism toward
    this paradigm of learning was largely unjustified, and that
    even quite a complex and linguistically motivated formalism
    like Lambek Grammars can be learnt according to it.

%% file: lambek_report.tex
\section{Lambek Grammars}
    In 1958 Joachim Lambek proposed (see \cite{lambek58}) to extend
    the formalism of Classical Categorial Grammars (sometimes referred
    to also as Basic Categorial Grammars or BCGs) 
    by a deductive system to derive type-change rules.
    A BCG is basically as a finite relation between the 
    finite set of symbols of the alphabet (usually
    referred to as {\it words}) and a finite set of types.
    Combinatory properties of each word are completely determined by the
    shape of its types, which can be combined according to a
    small set of rules, fixed once and for all BCGs.
    Lambek's proposal marked the irruption of logics into grammars:
    Lambek grammars come with a whole deductive system that allows
    the type of a symbol to be replaced with a weaker type.

    It was first realized by van Benthem (in \cite{benthem87}) that the proofs of these
    type changes principles carry important information about
    their {\it semantic} interpretation, following the
    Curry-Howard isomorphism. Thus, the notion of a proof
    theoretical grammar was proposed that replaces {\it formal
    grammars} (see \cite{chomsky56}) with {\it deductive systems}
    and that includes a systematic semantics for natural languages
    based on the relationship between proof theory and type
    theory. Thus, rather than considering grammatical categories
    as unanalyzed primitives, they are taken to be formulas
    constructed from atoms and connectives, and rather than
    defining grammars with respect to rewrite rules, grammars are
    defined by the rules of inference governing the connectives
    used in the syntactic categories.

    Due to the renewed interest in categorial grammars
    in the field of computational linguistics,
    Lambek (Categorial) Grammars (LCGs) are currently considered
    as a promising formalism. They enjoy the relative simplicity of a
    tightly constrained formalism as that for BCGs, together with the
    linguistically attractive feature of full lexicalization.

    Besides, although Pentus proved (in \cite{pentus97})
    that Lambek grammars generate exactly context-free (string) languages,
    in \cite{tiede99} it has been
    shown that their strong generative capacity is greater
    than that of context-free grammars.
    These features make them an interesting subject for our inquiry about
    their properties with respect to Gold's Learnability Theory.

    \subsection{Classical Categorial Grammars}
    \label{bcgsec}
    The main idea which lies behind the theory of Categorial Grammars
    is to conceive a grammar instead as a set of rules which generate any
    string of the language, as a system which assigns to each symbol of the alphabet
    a set of types which can be combined according to a small set of rules,
    fixed for the whole class of Classical Categorial Grammars.

    A context-free grammar {\it \'a la} Chomsky is made of a
    set of rules that generate all the strings of a given language
    in a ``top-down'' fashion, starting from an initial
    symbol which identifies all the well-formed strings. On the contrary, 
    a categorial grammar accepts a sequence of symbols of the
    alphabet as a well-formed string if and only if a sequence
    of types assigned to them {\it reduces} (in a ``bottom-up'' fashion) 
    according to a fixed set of rules, to a distinguished type which designates
    well-formed strings.
 
    \begin{definition}[Classical Categorial Grammar]\ \newline
        \label{CategorialGrammarDef}
        A Classical Categorial Grammar (henceforth CCG\index{CCG})
        is a quadruple $\seq{\Sigma, Pr, F, s}$,
        such that
        \begin{itemize}
            \item{$\Sigma$ is a finite set (the terminal symbols or vocabulary),}
            \item{Pr\index{Pr} is a finite set (the non-terminal symbols or atomic categories),}
            \item{F\index{F} is a function from $\Sigma$ to {\rm finite} subsets of Tp,
                 where Tp\index{Tp} is the smallest set such that:
                \begin{enumerate}
                    \item{$Pr \subseteq Tp$}
                    \item{if $A, B \in Tp$, then $(A/B), (A \backslash B) \in Tp$}
                \end{enumerate}
                If $F(a)=\set{A_1, \ldots, A_n}$ we usually write 
                $G: a \mapsto A_1, \ldots, A_n$.}
            \item{$s\in Pr$\index{s} is the distinguished atomic category}
        \end{itemize}
    \end{definition}

    In a CCG, combinatory properties are uniquely determined by
    their structure. There are only two modes of type combination:
    so-called (according to the notation introduced in \cite{lambek58}
    and almost universally adopted) {\it Backward Application}:
    \begin{displaymath}
        A, A\backslash B \Rightarrow B
    \end{displaymath}
    and {\it Forward Application}:
    \begin{displaymath}
        B/A, A \Rightarrow B.
    \end{displaymath}
    A non-empty sequence of types $A_1, \ldots, A_n$ is said to
    {\it derive} a type $B$, that is
    \begin{displaymath}
        A_1, \ldots, A_n \Rightarrow B,
    \end{displaymath}
    if repeated applications of the rules of Backward and Forward
    application to the sequence $A_1, \ldots, A_n$ results in $B$.\newline

    In order to define the language generated by a CCG 
    we have to establish a criterion to identify a string belonging 
    to that language. That's what is done by the following
    \begin{definition}
        \label{binreldef}
        The binary relation
        \begin{displaymath}
            \Rightarrow\index{$\Rightarrow$} \subseteq Tp^\ast \times Tp^\ast
        \end{displaymath}
        is defined as follows. Let $A, B \in Tp$, let $\alpha, \beta \in Tp^\ast$,
        \begin{eqnarray*}
            \alpha \ A \ A\backslash B \ \beta \ \Rightarrow \ \alpha \ B \ \beta \\
            \alpha \ B/A \ A \ \beta \ \Rightarrow \ \alpha \ B \ \beta
        \end{eqnarray*}
        The language generated by a CCG $G$ is the set
        \begin{displaymath}
        	\set{a_1\cdots a_n \in {\Sigma}^*\ |\ \mbox{for\ }1 \le i \le n,
        	\ \exists A_i,\ G: a_i\mapsto A_i,\ \mbox{and} 
        	\ A_1\ldots A_n \stackrel{*}{\Rightarrow} s}
        \end{displaymath}
        where $\stackrel{*}{\Rightarrow}$ is the reflexive,
        transitive closure of $\Rightarrow$.
    \end{definition}
    Informally, we can say that a string of symbols belongs 
    to the language generated by a CCG if there exists a derivation 
    of the distinguished category $s$ out of at least one sequence 
    of types assigned by the grammar to the symbols of the string.

    \begin{example}
        \label{CCGex1}
        The following grammar generates the language $\{a^n b^n\ |\ n>0\}$:
        \begin{eqnarray*}
            a & : & s/B,\\
            b & : & B,\ s\backslash B
        \end{eqnarray*}
        Here is a derivation for $a^3b^3$:
        \begin{eqnarray*}
            s/B\ s/B\ \underline{s/B\ B}\ s\backslash B\ s\backslash B\
            &\Rightarrow&\ s/B\ s/B\ \underline{s\ s\backslash B}\ s\backslash B\ \Rightarrow\\
            s/B\ \underline{s/B\ B}\ s\backslash B\ &\Rightarrow&\ s/B\ \underline{s\ s\backslash B}\ \Rightarrow\\
            \underline{s/B\ B}\ &\Rightarrow&\ s
        \end{eqnarray*}
    \end{example}

    Weak generative capacity of CCGs was characterized
    by Gaifman (see \cite{bar-hillel64}):
    \begin{theorem}[Gaifman, 1964]
        The set of languages generated by CCGs
        coincides with the set of context-free languages.
    \end{theorem}
    From the proof of Gaifman's theorem, we immediately
    obtain the following normal form theorem:
    \begin{theorem}[Gaifman normal form]
        Every categorial grammar is equivalent to a categorial 
        grammar which assigns only categories of the form
        \begin{displaymath}
            A, A/B, (A/B)/C.
        \end{displaymath}
    \end{theorem}

    \begin{example}
        \label{CCGex2}
        A CCG equivalent to that in example \ref{CCGex1}
        in Gaifman normal form is the following
        \begin{eqnarray*}
            a & : & s/B,\ (s/B)/s \\
            b & : & B
        \end{eqnarray*}
        and here is a derivation for $a^3b^3$:
        \begin{displaymath}
            \begin{prooftree}
                \begin{prooftree}
                    (s/B)/B
                    \begin{prooftree}
                        \begin{prooftree}
                            (s/B)/s
                            \begin{prooftree}
                                s/B\ \ \ B
                                \justifies
                                s
                            \end{prooftree}
                            \justifies
                            s/B
                        \end{prooftree}
                        B
                        \justifies
                        s
                    \end{prooftree}
                    \justifies
                    s/B
                \end{prooftree}
                B
                \justifies
                s
            \end{prooftree}
        \end{displaymath}
    \end{example}

    In the previous example we make use for the first time of a
    ``natural deduction'' notation for derivations, that in the present
    work will substitute the cumbersome notation used in
    example \ref{CCGex1}.

    \subsection{Extensions of Classical Categorial Grammars}
    \label{extensions}
    As stated in the previous section, CCG formalism comes with only two
    reduction rules which yield smaller types out of larger ones.
    Montague's work on semantics (see \cite{montague73}) led to
    the definition of two further ``type-raising'' rules, by which it is
    possible to construct new syntactic categories out of atomic ones.
    We can extend the definition of CCGs
    as presented in the previous section by adding to the former definition
    two new type change rules:
    \begin{eqnarray*}
        \alpha B \beta \Rightarrow \alpha (A/B)\backslash A \beta\\
        \alpha B \beta \Rightarrow \alpha A/(B\backslash A) \beta
    \end{eqnarray*}
    Other type-change rules that were proposed are the composition:
    \begin{displaymath}
        \frac{A/B \ \ B/C}{A/C} \ \ \ \frac{C\backslash B \ \ B\backslash A}{C \backslash A}
    \end{displaymath}
    and the Geach Rules:
    \begin{displaymath}
        \frac{A/B}{(A/C)/(B/C)} \ \ \ \frac{B \backslash A}{(C\backslash B)\backslash(C \backslash A)}
    \end{displaymath}

    We can extend the formalism of CCG by adding to definition
    \ref{binreldef} any type change rule we need to formalize specific phenomena in
    natural language. Such a {\it rule-based} approach was adopted by Steedman
    (see \cite{steedman93}) who enriches classical categorial grammar formalism 
    with a {\it finite} number of type-changes rules.
    On the other hand, as it will be made clear in the following section,
    Lambek's approach is a {\it deductive} one: he defines
    a calculus in which type changes rules spring out as a consequence
    of the operations performed on the types.

    One could ask why we should follow the deductive 
    rather than the rule-based approach.
    To begin with, as proved in \cite{zielonka89}, 
    Lambek Calculus {\it is not finitely axiomatizable},
    that is to say that adding a finite number of type-change 
    rules to the formalism of CCG one cannot derive all the 
    type change rules provable in the Lambek
    Calculus. Moreover, the two approaches are very 
    different under a theoretical viewpoint.

    From a linguistic perspective, Steedman pointed out 
    that there is no reason why we should stick to a deductive 
    approach instead of to a rule based one: he underlines the importance
    of introducing {\it ad hoc} rules to formalize specific 
    linguistic phenomena. Why should we subordinate the use of 
    specific type change rules to their derivability in some calculus?

    One of the most compelling reasons to do so is given by 
    Moortgat (see \cite{moortgat97}) who stresses the systematicity 
    of the relation between syntax and semantics provided
    in a deductive framework. Also, Lambek Calculus enjoys an important property:
    it is sound and complete with respect to free semigroup model, i.e. an interpretation
    with respect to formal languages . 
    That is to say, rules that are not deducible in Lambek Calculus
    are not sound, and so they can be considered as linguistically implausible.

    \subsection{(Associative) Lambek Calculus}
    Categorial grammars can be analyzed from a proof
    theoretical perspective by observing the close
    connection between the ``slashes'' of a categorial
    grammar and implication in intuitionistic logics.
    The rule that allows us
    to infer that if {\it w} is of  type $A/B$ and {\it v}
    is of type $B$, then {\it wv} is of type $A$,
    behaves like the {\it modus ponens} rule of inference in logic.
    On the basis of this similarity Lambek proposed an
    architecture for categorial grammars based on two
    levels:
    \begin{itemize}
        \item{a syntactic calculus, i.e. a deductive system 
            in which statement of the form
            \begin{displaymath}
                A_1,\ldots,A_n \vdash B,
            \end{displaymath}
            to be read ``from the types $A_1,\ldots,A_n$ 
            we can infer type $B$'' can be proved;}
        \item{a categorial grammar as presented in definition~\ref{CategorialGrammarDef},
            wherein the relation $\Rightarrow$ is changed to
            allow any type change rule that could be deduced
            at the previous level.}
    \end{itemize}
    In doing so, instead of adding a finite number of type
    change rules to our grammar, every type change rule
    that can be derived in the Lambek Calculus is added to the
    categorial grammar.

    The following formalizations for Lambek Calculus  are presented
    according, respectively, to the formalism
    of sequent calculus and to the formalism of natural deduction.
    Note that in the present work we will use the expression {\it Lambek
    Calculus} to refer to {\it product-free Lambek Calculus}:
    indeed we will never make use of the product `$\cdot$' (which
    corresponds to the tensor of linear logic).

    \begin{definition}
    The sequent calculus formalization of the Lambek
    calculus contains the axiom [ID] and the rules of
    inference [/R], [/L], [$\backslash R$], [$\backslash L$],
    and [Cut]:
        \begin{displaymath}
            \begin{prooftree}
                \justifies
                A \vdash A
                \using [ID]
            \end{prooftree}
        \end{displaymath}
        \begin{displaymath}
            \begin{prooftree}
                \Gamma, A \vdash B
                \justifies
                \Gamma \vdash B/A
                \using [/R]
            \end{prooftree}\ \ \ \ \
            \begin{prooftree}
                \Gamma \vdash A \ \ \
                \Delta,B,\Pi \vdash C
                \justifies
                \Delta,B/A,\Gamma,\Pi \vdash C
                \using [/L]
            \end{prooftree}
        \end{displaymath}
        \begin{displaymath}
            \begin{prooftree}
                A, \Gamma \vdash B
                \justifies
                \Gamma \vdash A \backslash B
                \using [\backslash R]
            \end{prooftree}\ \ \ \ \
            \begin{prooftree}
                \Gamma \vdash A \ \ \
                \Delta, B, \Pi \vdash C
                \justifies
                \Delta, \Gamma, A \backslash B, \Pi \vdash C
                \using [\backslash L]
            \end{prooftree}
        \end{displaymath}
        \begin{displaymath}
            \begin{prooftree}
                \Delta \vdash B \ \ \
                \Gamma, B, \Pi \vdash A
                \justifies
                \Gamma, \Delta, \Pi \vdash A
                \using [Cut]
            \end{prooftree}
        \end{displaymath}
        Note: in $[/R]$ and $[\backslash R]$ there is a side condition stipulating that
        $\Gamma \not= \emptyset$.

    \end{definition}
    The side condition imposed for $[/R]$ and $[\backslash R]$ 
    rules formalizes the fact that in Lambek Calculus one is not 
    allowed to cancel all the premises from the left-hand side
    of a derivation. Otherwise stated, in Lambek Calculus 
    there are no deductions of the form
    \begin{displaymath}
        \vdash A.
    \end{displaymath}

    Coherently with our interpretation of Lambek Calculus 
    as a deductive system to derive the type of a sequence 
    of symbols of the alphabet out of the types of each symbol, 
    such a derivation makes no sense, since it would mean 
    assigning a type to an empty sequence of words.

    \begin{definition}
        The natural deduction formalization of the Lambek
        Calculus is defined as follows:
        \begin{displaymath}
            A \ \ [ID]
        \end{displaymath}
        \begin{displaymath}
            \begin{prooftree}
                \begin{prooftree}
                    \proofdotnumber=4
                    \leadsto A/B
                \end{prooftree}
                \begin{prooftree}
                    \proofdotnumber=4
                    \leadsto B
                \end{prooftree}
                \justifies A
                \using [/E]
            \end{prooftree}
            \ \ \
            \begin{prooftree}
                \begin{prooftree}
                    \proofdotnumber=4
                    \leadsto B
                \end{prooftree}
                \begin{prooftree}
                    \proofdotnumber=4
                    \leadsto B\backslash A
                \end{prooftree}
                \justifies A
                \using [\backslash E]
            \end{prooftree}
        \end{displaymath}
        \begin{displaymath}
            \begin{prooftree}
                \begin{prooftree}
                    \ \ \ \ \ \ \ [B]
                    \proofdotnumber=4
                    \leadsto A
                \end{prooftree}
                \justifies A/B
                \using [/I]
            \end{prooftree}
            \ \ \
            \begin{prooftree}
                \begin{prooftree}
                    [B] \ \ \ \ \ \ \
                    \proofdotnumber=4
                    \leadsto A
                \end{prooftree}
                \justifies B \backslash A
                \using [\backslash I]
            \end{prooftree}
        \end{displaymath}
    Note: in [/I] and [$\backslash$I] rules the cancelled 
    assumption is always, respectively, the rightmost
    and the leftmost uncancelled assumption, and there 
    must be at least another uncancelled hypothesis.
    \end{definition}

    Both formalisms have advantages and disadvantages. However, due to the close
    connection between natural deduction proofs and $\lambda$-terms and
    because the tree-like structure of deductions resembles
    derivations trees of grammars, the natural deduction version
    will be the primary object of study in the
    present work.

    For later purposes we introduce here the notion of {\it
    derivation} in Lambek calculus that will be useful
    later for the definition of the {\it structure} of a sentence
    in a Lambek grammar.
    A derivation of $B$ from $A_1, \ldots, A_n$ is a
    certain kind of unary-binary branching
    tree that encodes a proof of $A_1, \ldots, A_n \vdash B$.
    Each node of a derivation
    is labeled with a type, and each internal node has an additional label which, for
    Lambek grammars, is either $/E, \backslash E, /I$, or $\backslash I$ and
    that indicates which Lambek calculus rule is used at each step
    of a derivation. For each occurrence of an introduction
    rule there must be a corresponding previously unmarked leaf type $A$
    which must be marked as $[A]$ (that corresponds to ``discharging'' 
    an assumption in natural deduction).\newpage

    The set of derivations is inductively defined as follows:
    \begin{definition}
        \label{derivationdef}
        Let $A, B \in {\rm Tp}$ and $\Gamma, \Delta \in {\rm Tp}^+$,
        \begin{itemize}
            \item $A$ (the tree consisting of a single node
                labeled by $A$) is a derivation of $A$ from $A$.
            \item {\bf "Backslash elimination".} If
                \begin{figure}[htbp]
                    \begin{center}
                        \includegraphics[height=2.2cm]{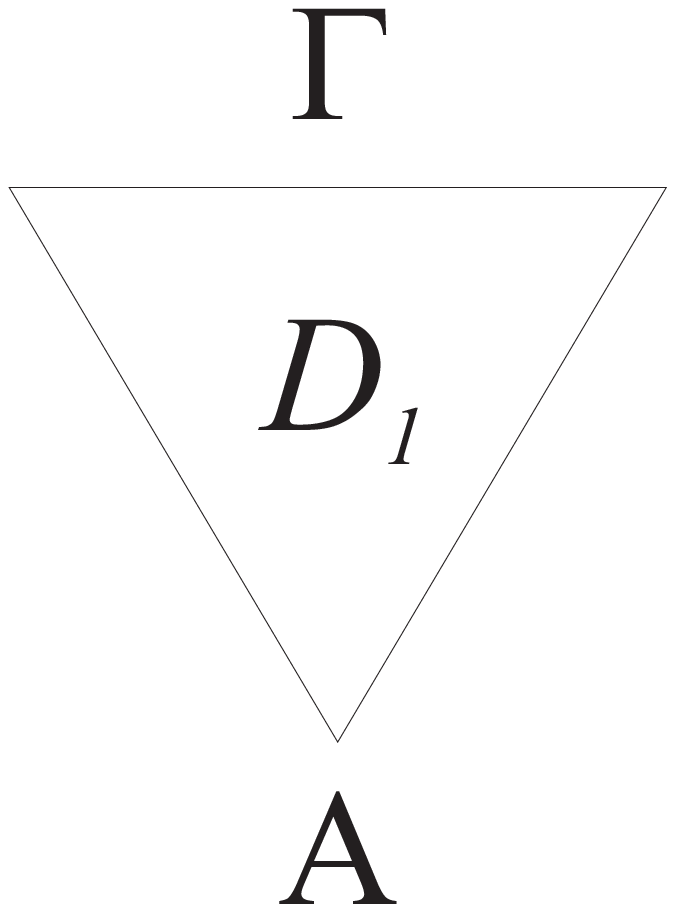}
                    \end{center}
                \end{figure}\newline
                is a derivation of $A$ from $\Gamma$ and
                \begin{figure}[htbp]
                    \begin{center}
                        \includegraphics[height=2.2cm]{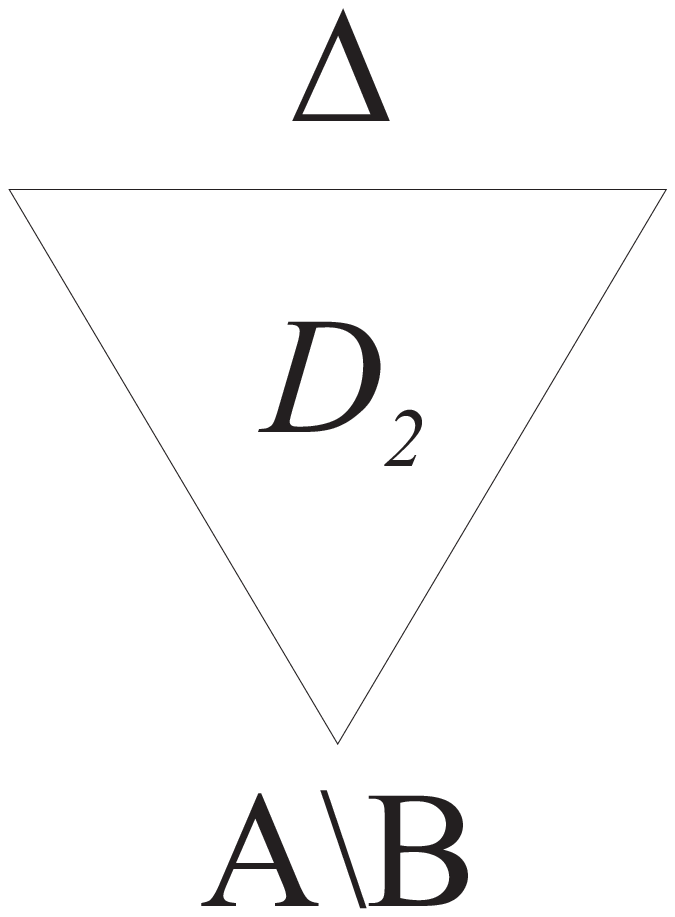}
                    \end{center}
                \end{figure}\newline
                is a derivation of $A\backslash B$ from $\Delta$, then
                \begin{figure}[htbp]
                    \begin{center}
                        \includegraphics[height=3.5cm]{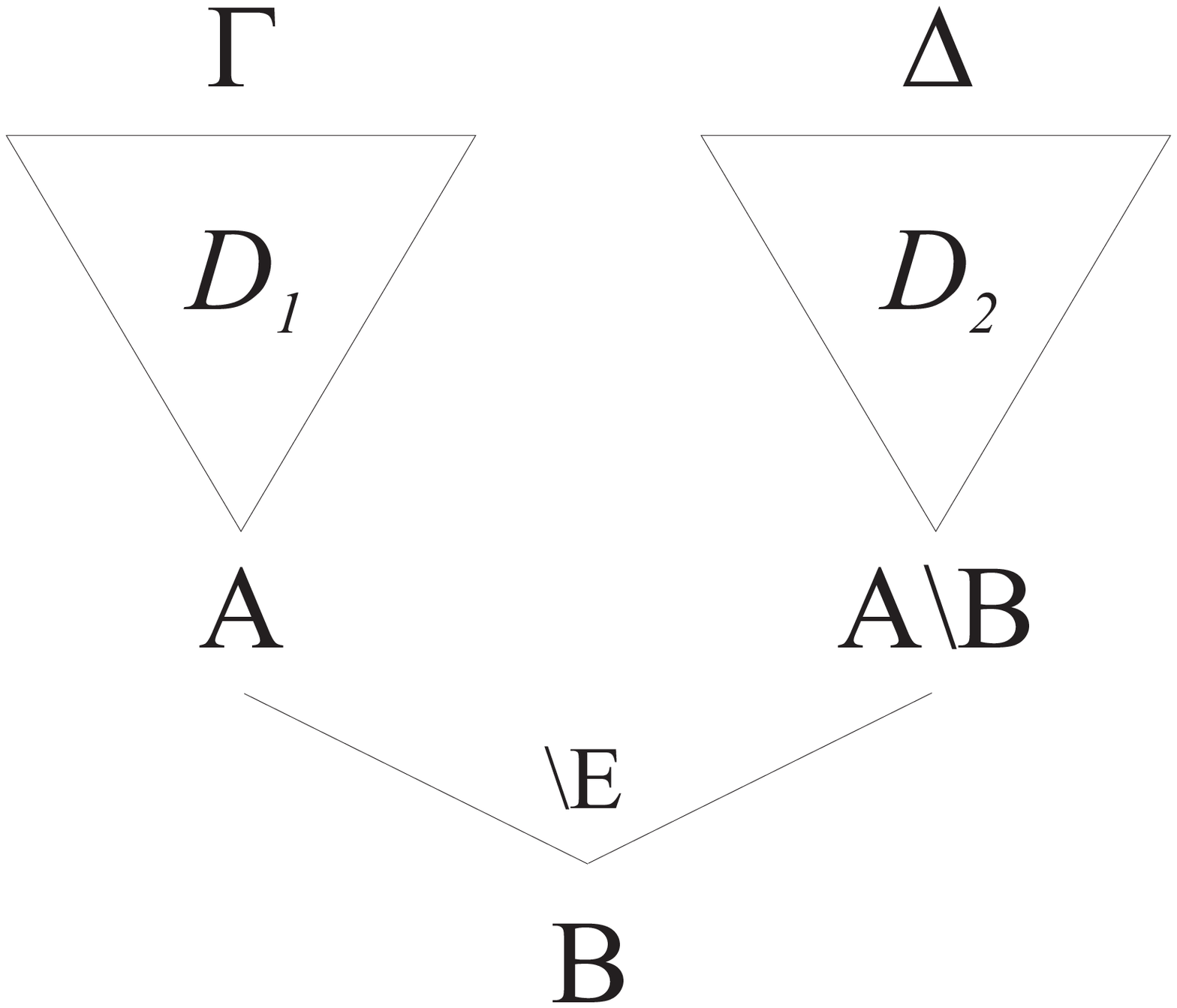}
                    \end{center}
                \end{figure}\newline
                is a derivation of $B$ from $\Gamma,
                \Delta$.\newpage
            \item {\bf "Backslash introduction".} If
                \begin{figure}[htbp]
                    \begin{center}
                        \includegraphics[height=2.2cm]{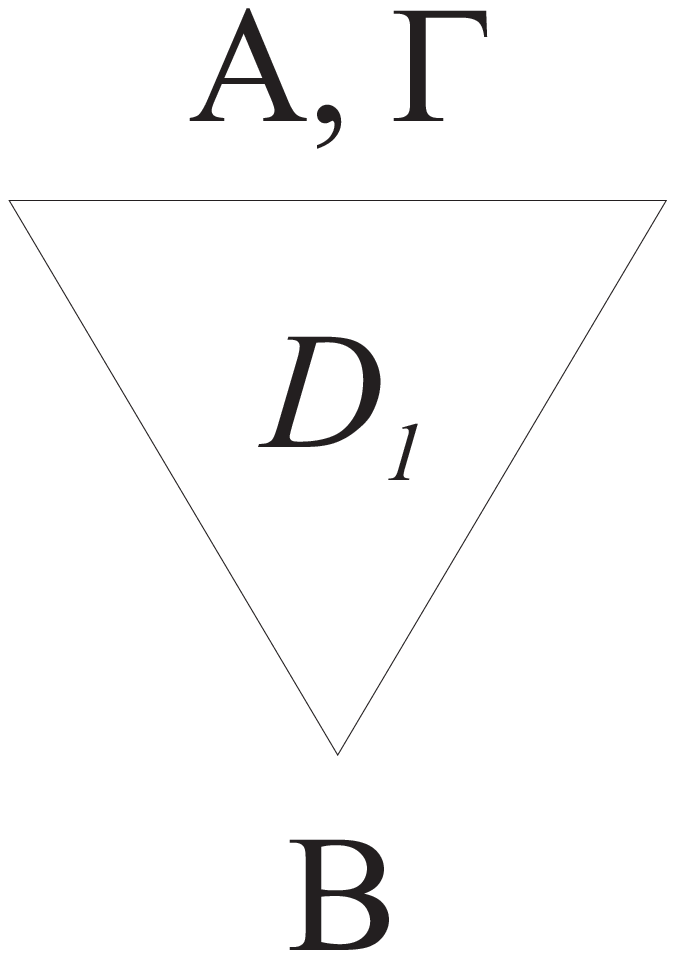}
                    \end{center}
                \end{figure}\newline
                is a derivation of B from $\set{A, \Gamma}$,
                then
                \begin{figure}[htbp]
                    \begin{center}
                        \includegraphics[height=3.6cm]{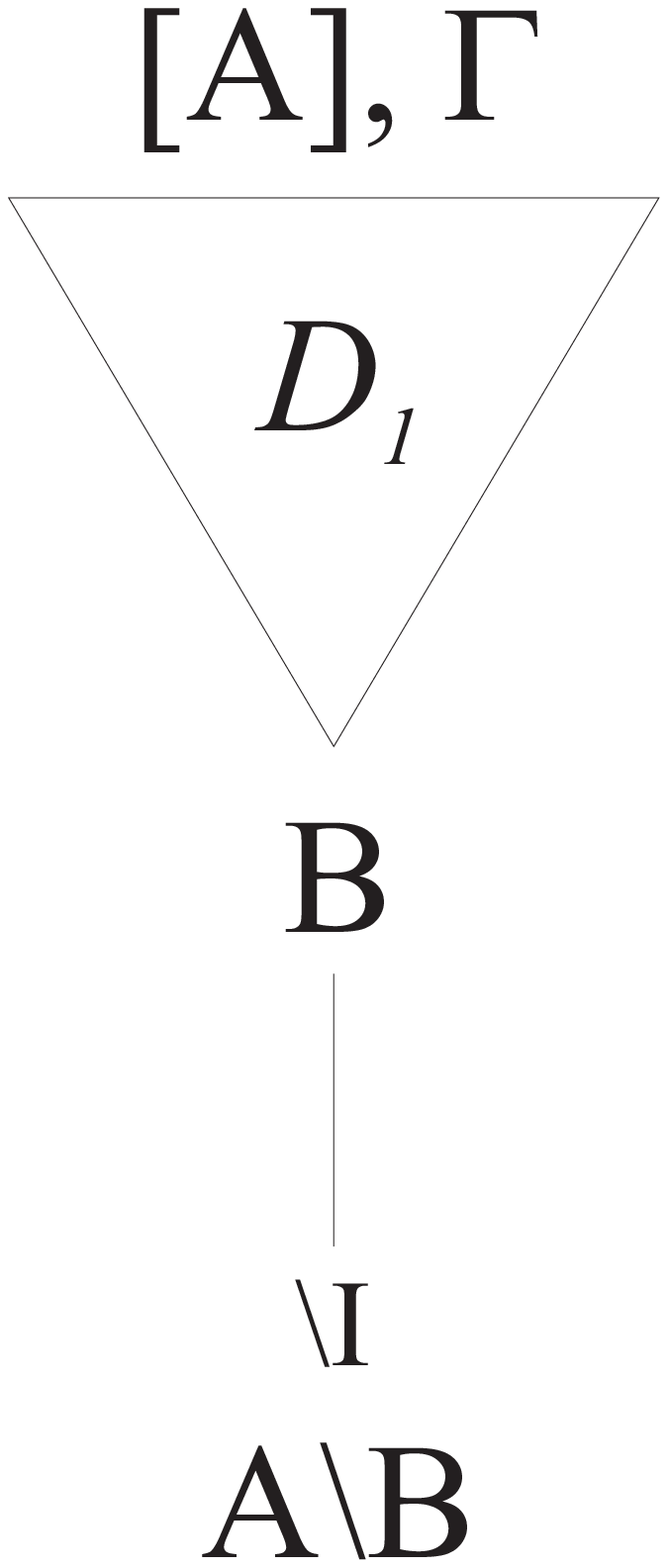}
                    \end{center}
                \end{figure}\newline
                is a derivation of A$\backslash$B from $\Gamma$. The leaf labeled by
                {\rm [A]} is called a {\rm discharged}
                leaf.\newpage
            \item {\bf "Slash elimination".} If
                \begin{figure}[htbp]
                    \begin{center}
                        \includegraphics[height=2.2cm]{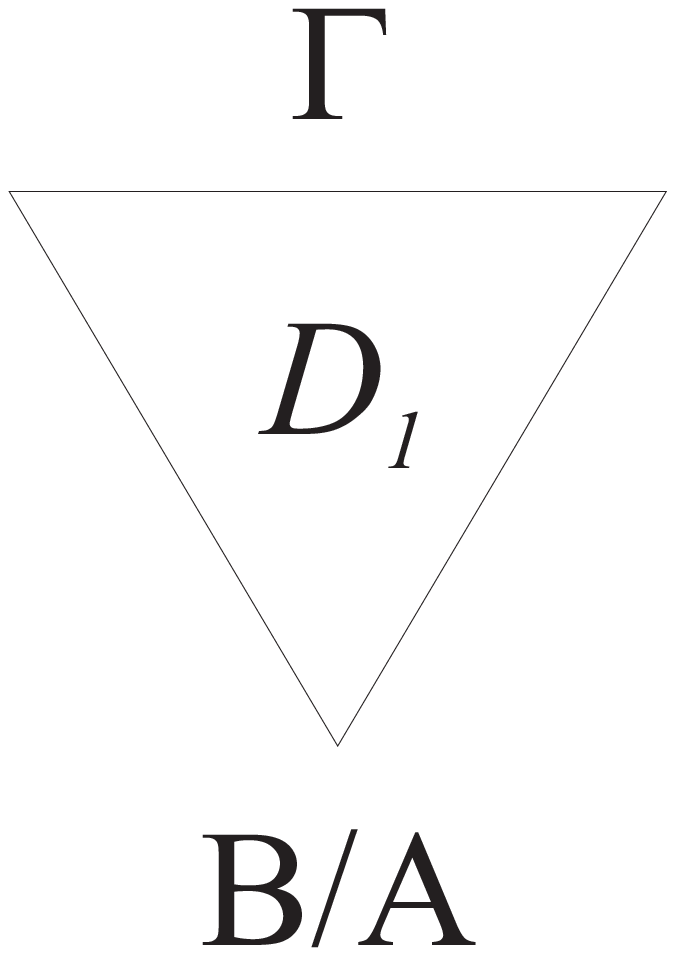}
                    \end{center}
                \end{figure}\newline
                is a derivation of B/A from $\Gamma$ and
                \begin{figure}[htbp]
                    \begin{center}
                        \includegraphics[height=2.2cm]{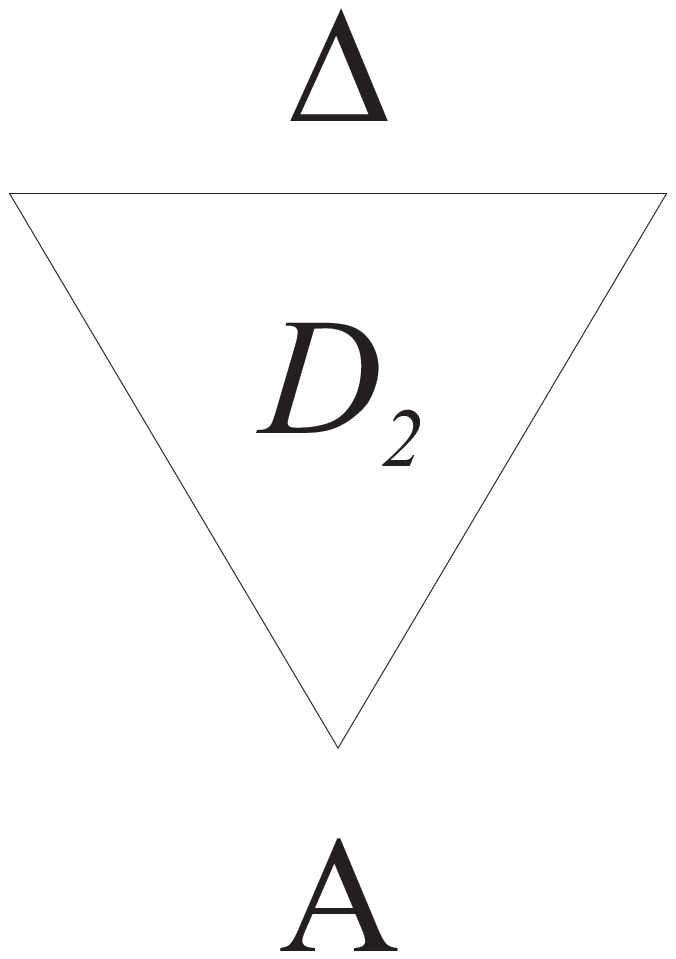}
                    \end{center}
                \end{figure}\newline
                is a derivation of A from $\Delta$, then
                \begin{figure}[htbp]
                    \begin{center}
                        \includegraphics[height=3.5cm]{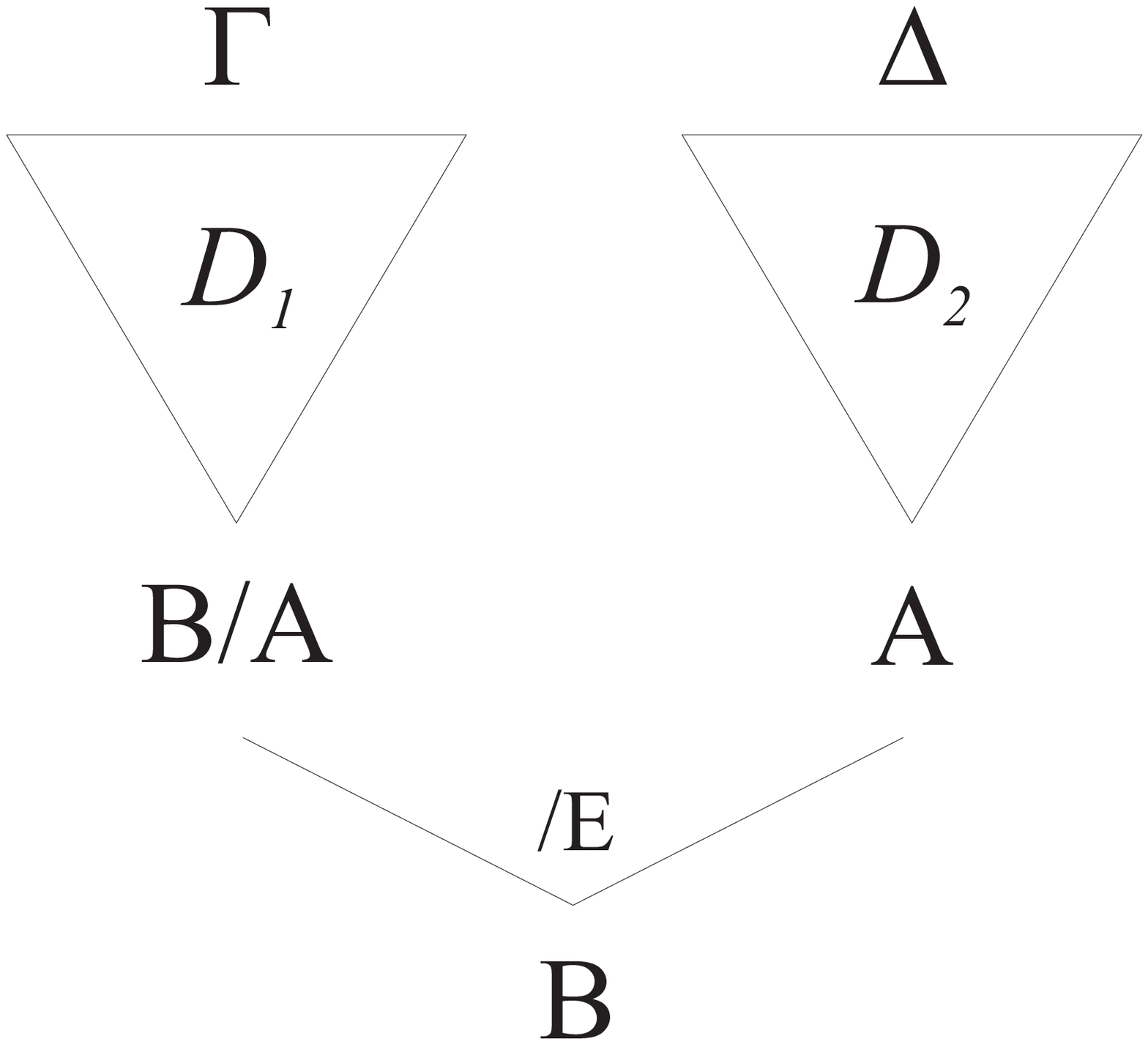}
                    \end{center}
                \end{figure}\newline
                is a derivation of B from $\Gamma$, $\Delta$.\newpage
	\item {\bf "Slash introduction".} If
                \begin{figure}[htbp]
                    \begin{center}
                        \includegraphics[height=2.2cm]{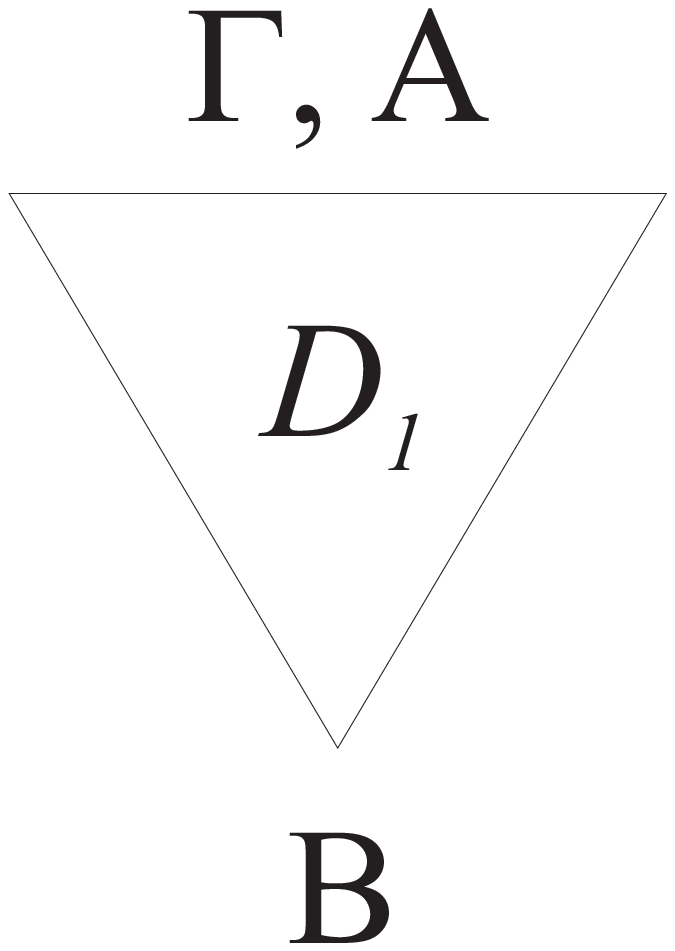}
                    \end{center}
                \end{figure}\newline
                is a derivation of B from $\{\Gamma, A\}$ then
                \begin{figure}[htbp]
                    \begin{center}
                        \includegraphics[height=3.6cm]{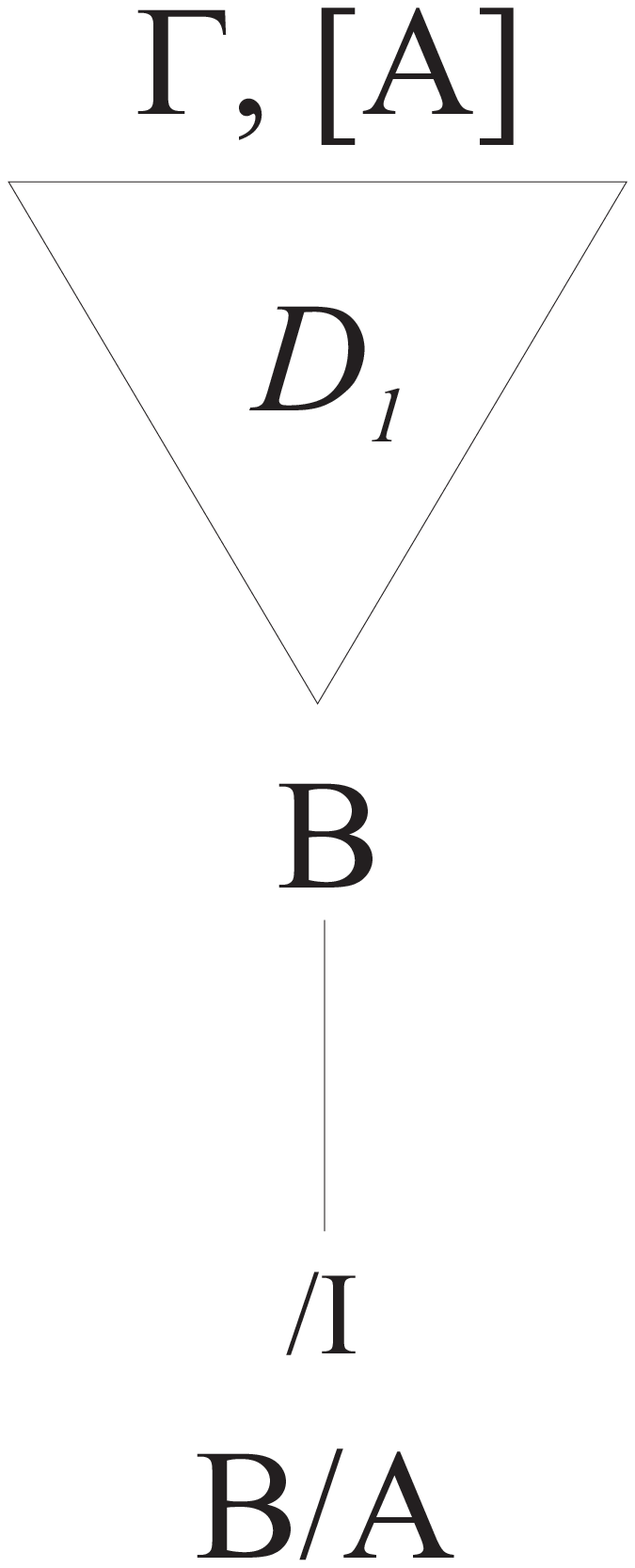}
                    \end{center}
                \end{figure}\newline
                is a derivation of B/A from $\Gamma$. The leaf labeled by {\rm [A]}
                is called a {\rm discharged} leaf.
            \end{itemize}
    \end{definition}
    \begin{example}
        \label{type-raising1}
        The following example is a derivation
        of $x$ from $y/(x\backslash y)$ (which
        proves one of the two {\it type-raising} rules in Lambek Calculus):
        \begin{figure}[htbp]
            \begin{center}
                \includegraphics[height=3.5cm]{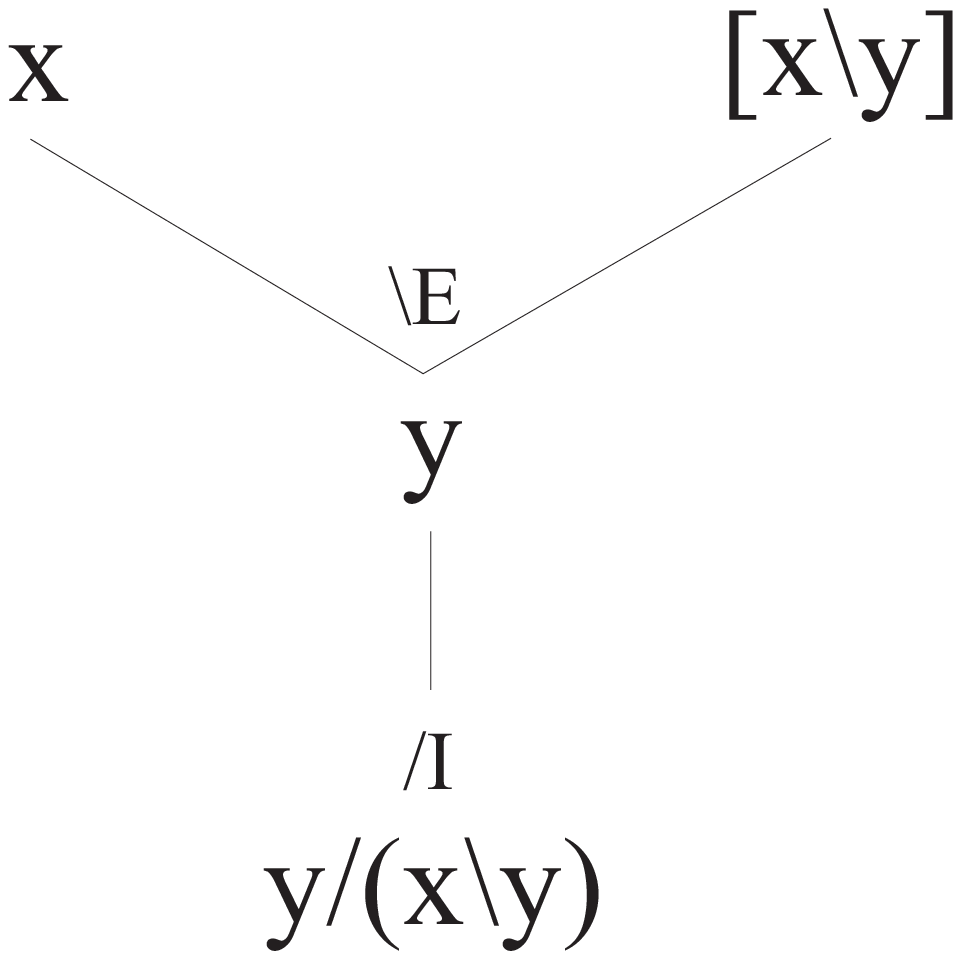}
            \end{center}
        \end{figure}
    \end{example}

    \subsection{Non-associative Lambek Calculus}
    Lambek Calculus, as defined in the previous
    section, is implicitly associative. In order to use Lambek
    calculus to describe some linguistic phenomena we
    have to forbid associativity and so  the hierarchical
    embedding of hypotheses is respected. Another linguistically
    attractive feature of non-associative Lambek calculus is that
    it provides useful logical to support semantics,
    but at the same time it prohibits transitivity, that sometimes leads to
    overgeneration.

    \begin{definition}
        The natural deduction formalization of the
        non-associative Lambek Calculus (SND) has the following
        axioms and rules of inference, presented in the
        sequent format:
        \begin{displaymath}
            \begin{prooftree}
                \justifies A \vdash A
                \using [ID]
            \end{prooftree}
        \end{displaymath}
        \begin{displaymath}
            \begin{prooftree}
                \Gamma \vdash A/B \ \
                \Delta \vdash B
                \justifies (\Gamma, \Delta) \vdash A
                \using [/E]
            \end{prooftree}
            \ \ \
            \begin{prooftree}
                \Gamma \vdash B \ \
                \Delta \vdash B \backslash A
                \justifies (\Gamma, \Delta) \vdash A
                \using [\backslash E]
            \end{prooftree}
        \end{displaymath}

        \begin{displaymath}
            \begin{prooftree}
                (\Gamma,B) \vdash A
                \justifies \Gamma \vdash A/B
            \using [/I]
            \end{prooftree}
            \ \ \
            \begin{prooftree}
                (B,\Gamma) \vdash A
                \justifies \Gamma \vdash B \backslash A
                \using [\backslash I]
            \end{prooftree}
        \end{displaymath}
        Note: in $[/I]$ and $[\backslash I]$ there is a side condition stipulating that
            $\Gamma \not= \emptyset$.
    \end{definition}

    \subsection{Normalization and Normal Forms}
    \label{normsec}
    As one can easily see, in Lambek Calculus there are infinitely many proofs
    for any deduction $A_1, \ldots, A_n \vdash B$.
    Since, as it will be extensively explained in section \ref{ptreessec},
    proofs in Lambek Calculus play a decisive role in defining the notion
    of {\it structure} for a sentence generated by a Lambek grammar, such an
    arbitrary proliferation of proofs for deductions is quite undesirable.

    The following definition introduces a useful
    relation between proofs in Lambek Calculus
    that formalizes our idea of a ``minimal'' proof for any deduction.
    It provides two normalization schemes
    that can be applied to a derivation
    to produce a ``simpler'' derivation of the same result.

    \begin{definition}
        The relation $>_1$ between proofs in the natural deduction formalization of
        Lambek Calculus is defined in the following way:
        \begin{displaymath}
            \begin{prooftree}
                \begin{prooftree}
                    \proofdotnumber = 4
                    \leadsto A
                \end{prooftree}
                \begin{prooftree}
                    \begin{prooftree}
                        [A] \ \ \ \
                        \proofdotnumber = 4
                        \leadsto B
                    \end{prooftree}
                    \justifies A \backslash B
                    \using [\backslash I]
                \end{prooftree}
                \justifies B
                \using [\backslash E]
            \end{prooftree}
            >_1
            \begin{prooftree}
                \begin{prooftree}
                    \proofdotnumber = 4
                    \leadsto A
                \end{prooftree}
                \begin{prooftree}
                \end{prooftree}
                \leadsto
                \begin{prooftree}
                    \proofdotnumber = 1
                    \leadsto B
                \end{prooftree}
            \end{prooftree}
            \ \ \ \ \ \ \ \ \ \
            \begin{prooftree}
                \begin{prooftree}
                    \begin{prooftree}
                        \ \ \ \ [A]
                        \proofdotnumber = 4
                        \leadsto B
                    \end{prooftree}
                    \justifies B / A
                    \using [ / I]
                \end{prooftree}
                \begin{prooftree}
                    \proofdotnumber = 4
                    \leadsto A
                \end{prooftree}
                \justifies B
                \using [/ E]
            \end{prooftree}
            >_1
            \begin{prooftree}
                \begin{prooftree}
                \end{prooftree}
                \begin{prooftree}
                    \proofdotnumber = 4
                    \leadsto A
                \end{prooftree}
                \leadsto
                    \begin{prooftree}
                        \proofdotnumber = 1
                        \leadsto B
                    \end{prooftree}
            \end{prooftree}
        \end{displaymath}

        \begin{displaymath}
            \begin{prooftree}
                \begin{prooftree}
                    [B] \ \ \ \ \
                    \begin{prooftree}
                        \proofdotnumber = 4
                        \leadsto B \backslash A
                    \end{prooftree}
                    \justifies A
                    \using [\backslash E]
                \end{prooftree}
                \justifies B \backslash A
                \using [\backslash I]
            \end{prooftree}
            >_1
            \begin{prooftree}
                \proofdotnumber = 4
                \leadsto B \backslash A
            \end{prooftree}
            \ \ \ \ \ \ \ \ \ \
            \begin{prooftree}
                \begin{prooftree}
                    \begin{prooftree}
                        \proofdotnumber = 4
                        \leadsto A / B
                    \end{prooftree}
                    [B]
                    \justifies A
                    \using [/E]
                \end{prooftree}
                \justifies A / B
                \using [/I]
            \end{prooftree}
            >_1
            \begin{prooftree}
                \proofdotnumber = 4
                \leadsto A / B
            \end{prooftree}
        \end{displaymath}
        The symbol $\ge$ stands for reflexive and transitive 
        closure of $>_1$. Relation $>_1$ is usually defined 
        as {\it $\beta$-$\eta$-conversion}, while $\ge$ 
        as {\it $\beta$-$\eta$-reduction}.
    \end{definition}
    The relation $\ge$ satisfies the following properties 
    (see \cite{wansing93}, \cite{roorda91}):
    \begin{theorem}[Wansing, 1993]
        \label{wansingth}
        The relation $\ge$ is confluent (in the Church-Rosser meaning), i.e.
        if $\delta_1 \ge \delta_2$ and $\delta_1 \ge \delta_3$, then there exists
        a $\delta_4$ such that $\delta_2 \ge \delta_4$ and $\delta_3 \ge \delta_4$.
    \end{theorem}

    \begin{theorem}[Roorda, 1991]
        \label{wsnorm}
        The relation $\ge$ is both weakly and strongly normalizing, 
        that is, every proof can be reduced
        in normal form and every reduction terminates 
        after at most a finite number of steps.
    \end{theorem}

    \begin{definition}[$\beta$-$\eta$-normal form]
        \label{normalformdef}
        A proof tree for the Lambek Calculus is said to be 
        in {\rm $\beta$-$\eta$-normal form} is none of its subtrees 
        is of the form
        \begin{displaymath}
            \begin{prooftree}
                \begin{prooftree}
                    \begin{prooftree}
                        \ \ \ \ [B]
                        \proofdotnumber=4
                        \leadsto A
                    \end{prooftree}
                    \justifies A/B
                    \using [/I]
                \end{prooftree}
                B
                \justifies A
                \using [/E]
            \end{prooftree}
            \ \ \
            \begin{prooftree}
                B
                \begin{prooftree}
                    \begin{prooftree}
                        [B] \ \ \ \
                        \proofdotnumber=4
                        \leadsto A
                    \end{prooftree}
                    \justifies B \backslash A
                    \using [\backslash I]
                \end{prooftree}
                \justifies A
                \using [\backslash E]
            \end{prooftree}
        \end{displaymath}

        \begin{displaymath}
            \begin{prooftree}
                \begin{prooftree}
                    A / B \ \ \ \
                    [B]
                    \justifies A
                    \using [/E]
                \end{prooftree}
                \justifies A/B
                \using [/I]
            \end{prooftree}
            \ \ \ \
            \begin{prooftree}
                \begin{prooftree}
                    [B] \ \ \ \
                    B \backslash A
                    \justifies A
                    \using [\backslash E]
                \end{prooftree}
                \justifies B \backslash A
                \using [\backslash I]
            \end{prooftree}
        \end{displaymath}
    \end{definition}
  
    \subsection{Basic Facts about Lambek Calculus}
    Let's summarize here some meaningful properties for Lambek
    calculus, which is:
    \begin{itemize}
        \item {\it intuitionistic}: only one formula is allowed on
        the right-hand side of a deduction. This means there is
        neither involutive negation, nor disjunction;
        \item {\it linear}: so-called structural rules of logics
        are not allowed: two equal hypotheses can't be considered as
        only one, and on the other hand we are not allowed to
        ``duplicate'' hypotheses at will. Lambek calculus is what we call a
        resource-aware logics, wherein hypotheses must be
        considered as consumable resources;
        \item {\it non-commutative}: hypotheses don't commute
        among them, that is, the implicit operator ``$\cdot$'' in
        this calculus is not commutative. This is what makes
        possible the existence of the two ``implications'' ($/$ and
        $\backslash$), the first one consuming its right argument,
        the second one its left argument.
    \end{itemize}

    Since Lambek proved a cut-elimination theorem for his 
    calculus (see \cite{lambek58}), among the many consequences of 
    the normalization theorems there are the subformula property, that is:
    \begin{proposition}
        Every formula that occurs in a normal form natural deduction proof of
        cut-free sequent calculus proof is either a subformula of the (uncancelled)
        assumptions or of the conclusion;
    \end{proposition}
    and decidability for Lambek calculus:
    \begin{proposition}
        \label{derivabilityprop}
        Derivability in the Lambek Calculus is decidable.
    \end{proposition}
    In fact, given a sequent to prove in Lambek calculus,
    cut-elimination property authorizes us to look for a cut-free
    proof. But if the sequent comes from the application of a rule
    other that cut, this can't but be made in a finite number of
    different ways, and in any case we have to prove one or two
    smaller (i.e. with less symbols) sequents. This is enough to
    prove decidability for Lambek calculus.\newline

    Theorem \ref{wsnorm} states that any proof has a normal form
    and theorem \ref{wansingth} that this normal form is unique.
    This doesn't mean that there is a unique normal form proof for any deduction.
    The following theorem by van Benthem sheds light on this point:
    \begin{theorem}[van Benthem]
        \label{benthemth}
        For any sequent
        \begin{displaymath}
            A_1, \ldots, A_n \vdash B
        \end{displaymath}
        there are only finitely many different normal form proofs in the Lambek Calculus.
    \end{theorem}

    This is quite an unsatisfactory result: we still have a
    one-to-many correspondence between a sequent and its
    normal proofs. This leads to what is generally known as the
    problem of {\it spurious ambiguities} for Lambek grammars.

    \subsection{Lambek Grammars}
    A Lambek grammar extends the traditional notion of categorial
    grammars as presented in section \ref{bcgsec} by a whole deductive
    system in the following way:
    \begin{itemize}
        \item a lexicon assigns to each word $w_i$ a finite set of
        types
        \begin{displaymath}
            F(w_i)=\set{t_i^1, \ldots, t_i^{k_i}} \subset \wp(Tp);
        \end{displaymath}
        \item the language generated by this fully lexicalized grammar
        is the set of all the sequences $w_1 \cdots w_n$ of words
        of the lexicon such that for each $w_i$ there exists a
        type $t_i \in F(w_i)$ such that 
        \begin{displaymath}
		t_1, \ldots, t_n \vdash s
	\end{displaymath}
        is provable in Lambek calculus.
    \end{itemize}

    Formally:
    \begin{definition}[Lambek grammar]
        \label{LambekGrDef}
        A Lambek grammar is a triple
        $G=\seq{\Sigma,s,F}$, such that
        \begin{itemize}
            \item{$\Sigma$ is a finite set (the vocabulary),}
            \item{s is the distinguished category (a propositional
                variable),
            }
            \item{$F: \Sigma \rightarrow \wp(Tp)$ is a function which maps
                each symbol of the alphabet into the set if its types.
                If $F(a) = \{A_1, \ldots, A_n \}$ we write $G: a\mapsto A_1, \ldots, A_n$.
            }
        \end{itemize}
    \end{definition}
    For $w \in \Sigma^*, w=a_1 \cdots a_n$, we say that $G$
    {\it accepts} $w$ if there is a proof in Lambek calculus of
    \begin{displaymath}
        A_1,\ldots,A_n \vdash s
    \end{displaymath}
    with $G: a_i \mapsto A_i$ for each $i$.\newline\newline
    The language generated by a Lambek grammar $G$ is
    \begin{displaymath}
        \naming(G)=\set{a_1 \cdots a_n \in {\Sigma}^* \ |\ \mbox{for }1 \leq i \leq n,\
        \exists A_i,\ G: a_i\mapsto A_i\  \mbox{and } \ A_1,\ldots,A_n
        \vdash s}.
    \end{displaymath}

    \begin{example}
       Let $\Sigma=\{${\bf Mary, cooked, the, beans}$\}$ 
       be our alphabet and $s$ our distinguished
       category. Let's take F such that
       \begin{eqnarray*}
           {\bf Mary} &:&  np\\
           {\bf cooked}&:& (np\backslash s)/np \\
           {\bf the} &:& np/n\\
           {\bf beans} &:& n
       \end{eqnarray*}
       Then {\rm Mary cooked the beans} belongs to the
       language generated by this grammar, because in Lambek
       calculus we can prove:
       \begin{displaymath}
           np, (np\backslash s)/np, np/n, n \vdash s
       \end{displaymath}
    \end{example}
    \newpage
    Weak generative capacity for associative Lambek grammars 
    was characterized (see \cite{pentus97})
    by the following celebrated theorem, one of the finest and
    most recent achievements in this field:
    \begin{theorem}[Pentus, 1997]
        \label{PentusTheo}
        The languages generated by associative Lambek grammars
        are exactly the context-free languages.
    \end{theorem}
    Analogously, for non-associative Lambek grammars Buszkowski 
    proved (see \cite{buszkowski86}):
    \begin{theorem}[Buszkowski, 1986]
        \label{BuszkowskyTheo}
        The languages generated by non-associative Lambek
        grammars are exactly the context-free languages.
    \end{theorem}
   
\section{Proofs as Grammatical Structures}
    \label{ptreessec}
    In this  section we will introduce the notion of {\it structure}
    for a sentence generated by a Lambek grammar. On the basis of
    a recent work by Hans-Joerg Tiede (see \cite{tiede99}) who
    proved some important theorems about the tree language
    of proof trees in Lambek calculus, we will adopt as the
    underlying structure of a sentence in a Lambek grammar a
    proof of its well-formedness in Lambek calculus. We will see
    in section \ref{learnlambeksection} how this choice affects the
    process of learning a rigid Lambek grammar on the basis of
    {\it structured positive data}.

    \subsection{(Partial) Parse Trees for Lambek Grammars}
    Just as a derivation encodes a proof of $A_1, \ldots, A_n \vdash B$, the
    notion of {\it parse tree} introduced by the following definition encodes a
    proof of $a_1 \cdots a_n \in \naming(G)$ where $G$ is a Lambek grammar and
    $a_1, \ldots, a_n$ are symbols of its alphabet.

    \begin{definition}
        Let $G = \seq{\Sigma, s, F}$ be a Lambek grammar, then
        \begin{itemize}
            \item if ${\cal D}$ is a derivation of $B$ from $A_1, \ldots, A_n$, and
                $a_1, \ldots, a_n$ are symbols of alphabet $\Sigma$ such that
                $G: a_i \mapsto A_i$ for $1 \leq i \leq n$,
                the result of attaching $a_1, \ldots, a_n$,
                from left to right in this order, to the undischarged leaf nodes of ${\cal D}$
                is a {\rm partial parse tree} of $G$.
                \begin{figure}[htbp]
                    \begin{center}
                        \includegraphics[height=2.7cm]{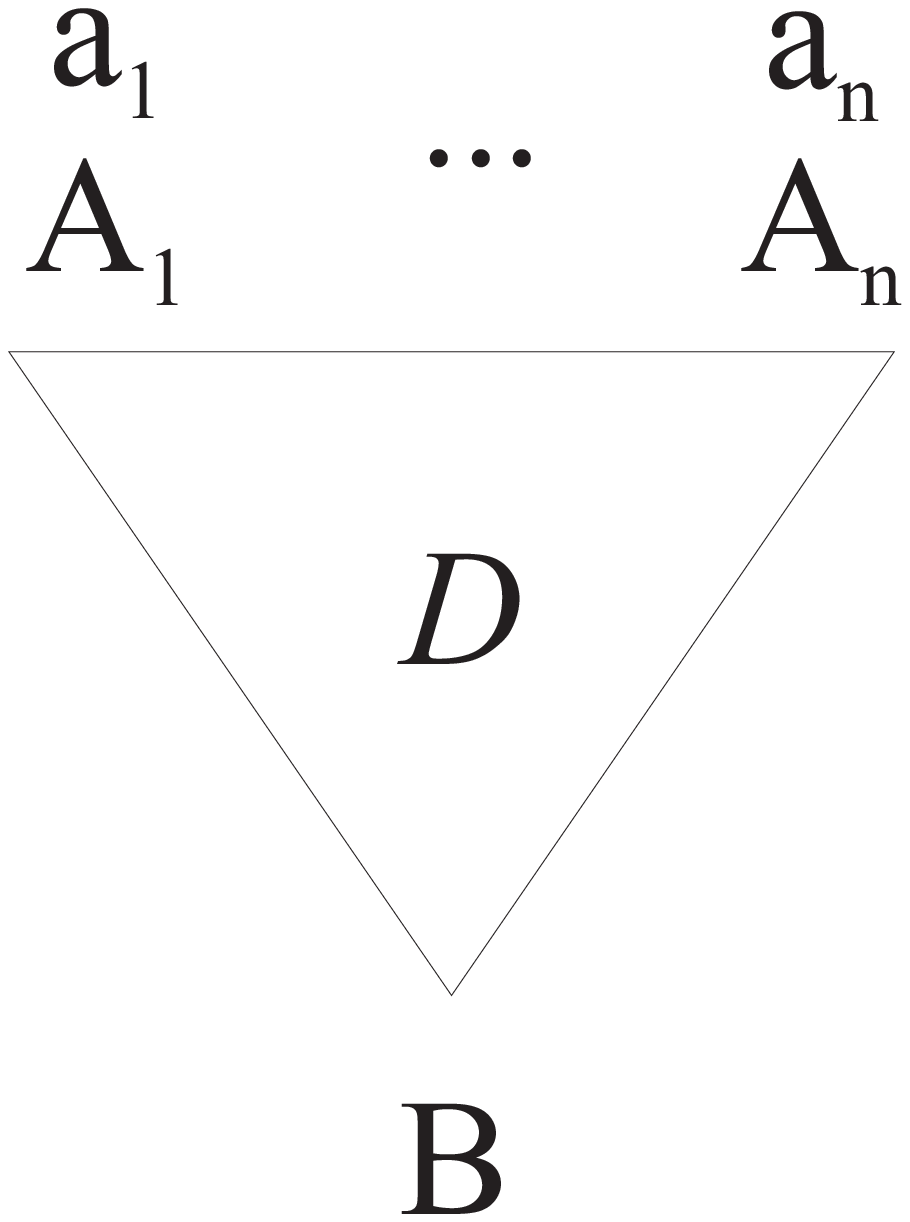}
                    \end{center}
                \end{figure}
            \item A {\rm parse tree} of G is a partial parse tree of G whose root node is
                labeled by the distinguished category $s$.
        \end{itemize}
    \end{definition}
    If $a_1 \cdots a_n$ is the string of symbols attached to the leaf nodes of
    a partial parse tree ${\cal P}$, $a_1 \cdots a_n$ is said to be the {\it yield}
    of ${\cal P}$. If a parse tree ${\cal P}$ of $G$ yields $ a_1 \cdots a_n$, then
    ${\cal P}$ is called a {\it parse} of $a_1 \cdots a_n$ in $G$.
    \begin{example}
        \label{helikeshimex}
        Let $\Sigma=\{{\bf he}, {\bf him}, {\bf likes}\}$ be our alphabet and let $G$ a Lambek
        grammar such that
        \begin{eqnarray*}
                G: {\bf likes}&\mapsto& (np\backslash s)/np,\\
                {\bf he}&\mapsto& s/(np\backslash s),\\
                {\bf him}&\mapsto& (s/np)\backslash s.
        \end{eqnarray*}
        Then the following is a parse for {\bf he likes him}:
        \begin{figure}[htbp]
            \begin{center}
                \includegraphics[height=7cm]{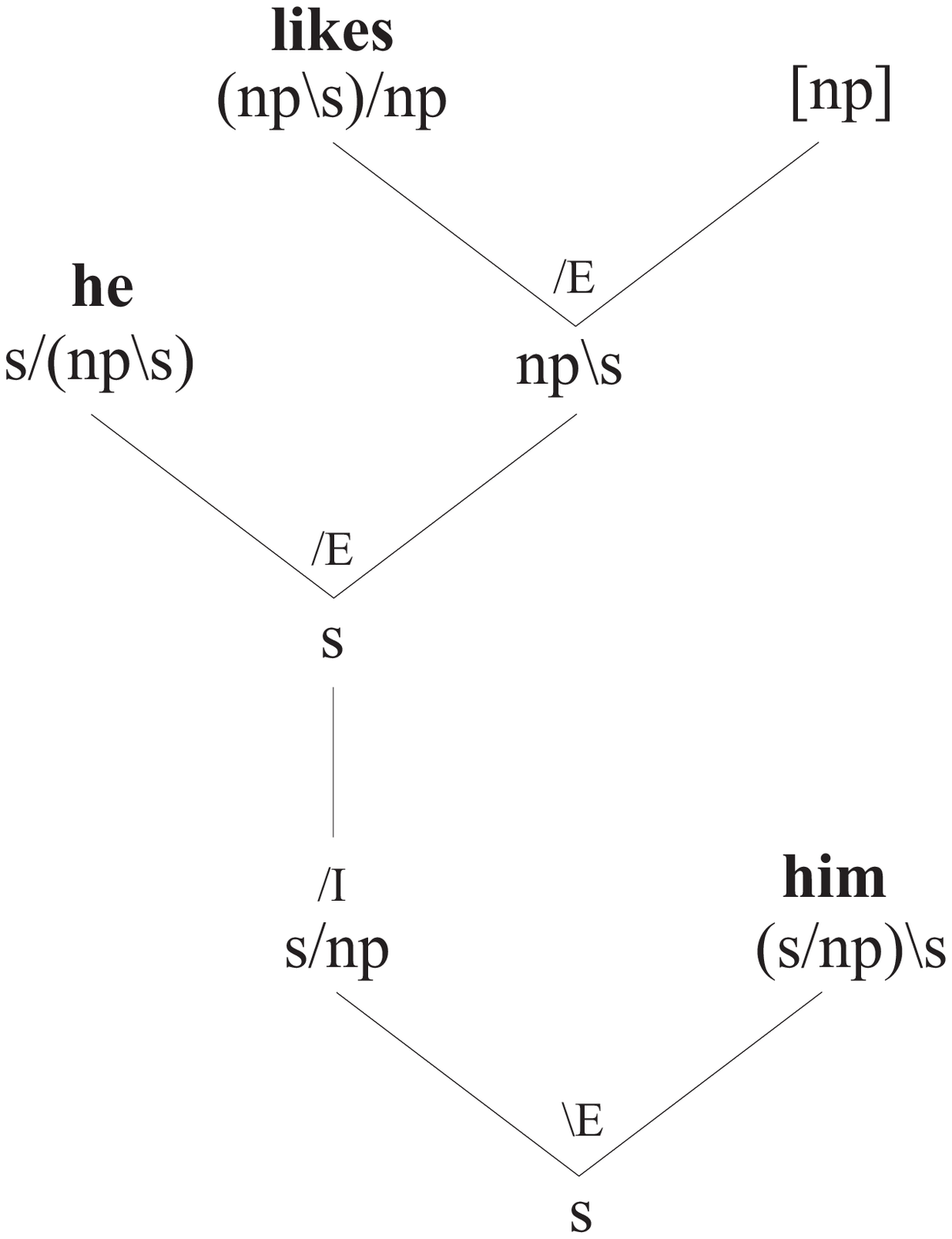}
            \end{center}
        \end{figure}
    \end{example}

    \subsection{Tree Languages and Automata}
    In order to fully appreciate the peculiarity of Lambek grammars with respect to
    their strong generative capacity, we recall here some basic definitions
    about the notion of {\it tree language} as presented in \cite{tiede99}.

    \begin{definition}[Trees and tree languages]
        A {\rm tree} is a term over a finite signature $\Sigma$ containing function and
        constant symbols. The set of n-ary function symbols in $\Sigma$ will be denoted
        by $\Sigma_n$. The set of all terms over $\Sigma$ will be denoted by
        $T_\Sigma$; a subset of $T_\Sigma$ is called a {\rm tree language} or a {\rm forest}.
    \end{definition}
    \begin{definition}[Yield of a tree]
        The {\rm yield} of a tree t is defined by
        \begin{eqnarray*}
            yield(c) &=& c, \mbox{\ \ for } c \in \Sigma_0\\
            yield(f(t_1, \ldots, t_n)) &=& yield(t_1), \ldots, yield(t_n), \mbox{\ \ for } f \in \Sigma_n, n>0
        \end{eqnarray*}
        Thus, the yield of a tree is the string of symbols occurring as its leaves.
    \end{definition}
    \begin{definition}[Root of a tree]
        The {\rm root} of a tree $t$ is defined by
        \begin{eqnarray*}
            root(c) &=& c, \mbox{\ \ for } c \in \Sigma_0\\
            root(f(t_1, \ldots, t_n)) &=& f, \mbox{\ \ for } f \in \Sigma_n, n>0.
        \end{eqnarray*}
    \end{definition}

    In the following subsections three increasingly more powerful 
    classes of tree languages are presented: local, regular and 
    context-free tree languages. Note that even if the names for
    these classes of tree languages are the same as those for 
    classes of string languages, their meaning is very different.

    \subsubsection{Local Tree Languages}
    We can think of a local tree language as a tree 
    language whose membership problem can be decided
    by just looking at some very simple (local) properties 
    of trees. A formalization of such an intuitive notion
    is given by the following definitions:

    \begin{definition}[Fork of a tree]
        The {\rm fork} of a tree t is defined by
        \begin{eqnarray*}
            fork(c) &=& \emptyset, \mbox{\ \ for } c\in \Sigma_0\\
            fork(f(t_1, \ldots, t_n)) &=& \{\langle f, root(t_1), 
            \ldots, root(t_n)\rangle\} \cup \bigcup_{i=1}^{n} fork(t_i)
        \end{eqnarray*}
    \end{definition}
    \begin{definition}[Fork of a tree language]
        For a tree language L, we define
        \begin{displaymath}
            fork(L)= \bigcup_{t \in L} fork(t)
        \end{displaymath}
    \end{definition}
    Note that, since $\Sigma$ is finite, $fork(T_\Sigma)$ is always finite.

    \begin{definition}[Local tree language]
        A tree language $L \subseteq T_\Sigma$ is {\rm local} 
        if there are sets $R \subseteq \Sigma$ and
        $E \subseteq fork(T_\Sigma)$, such that, for all 
        $t \in T_\Sigma$, $t \in L$ iff $root(t) \in R$ and $fork(t) \subseteq E$.
    \end{definition}

    Thatcher (see \cite{thatcher67}) characterized the relation 
    between local tree languages and the derivation trees of
    context-free string grammars by the following
    \begin{theorem}[Thatcher, 1967]
        \label{thatcherth}
        S is the set of derivation trees of some context-free string grammar iff S is local.
    \end{theorem}

    \subsubsection{Regular Tree Languages}
    Among many different equivalent definitions for regular 
    tree languages, we follow Tiede's approach in choosing 
    the following one, based on finite tree automata.
    \begin{definition}[Finite tree automaton]
        A finite tree automaton is a quadruple 
        $\langle \Sigma, Q, q_0, \Delta\rangle$, such that
        \begin{itemize}
            \item $\Sigma$ is a finite signature,
            \item $Q$ is a finite set of unary states,
            \item $q_0 \in Q$ is the start state,
            \item $\Delta$ is a finite set of transition rules of the following type:
                \begin{eqnarray*}
                    q(c) &\rightarrow& c  \mbox{\ \ for } c\in \Sigma_0\\
                    q(f(v_1, \ldots, v_n)) &\rightarrow& f(q_1(v_1), 
                    \ldots, q_n(v_n)) \mbox{\ \ for }
                    f \in \Sigma_n,\ q, q_1, \ldots, q_n \in Q
                \end{eqnarray*}
        \end{itemize}
    \end{definition}
    We can think of a finite tree automaton as a device which scans
    non-deterministically a tree from root to frontier.
    It accepts a tree if it succeeds in reading the whole tree, 
    it rejects it otherwise.

    In order to define the notion of tree language accepted 
    by a regular tree automaton we need
    to define the transition relation for finite tree automata.
    \begin{definition}
        A {\rm context} is a term over $\Sigma \cup \{x\}$ 
        containing the zero-ary term x exactly once.
    \end{definition}
    \begin{definition}
        \label{transitionrel}
        Let $M = \langle \Sigma, Q, q_0, \Delta\rangle$ be a finite tree automaton,
        the derivation relation
        \begin{displaymath}
            \Rightarrow_M \subseteq T_{Q \cup \Sigma} \times T_{Q \cup \Sigma}
        \end{displaymath}
        is defined by $t \Rightarrow_M t^\prime$ if for some 
        context s and some $t_1, \ldots, t_n \in T_\Sigma$,
        there is a rule in $\Delta$
        \begin{displaymath}
            q(f(v_1, \ldots, v_n)) \rightarrow f(q_1(v_1), \ldots, q_n(v_n))
        \end{displaymath}
        and
        \begin{eqnarray*}
            t &=& s[x \mapsto q(f(t_1, \ldots, t_n))]\\
            t^\prime &=& s[x \mapsto f(q_1(t_1), \ldots, q_n(t_n))].
        \end{eqnarray*}
        If we use $\Rightarrow_M^*$ to denote the reflexive, 
        transitive closure of $\Rightarrow_M$, we say that
        a finite automaton M {\rm accepts} a term $t \in T_\Sigma$ 
        if $q_0(t) \Rightarrow_M^* t$.
        The tree language accepted by a finite tree automaton M is
        \begin{displaymath}
            \{t \in T_\Sigma\ |\ q_0(t) \Rightarrow_M^* t \}.
        \end{displaymath}
    \end{definition}
    \begin{definition}[Regular tree language]
        A tree language is {\rm regular} if it is accepted by a finite tree automaton.
    \end{definition}
    The following theorem (see \cite{kozen97})
    defines the relation between local and regular tree languages:
    \begin{theorem}
        Every local tree language is regular.
    \end{theorem}
    while the following (see \cite{gs84})
    establishes a relation between regular tree languages and context-free string languages:
    \begin{theorem}
        The yield of any regular tree language is a context-free string language.
    \end{theorem}

    \subsubsection{Context-free Tree Languages}
    The final step in the definition of more and more 
    powerful tree language classes is made possible
    by introducing the notion of {\it pushdown tree automaton}. 
    Again, we stick to Tiede's approach in
    choosing Guesserian's useful definition (see \cite{guesserian83}):
    \begin{definition}[Pushdown tree automaton]
        A pushdown tree automaton is a system 
        $\langle \Sigma, \Gamma, Q, q_0, Z_0, \Delta\rangle$,
        such that
        \begin{itemize}
            \item $\Sigma$ is a finite signature (the input signature),
            \item $\Gamma$ is a finite signature (the pushdown signature;
                we assume $\Sigma \cap \Gamma = \emptyset$),
            \item Q is a finite set of binary states,
            \item $q_0 \in Q$ is the start state,
            \item $Z_0 \in \Gamma$ is the initial stack symbol,
            \item $\Delta$ is a finite set of rules of the form
                \begin{eqnarray*}
                    q(f(v_1, \ldots, v_n), E(x_1, \ldots, x_m)) &\rightarrow&
                        f(q_1(v_1, \gamma_1), \ldots, q_n(v_n, \gamma_n)),\\
                    q(v, E(x_1, \ldots, x_m)) &\rightarrow& q^\prime(v, \gamma^\prime),\\
                    q(c) &\rightarrow& c
                \end{eqnarray*}
                with
                \begin{itemize}
                    \item $q, q^\prime, q_1, \ldots, q_n \in Q$,
                    \item $c\in \Sigma_0$,
                    \item $f \in \Sigma_n$, $n > 0$,
                    \item $E \in \Gamma_m$,
                    \item $\gamma^\prime, \gamma_1, \ldots, \gamma_n \in T_{\Gamma \cup \{x_1, \ldots, x_m\}}$.
                \end{itemize}
        \end{itemize}
    \end{definition}
    The transition relation for pushdown tree automata $\Rightarrow$
    can be defined straightforwardly as a generalization of definition
    \ref{transitionrel}. A term {\it t} is accepted by a pushdown
    automaton if $q_0(t, Z_0) \Rightarrow^* t$,
    where $\Rightarrow^*$ is the reflexive, transitive closure of $\Rightarrow$.
    \begin{definition}[Context-free tree language]
        The language accepted by a pushdown tree automaton 
        is called a {\rm context-free tree language}.
    \end{definition}
    The relationship between regular and context-free tree 
    languages is exemplified by the following proposition:
    \begin{proposition}
        The intersection of a regular and a context-free tree language is context-free.
    \end{proposition}
    We know that the yield of a regular tree language is a 
    context-free string language: there is a similar connection
    between the class of context-free tree languages and the class 
    of indexed languages, as stated by the following
    \begin{proposition}
        The yield of any context-free tree language is an indexed string language.
    \end{proposition}
    Indexed languages have been proposed as an upper bound of the complexity of
    natural languages, after it was shown that certain phenomena in natural languages
    cannot be described with context-free grammars (see \cite{gazdar88}).

    \subsection{Proof Trees as Structures for Lambek Grammars}
    \label{rulesec}
    In \cite{tiede99} Hans-Joerg Tiede proposes, in contrast with a previous
    approach by Buszkowski, to take as the structure
    underlying a sentence generated by a Lambek grammar, one of
    the infinite proof trees
    of the deduction $A_1, \ldots, A_n \vdash s$, where $A_1, \ldots, A_n$
    is a sequence of types assigned by the grammar to each symbol,
    and $s$ is the distinguished atomic category.

    Following Tiede's approach, we give the following
    \begin{definition}[Proof tree]
    A {\rm proof tree} for a Lambek grammar is a term over the
    signature $\Sigma = \set{[/E], [\backslash E], [/I], [\backslash I], [ID]}$ where
        \begin{itemize}
            \item {$[ID]$ is the 0-ary function symbol},
            \item {$[/E]$ and $[\backslash E]$ are the binary function symbols},
            \item {$[/I]$ and $[\backslash I]$ are the unary function symbols}.
        \end{itemize}
    \end{definition}
    The terms over this signature represent proof trees that neither have information
    about the formulas for which they are a proof, nor about the strings that are
    generated by a grammar using this proof.
    These terms represent proofs unambiguously, since the assumption
    discharged by an introduction rule is univocally
    determined by the position of the corresponding
    $[/I]$ or $[\backslash I]$ function symbol in the proof tree.

    \begin{example}
        The term $t = [\backslash I]([/E]([ID],[ID]))$ is an 
        example of well-formed term over this signature.
        There's no need for additional information about the discharged assumption since, as
        we can see from the tree-like representation of the term, the discharged
        assumption is unambiguously identified.\newline
        \begin{figure}[htbp]
            \begin{center}
                \includegraphics[height=3.5cm]{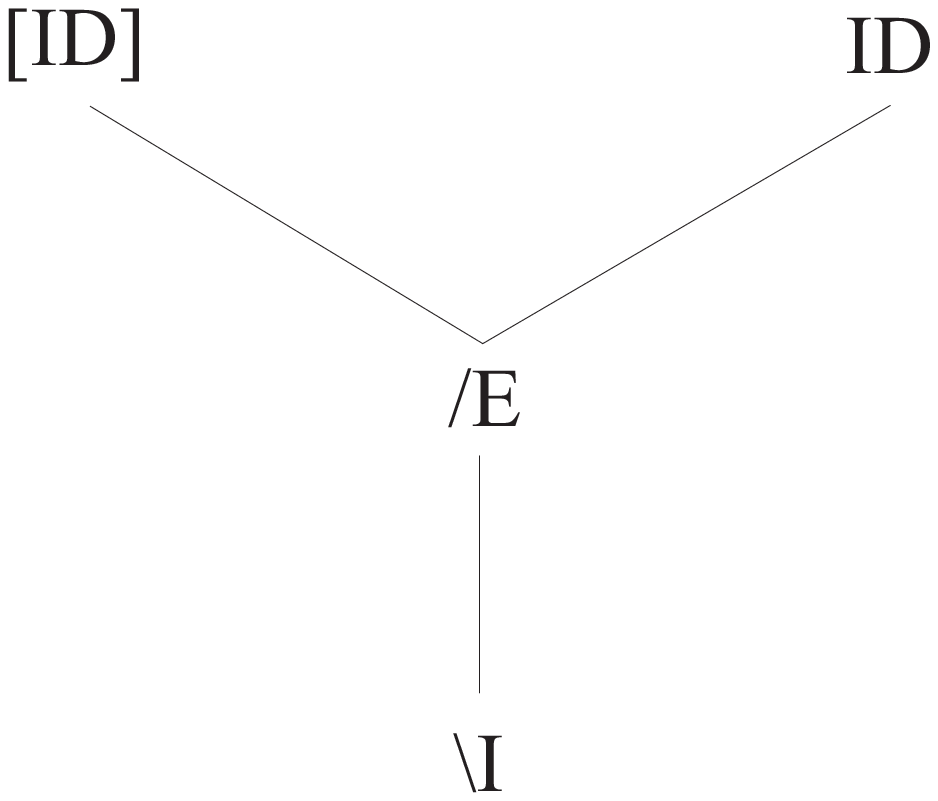}
            \end{center}
        \end{figure}
    \end{example}

    The following terms are examples of not well-formed proof trees for the tree
    language generated by any Lambek grammar:
    \begin{itemize}
        \item {$[\backslash E](x, [/I](y))$. Since the major premise of the $\backslash E$
        function symbol is something with a $(\ldots)\backslash(\ldots)$ shape,
        there's no way to reduct that term by a $\backslash E$ rule;}
        \item {$[/E[([\backslash I](x),y)$. Analogous to the previous situation;}
        \item {$[\backslash I]([ \backslash E](x, [ID]))$ if the term x does not contain at
            least two uncancelled assumptions;}
        \item {$[/I]([/E]([ID], x))$, if the term x does not contain at least two uncancelled assumptions.}
    \end{itemize}

    By taking a proof tree as the structure of a sentences generated by
    Lambek grammars, Tiede proved some important results about their
    strong generative capacity, that is, the set of the structures assigned
    by a grammar to the sentences it generates. Since strong generative capacity
    can provide a formal notion of the linguistic concept of {\it structure} of
    a sentence, this result justifies the current interest toward Lambek
    Grammars as a promising mathematical tool for linguistic purposes.

    \begin{theorem}[Tiede, 1999]
        \label{tiedeth}
        The set of well-formed proof trees of the Lambek Calculus is not
        regular.
    \end{theorem}
    \begin{theorem}[Tiede, 1999]
        \label{tiedeth2}
        The set of proof trees of the Lambek Calculus is a context-free tree language.
    \end{theorem}
    These two theorems show that the language of
    proof trees is properly a context-free tree language.

    In particular, these theorems show that Lambek grammars are more
    powerful, with respect to strong generative capacity,
    than context-free grammars, whose structure language
    is a local tree language as shown in theorem \ref{thatcherth}.\newline

    We can easily introduce the notion of {\it normal form proof tree} by simply extending
    the notion of normal form proof as presented in definition \ref{normalformdef}.
    We can say that for normal form trees,
    in addition to the rules that prohibit terms of the form
    \begin{eqnarray*}
        [\backslash E](x, [/I](y)),\\
        {[/E]([\backslash I](x), y),}
    \end{eqnarray*}
    we have rules that prohibit terms of the form
    \begin{eqnarray*}
        [\backslash E](x, [\backslash I](y)) \\
        {[/E]([/I](x),y)} \\
    \end{eqnarray*}
    and terms of the form
    \begin{eqnarray*}
        {[/I]([/E](x, [ID]))} \\
        {[\backslash I]([\backslash E]([ID],y))}
    \end{eqnarray*}
    which correspond to $\beta$-redexes and $\eta$-redexes,
    respectively, as one can easily see from definition
    \ref{normalformdef}.

    We can easily extend to the formalism of proof trees the
    ``reduction rules'' we've seen in section \ref{normsec} to get a
    normal form proof tree out of a non-normal one.
    \begin{figure}[htbp]
        \begin{center}
            \includegraphics[height=3.5cm]{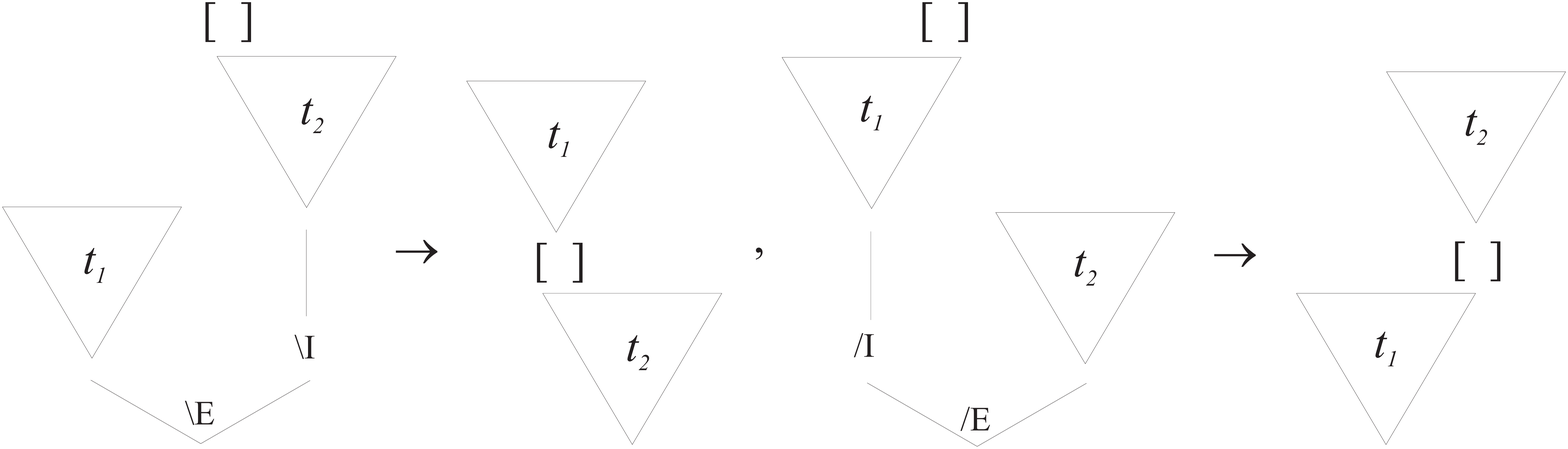}
        \end{center}
    \end{figure}\newpage
    \begin{figure}[htbp]
        \begin{center}
            \includegraphics[height=3cm]{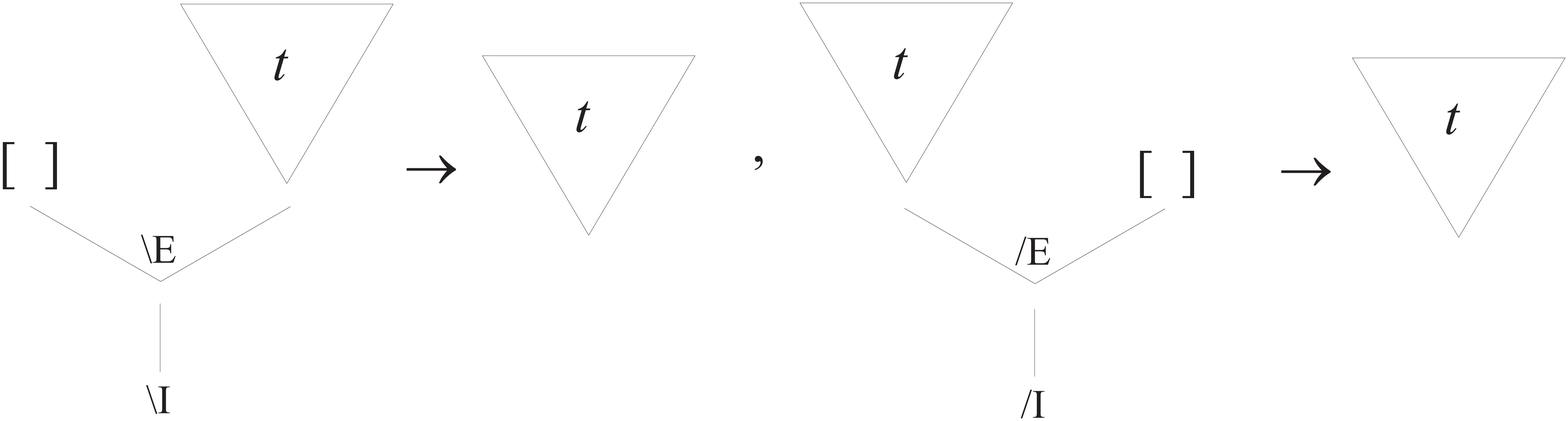}
        \end{center}
    \end{figure}

    As a corollary of theorem \ref{tiedeth}, Tiede proves that
    \begin{theorem}[Tiede, 1999]
        The set of normal form proof trees of the Lambek Calculus is not
        regular,
    \end{theorem}
    which, together with
    \begin{theorem}
        The set of normal form proofs of the Lambek Calculus is a context-free
        tree language
    \end{theorem}
    shows that the tree language of normal form proof trees of Lambek Calculus
    is properly a context-free tree language.\newline

    \subsection{Proof-tree Structures}
    \label{ptsec}
    Given a Lambek grammar $G$, a {\it proof-tree structure}
    over its alphabet $\Sigma$ is a unary-binary branching
    tree whose leaf nodes are labeled by
    either $[ID]$ (these are called "discharged leaf nodes")
    or symbols of $\Sigma$ and whose
    internal nodes are labeled by either $\backslash E, /E,
    \backslash I$, or $/I$.

    The set of proof-tree structures over $\Sigma$ is denoted
    $\Sigma^P$. Often we will simply say `structure' to mean proof-tree
    structure. A set of proof-tree structures over $\Sigma$ is
    called a {\it structure language} over $\Sigma$.

    \begin{example}
        The following is an example of a proof-tree structure for
        the sentence {\it he likes him} seen in example
        \ref{helikeshimex}:
        \begin{figure}[htbp]
            \begin{center}
                \includegraphics[height=6cm]{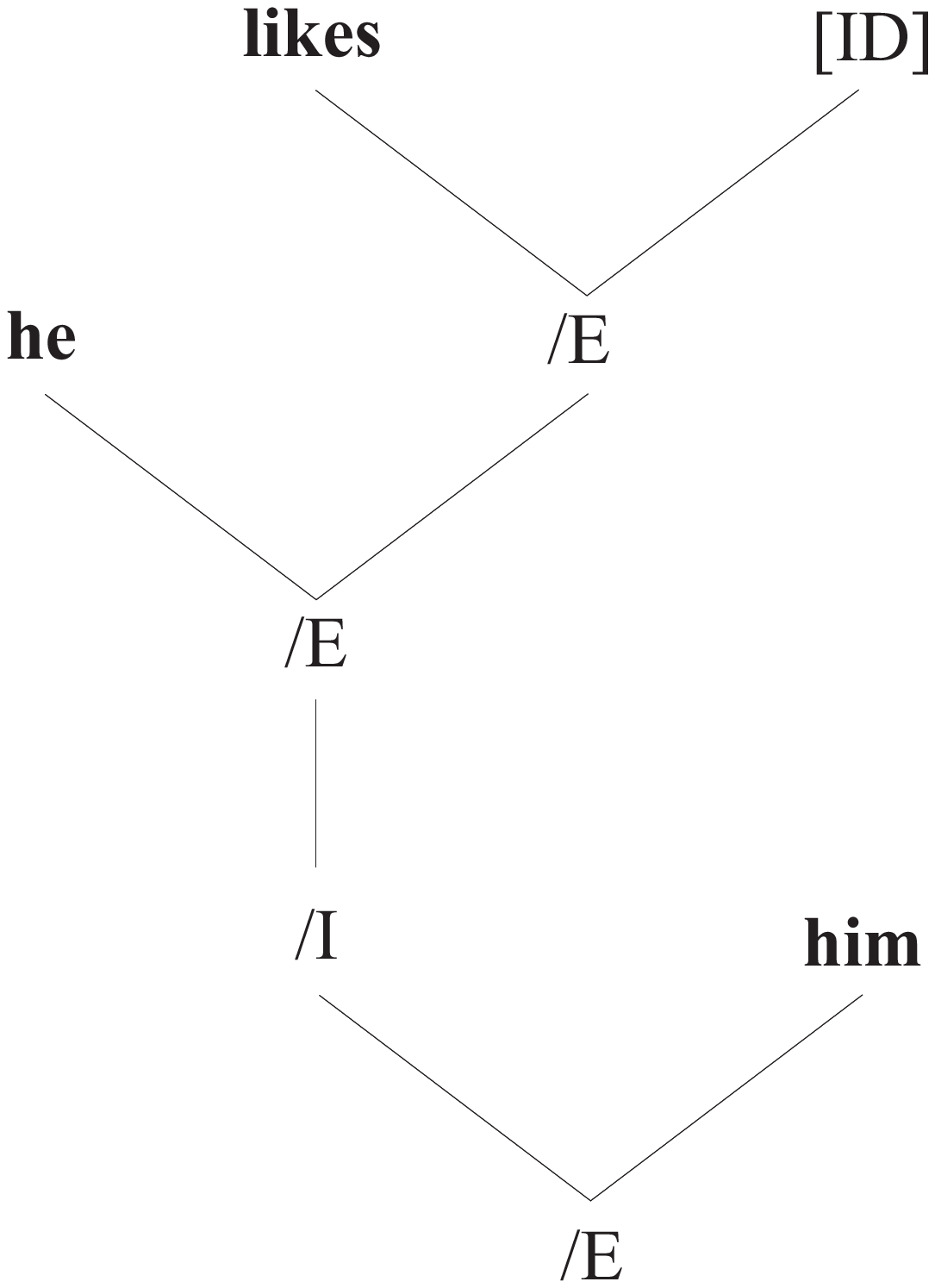}
            \end{center}
        \end{figure}
    \end{example}\newpage

    Let $G$ be a Lambek grammar, and let ${\cal P}$ be a partial
    parse tree of $G$. The result of stripping ${\cal P}$ of its
    type labels is a proof-tree structure, that is called the
    proof-tree structure of ${\cal P}$. If $T$ is the structure of
    a parse tree ${\cal P}$, we say that ${\cal P}$ is a {\it
    parse} of $T$.

    We say that a Lambek grammar $G$ {\it generates} a structure
    $T$ if and only if for some parse tree
    ${\cal P}$ of $G$, $T$ is the structure of ${\cal P}$.
    The set of structures generated by $G$ is called the {\it (proof-tree)
    structure language} of $G$ and is denoted $\pl(G)$. In order to
    distinguish $\naming(G)$, the language of $G$, from $\pl(G)$, its
    structure language, we often call the former the {\it string
    language} of $G$.

    The {\it yield} of a proof-tree structure $T$ is the string of
    symbols $a_1, \ldots, a_n$ labeling the {\it undischarged} leaf
    nodes of $T$, from left to right in this order. The yield of
    $T$ is denoted $yield(T)$. Note that $\naming(G)=\set{yield(T)\ |\ T \in \pl(G)}$.

    \subsection{Decidable and Undecidable Problems about \\ Lambek
    Grammars}

    Since, as stated in by theorem \ref{derivabilityprop}, Lambek
    calculus is decidable, the universal membership problem ``$s\in
    \naming(G)$'' is decidable for any sentence $s$ and any Lambek
    grammar $G$.

    On the other hand, the questions ``$\naming(G_1)=\naming(G_2)$''
    and ``$\naming(G_1) \subseteq \naming(G_2)$'' for arbitrary
    Lambek grammars $G_1$ and $G_2$ are undecidable, because the
    same questions are undecidable for context-free grammars and
    there exists an effective procedure for converting a
    context-free grammar $G^\prime$ to a Lambek grammar $G$ such
    that $L(G^\prime)=L(G)$.

    Given a proof-tree structure $t$ the question ``$t \in \pl(G)$''
    is decidable. In fact, as shown by Tiede in \ref{tiedeth2},
    every proof tree language of a Lambek Grammar is a
    context-free tree language; and that problem is decidable for
    context-free tree languages (you just have to run the pushdown
    tree automata on $t$).

    Unfortunately, the question ``$\pl(G_1)\subseteq
    \pl(G_2)$'' has been proved decidable only for $G_1, G_2$
    non-associative Lambek grammars. Whether it is decidable or
    not for (associative) Lambek grammars is still an open question
    and the subject of active research in this field.

     \subsection{Substitutions}
    In this section we introduce the notion of a
    Lambek grammar being a {\it substitution instance} of another.
    Besides, we define a notion of {\it size} of a Lambek
    grammar that will be decisive in our proof of learnability for
    Rigid Lambek Grammars presented in section \ref{algorithm}.

    First of all, let's define what we mean when we say that a
    Lambek grammar is subset of another one:
    \begin{definition}
        Let $G_1$, $G_2$ be Lambek grammars; we say that $G_1
        \subseteq G_2$ if and only if for any $a \in \Sigma$ such
        that $G_1: a \mapsto A$ we have also $G_2: a \mapsto A$.
    \end{definition}

    \begin{example}
        Let $\set{{\bf Francesca, loves, Paolo}}\subseteq\Sigma$ and let
        \begin{eqnarray*}
            G_1 : \ \ \ {\bf Francesca} &\mapsto& np\\
            {\bf loves} &\mapsto& np\backslash s\\
            G_2: \ \ \ {\bf Francesca} &\mapsto& np\\
                {\bf loves} &\mapsto& np\backslash s, np\backslash(s/np)\\
                {\bf Paolo} &\mapsto& np\\
        \end{eqnarray*}
        Obviously, $G_1 \subseteq G_2$
    \end{example}

    \begin{definition}
        A {\it substitution}\index{substitution} is a function
        $\sigma: Var \rightarrow Tp$ that maps variables to types.
        We can extend it to a function from types to types by setting
        \begin{eqnarray*}
            \sigma(t) &=& t\\
            \sigma(A/B) &=& \sigma(A)/\sigma(B)\\
            \sigma(A\backslash B) &=& \sigma(A)\backslash\sigma(B)
        \end{eqnarray*}
        for all $A, B \in Tp$.\newline
    \end{definition}

    We use the notation $\set{x_1 \mapsto A_1, \ldots, x_n \mapsto
    A_n}$ to denote the substitution $\sigma$ such that
    $\sigma(x_1)=A_1, \ldots, \sigma(x_n)=A_n$ and $\sigma(y)=y$
    for all other variables $y$.

     \begin{example}
        Let $\sigma = \set{x \mapsto x\backslash y, y \mapsto s, z
        \mapsto s/(s/x)}$. Then
        \begin{displaymath}
            \sigma((s/x)\backslash y)=(s/(x\backslash y))\backslash t
        \end{displaymath}
        and
        \begin{displaymath}
            \sigma(((s/x)\backslash y)/(x/z))=((s/(x\backslash y))\backslash
            s)/((x\backslash y)/(s/(s/x))).
        \end{displaymath}
    \end{example}
    \
    \newline
    The following definition introduce the notion of a Lambek
    grammar being a {\it substitution instance} of another:
    \begin{definition}
        Let $G=\langle \Sigma, s, F\rangle$ be a Lambek grammar, and $\sigma$
        a substitution. Then $\sigma[G]$ denotes the grammar obtained by applying
        $\sigma$ in the type assignment of $G$, that is:
        \begin{displaymath}
            \sigma[G] = \seq{\Sigma, s, \sigma\cdot F}
        \end{displaymath}
        $\sigma[G]$ is called a {\rm substitution instance} of G.
    \end{definition}

    It easy to prove also for Lambek grammars this straightforward
    but important fact that was first proved for CCGs in \cite{bp90}
    \begin{proposition}
        \label{bonatoprop}
        If $\sigma[G_1] \subseteq G_2$, then the set of proof-tree
        structures generated by $G_1$ is a subset of the set of
        proof-tree structures generated by $G_2$, that is
        $\pl(G_1) \subseteq \pl(G_2)$.
    \end{proposition}
    {\it Proof.} Suppose $\sigma[G_1]\subseteq G_2$. Let $T \in
    \pl(G_1)$ and let ${\cal P}$ be a parse of $T$ in $G_1$. Let
    $\sigma[{\cal P}]$ the result of replacing each type label $A$
    of ${\cal P}$ by $\sigma(A)$. Then it is easy to see that
    $\sigma[{\cal P}]$ is a parse of $T$ in $G_2$. Therefore, $T
    \in \pl(G_2)$.
   
    \begin{corollary}
        If $\sigma[G_1] \subseteq G_2$, then $\naming(G_1) \subseteq \naming(G_2)$.
    \end{corollary}
    {\it Proof.} Immediate from the previous proposition and the
    remark at the end of section \ref{ptsec}.\newline

    A substitution that is a one-to-one function from $Var$ to
    $Var$ is called a {\it variable renaming}. If $\sigma$ is a
    variable renaming, then $G$ and $\sigma[G]$ are called {\it
    alphabetic variants}. Obviously grammars that are alphabetic
    variants have exactly the same shape and are identical for all
    purposes. Therefore, grammars that are alphabetic variants are
    treated as identical.

    \begin{proposition}
        \label{alphaprop}
        Suppose $\sigma_1[G_1]=G_2$ and $\sigma_2[G_2]=G_1$. Then
        $G_1$ and $G_2$ are alphabetic variants and thus are
        equal.
    \end{proposition}
    {\it Proof.} For each symbol $c \in \Sigma$, $\sigma_1$ and
    $\sigma_2$ provide a one-to-one correspondence between
    $\set{A\ |\ G_1: c \mapsto A}$ and $\set{A\ |\ G_2: c \mapsto A}$.
    Indeed, if it didn't and, say, $\set{\sigma_1(A)\ |\ G_1: c \mapsto A}
    \subset \set{A\ |\ G_2: c \mapsto A}$, then
    $\sigma_2[G_2]=\sigma_2[\sigma_1[G_1]]$ couldn't be equal to
    $G_1$, and likewise for $\sigma_2$. Then, it is easy to see
    that $\sigma_1 \uparrow Var(G_1)$ is a one-to-one function
    from $Var(G_1)$ onto $Var(G_2)$, and $\sigma_2 \uparrow
    Var(G_2)=(\sigma_1 \uparrow Var(G_1))^{-1}$. One can extend
    $\sigma_1 \uparrow Var(G_1)$ to a variable renaming $\sigma$.
    Then $\sigma[G_1]=\sigma_1[G_1]=G_2$.

    \subsection{Grammars in Reduced Form}
    \begin{definition}
        A substitution $\sigma$ is said to be {\rm faithful to} a
        grammar $G$ if the following condition holds:
        \begin{quote}
            for all $c \in dom(G)$, if $G_1: c \mapsto A$, $G_1: c
            \mapsto B$, and $A \not = B$, then $\sigma(A) \not =
            \sigma(B)$.
        \end{quote}
    \end{definition}

    \begin{example}
        Let $G$ be the following grammar
        \begin{eqnarray*}
            G : \ \ \ {\bf Francesca} &\mapsto& x,\\
                {\bf dances} &\mapsto& x\backslash s, y\\
                {\bf well} &\mapsto& y\backslash(x\backslash s).
        \end{eqnarray*}
        Let
        \begin{eqnarray*}
            \sigma_1 &=& \set{y \mapsto x},\\
            \sigma_2 &=& \set{y\mapsto x\backslash s}.
        \end{eqnarray*}
        Then $\sigma_1$ is faithful to $G$, while $\sigma_2$ is
        not.
    \end{example}

    \begin{definition}
        Let $\sqsubseteq$ be a binary relation on grammars such
        that $G_1 \sqsubseteq G_2$ if and only if there exists a
        substitution $\sigma$ with the following properties:
        \renewcommand{\labelenumi}{(\alph{enumi})}
        \begin{itemize}
            \item $\sigma$ is faithful to $G_1$;
            \item $\sigma[G_1] \subseteq G_2$.
        \end{itemize}
    \end{definition}

    From the definition above and proposition \ref{alphaprop} it's
    immediate to prove the following:
    \begin{proposition}
        $\sqsubseteq$ is reflexive, transitive and antisymmetric.
    \end{proposition}

    \begin{definition}
        For any grammar $G$, define the {\rm size} of $G$,
        $size(G)$, as follows:
        \begin{displaymath}
            size(G)= \sum_{c \in \Sigma}\sum_{G: c\mapsto A} |A|,
        \end{displaymath}
        where, for each type $A$, $|A|$ is the number of symbol
        occurrences in $A$.
    \end{definition}

    \begin{lemma}
        \label{LemmaKan1}
        If $G_1 \sqsubseteq G_2$, then $size(G_1)\le size(G_2)$,
    \end{lemma}
    {\it Proof}. For any type $A$ and any substitution $\sigma$,
    $|A| \le |\sigma(A)|$. Then the lemma is immediate from the
    definition of $\sqsubseteq$.

    \begin{corollary}
        \label{sizecorollary}
        For any grammar $G$, the set $\{G^\prime\ |\ G^\prime
        \sqsubseteq G\}$ is finite.
    \end{corollary}
    {\it Proof}. By lemma \ref{LemmaKan1}, $\{G^\prime\ |\ G^\prime
        \sqsubseteq G\} \subseteq \{G^\prime\ |\ size(G^\prime)\le
        size(G)\}$. The latter set must be finite, because for any
        $n \in \naturals$, there are only finitely many grammars
        $G$ such that $size(G)=n$.\newline

    If we write $G_1 \sqsubset G_2$ to mean $G_1 \sqsubseteq G_2$
    and $G_1 \not = G_2$, we have
    \begin{corollary}
        $\sqsubset$ is well-founded.
    \end{corollary}

    \begin{definition}
        A grammar $G$ is said to be in {\it reduced form} if there
        is no $G^\prime$ such that $G^\prime\sqsubset G$ and
        $\pl(G)=\pl(G^\prime)$.
    \end{definition}
    \newpage

    \section{Lambek Grammars as a Linguistic Tool}

    \subsection{Lambek Grammars and Syntax}
    As explicitly stated in the original paper wherein Lambek laid the foundations
    of the Lambek Calculus, his aim was
    \begin{quote}
        [...] to obtain an effective rule (or algorithm) for distinguishing sentences from
        nonsentences, which works not only for the formal languages of
        interest to the mathematical logician, but also for natural languages
        such as English, or at least for fragments of such languages.
        (\cite{lambek58})
    \end{quote}

    That's why, even if Lambek grammars can be simply considered as
    interesting mathematical objects,
    it will be useful to underline here some
    properties that make them also an
    interesting tool to formalize some
    phenomena in natural languages.

    The importance of Lambek's approach to grammatical reasoning lies in
    the development of a uniform {\it deductive} account of the composition
    of form and meaning in natural language: formal grammar is presented
    as a {\it logic}, that is a system to reason about structured linguistic structures.

    The basic idea underlying the notion of Categorial Grammar on which Lambek
    based his approach is that a grammar is a formal device to assign to each
    word (a symbol of the alphabet of the grammar) or expression (an ordered
    sequence of words) one or more {\it syntactic types} that describe their
    function. Types can be considered as a formalization of the linguistic
    notion of {\it parts of speech}.

    CCGs assign to each symbol a fixed set of types, and provide
    two composition rules to derive the type of a sequence of words out of
    the types of its components. Such a ``fixed types'' approach leads to some
    difficulties: to formalize some linguistic phenomena we should add
    further rules to the two elimination rules defined for CCGs as
    described in section \ref{extensions}.
    In the following subsections we present some examples
    where the deductive approach of Lambek grammars leads to more an
    elegant and consistent formalization of such linguistic phenomena.

    In the following subsections we take $s$ as the primitive type of
    {\it well-formed sentences} in our language and $np$ as the primitive type for
    {\it noun phrases} (such as John, Mary, he).

    \subsubsection{Transitive verbs}
    Transitive verbs require a name both on their left and right hand sides,
    as it is apparent from the well-formedness of the following sentences.
    \begin{eqnarray*}
        \stackrel{np}{\mbox{John}} (\stackrel{(np\backslash s)/np}{\mbox{\ \ \ likes\ \ \ }} \stackrel{np}{\mbox{Mary}})\\
        (\stackrel{np}{\mbox{John}} \stackrel{np\backslash(s/np)}{\mbox{\ \ \ likes\ \ \ }}) \stackrel{np}{\mbox{Mary}}
    \end{eqnarray*}
    Both parenthesizations lead to a derivation of $s$ as type of the whole expression.
    This would mean that in an CCG we should assign to any transitive verb
    at least two distinct types: $(np\backslash s)/np$ and $np\backslash(s/np)$.

    On the contrary, in a Lambek grammar, since we can prove both
    \begin{displaymath}
        (np\backslash s)/np \vdash np\backslash(s/np)\\
    \end{displaymath}
    and
    \begin{displaymath}
        np\backslash(s/np) \vdash (np\backslash s)/np
    \end{displaymath}
    we can simply assign to a transitive verb the type $np\backslash s/np$ without any
    further parenthesizations.

    \subsubsection{Pronouns}
    If we try to assign a proper type to the personal pronoun {\it he} we notice that
    its type is such that the following sentences are well-formed:\newline
    \centerline{${\mbox{he}} \stackrel{np\backslash s}{\mbox{works}}$,} \\ \\
    \centerline{${\mbox{he}} \stackrel{np\backslash s/np}{\mbox{\ \ likes\ \ }} \stackrel{np}{\mbox{Jane}}$}\\ \\
    We have two choices: either we give {\it he} the same 
    type as a name (that is, $np$) or we
    give it the type $s/(np\backslash s)$. 
    In the first case there is a problem: expressions like
    {\it Jane likes he} are considered as well-formed sentences. 
    So, we assign to {\it he} the type $s/(np\backslash s)$.

    Analogously, since the personal pronoun {\it him} makes 
    the following sentences     well-formed:\newline
    \centerline{$\stackrel{np}{\mbox{Jane}} 
    \stackrel{np\backslash s/np}{\mbox{\ \ likes\ \ }} \mbox{him}$}\\ \\
    \centerline{$\stackrel{np}{\mbox{Jane}} \stackrel{np\backslash s}{\mbox{\ \ works\ \ }}
            \stackrel{s\backslash s/np}{\mbox{for}} \mbox{him}$,}\\ \\
    we assign to {\it him} the type $(s/np)\backslash s$ (and not type $np$, since expressions
    like {\it him likes John} would be well-formed).

    Since a pronoun is, according to its own
    definition, something that ``stands for a noun'',
    we wish that in our grammar each occurrence
    of a pronoun could be replaced by a name
    (while the converse is not always true): but
    this means that any name (say, John, of type $np$)
    should also be assigned the type of {\it he} and {\it him}, that is, respectively,
    type $s/(np\backslash s)$ and type $(s/np)\backslash s$. In other words,
    we need something that accounts for a type-raising. 
    But since in Lambek Calculus we can prove
    \begin{eqnarray*}
        np \vdash s/(np\backslash s)\\
        np \vdash (s/np)\backslash s
    \end{eqnarray*}
    for any $np$ and $s$, a  Lambek grammar provides a very natural
    formalization of the relationship between names and pronouns: while a name
    can always be substituted to a pronoun in a sentence (and the type-raising derivation
    guarantees that a name can always ``behave like'' a pronoun if we need it to), 
    the converse is not
    true (the converse of the type-raising proof doesn't hold in Lambek Calculus).
    The proof of the first deduction is reported in example \ref{type-raising1}
    as a derivation in a Lambek grammar.

    \subsubsection{Adverbs}
    If we look for the proper type for adverbs like {\it here} we can consider the well-formed
    sentence {\it John works here}. We can choose between two possible parenthesizations
    here, that is: \newline
    \centerline{$(\stackrel{np}{\mbox{John\ }} \stackrel{np\backslash s}{\mbox{works\ }}) \mbox{\ here}$}\\ \\
    \centerline{$\stackrel{np}{\mbox{John}} (\stackrel{np\backslash s}{\mbox{works}} \mbox{here})$}\\ \\
    The first one suggests for {\it here} the type $s\backslash s$, while the second one the type
    $(np\backslash s)\backslash(np\backslash s)$. The good news is that, while in a CCG we
    should assign each adverb at least two different types, in a Lambek grammar we can prove that
    \begin{displaymath}
        s \backslash s \vdash (np\backslash s) \backslash (np\backslash s)
    \end{displaymath}
    that is to say, in Lambek grammars any adverbial expression of type $s\backslash s$ has also
    type $(np\backslash s)\backslash(np\backslash s)$. More generally, we can show that in Lambek Calculus
    \begin{eqnarray*}
        x\backslash y \vdash (z\backslash x)\backslash(y\backslash x)\\
        x/y \vdash (x/z)/(y/z).
    \end{eqnarray*}

    \subsubsection{Hypothetical reasoning}
    In the following example, sentences $s$, noun phrases $np$, common nouns $n$,
    and propositions phrases $pp$ are taken to be ``complete expressions'', whereas
    the verb {\it dances}, the determiner {\it the}
    and the preposition {\it with} are categorized as
    incomplete with respect to these complete phrases. \newpage
    \begin{example}
        Here is the derivation for the sentence {\rm Francesca dances with the boy}.
        \begin{figure}[htbp]
            \begin{center}
                \includegraphics[height=6cm]{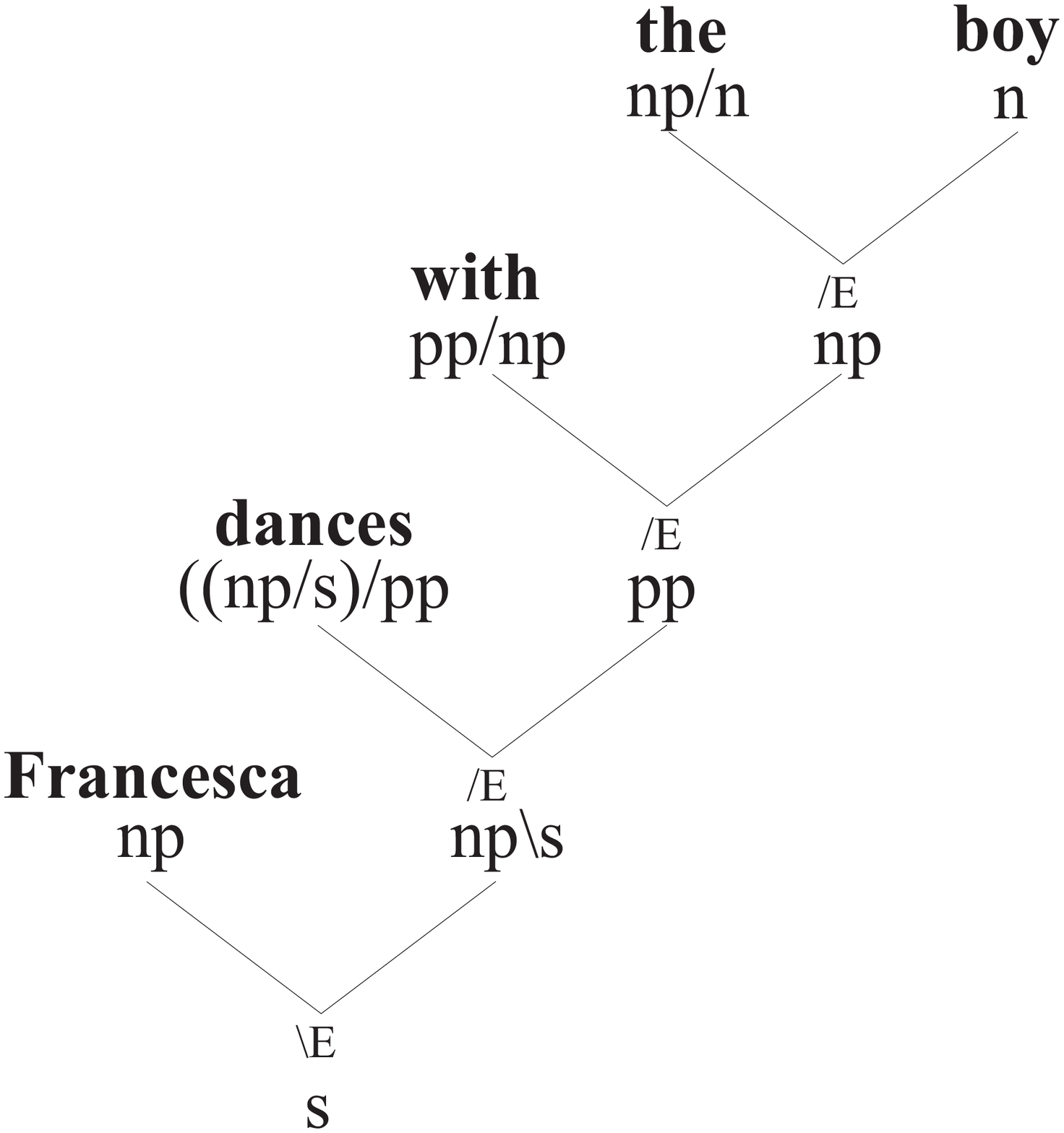}
            \end{center}
        \end{figure}\newline
    \end{example}
        This is an example of grammatical reasoning where, on the basis of
        the types we assigned to each word, we infer the well-formedness of a
        sequence of words. On the other hand we can assume a different perspective:
        knowing that a sentence is well-formed, what can be said about the type
        of its components? In the words of Lambek: ``Given the information about
        the categorization of a composite structure, what conclusions could be
        draw about the categorization of its parts?'' (\cite{lambek58}). That's
        where the following inference patterns come into play:
        \begin{eqnarray*}
            &&\mbox{from\ } \Gamma, B \vdash A, \ \ \ \ \ \mbox{infer\ } \Gamma \vdash A/B,\\
            &&\mbox{from\ } B, \Gamma \vdash A, \ \ \ \ \ \mbox{infer\ } \Gamma \vdash B\backslash A
        \end{eqnarray*}
        which gives a linguistic interpretation of the role 
        of the ``introduction'' rules. 
        That's what is done in the following
        derivation which allows us to infer that 
        the expression {\it the boy Francesca dances with}
        is of type $np$:\newpage
        \begin{figure}[htbp]
            \begin{center}
                \includegraphics[height=9cm]{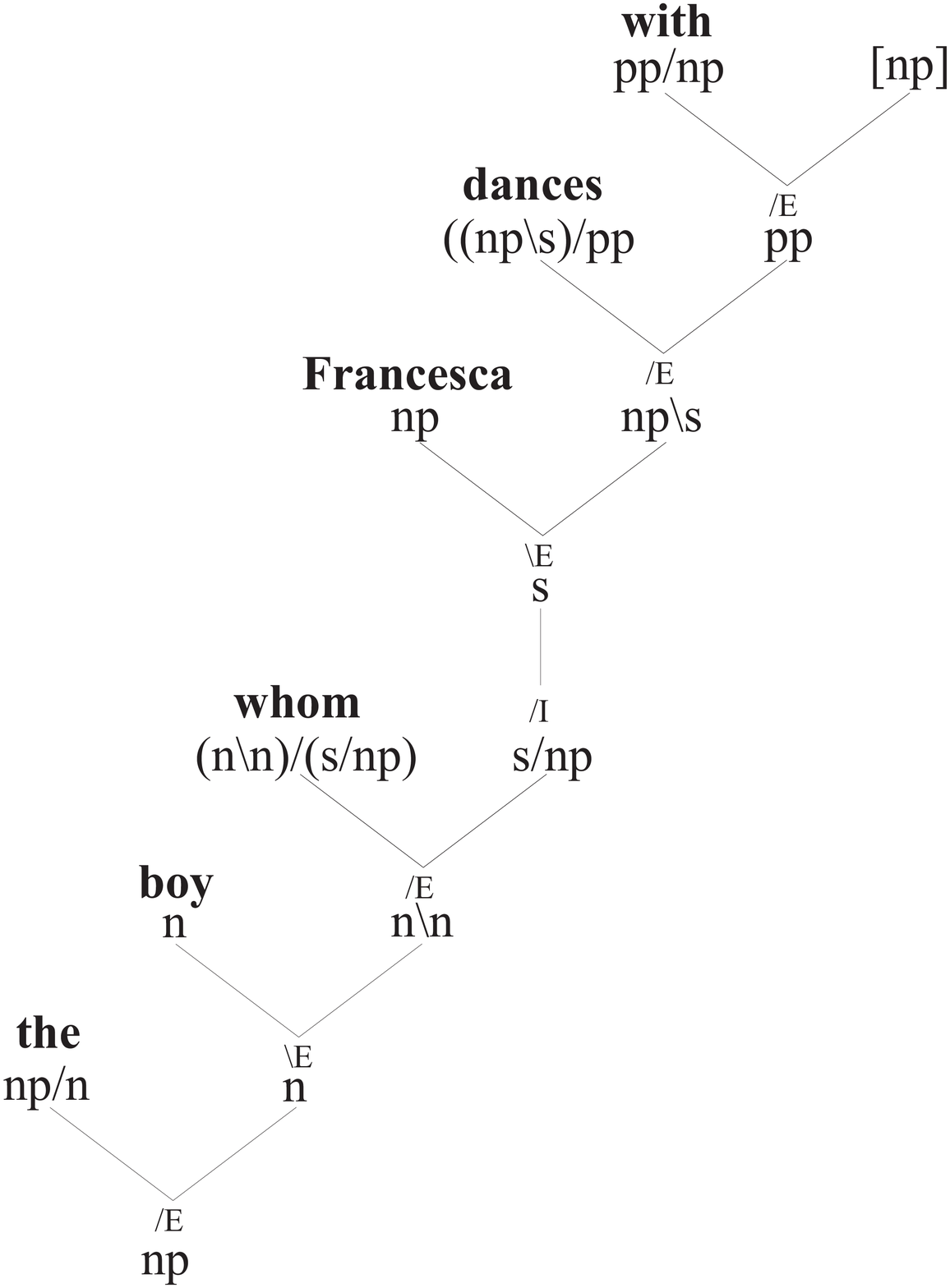}
            \end{center}
        \end{figure}

        Since the relative pronoun {\it whom} (of type $(n\backslash n)/(s/np)$)
        wants to enter into composition on its right
        with the relative clause body, we'd like to assign type $s/np$ to the latter.
        In order to show that {\it Francesca dances with} is indeed of type
        $s/np$, we make a {\it hypothetical assumption} and suppose to have
        a ``ghost word'' of type $np$ on its right. It's easy to derive the category
        $s$ for the sentence {\it Francesca dances with {\rm np}}. By withdrawing
        the hypothetical $np$ assumption, we conclude that {\it Francesca dances
        with} has type $s/np$.

        We can say that the cancelled  hypothesis is the analogous of a
        ``trace'' {\it à la} Chomsky moving {\it whom} before {\it Francesca}.

    \subsubsection{Transitivity}
    In the framework of CCGs a difficulty arises when we try to show the
    well-formedness of \newline
    \centerline{$\stackrel{s/(np\backslash s)}{\mbox{\ \ \ \ he\ \ \ \ }} \stackrel{np\backslash s/np}{\mbox{\ \ likes\ \ }}
            \stackrel{(s/np)\backslash s}{\mbox{\ \ him\ \ }}$}
    so some authors proposed to introduce two new rules,
    which are often referred to as `transitivity rules':
    \begin{eqnarray*}
        (x/y)(y/z) &\rightarrow& x/z,\\
        (x\backslash y)(y\backslash z) &\rightarrow& x\backslash z
    \end{eqnarray*}
    It's easy to show that such rules are derivable in Lambek Calculus, as we can easily see from
    the following proof tree:
    \begin{figure}[htbp]
        \begin{center}
            \includegraphics[height=5cm]{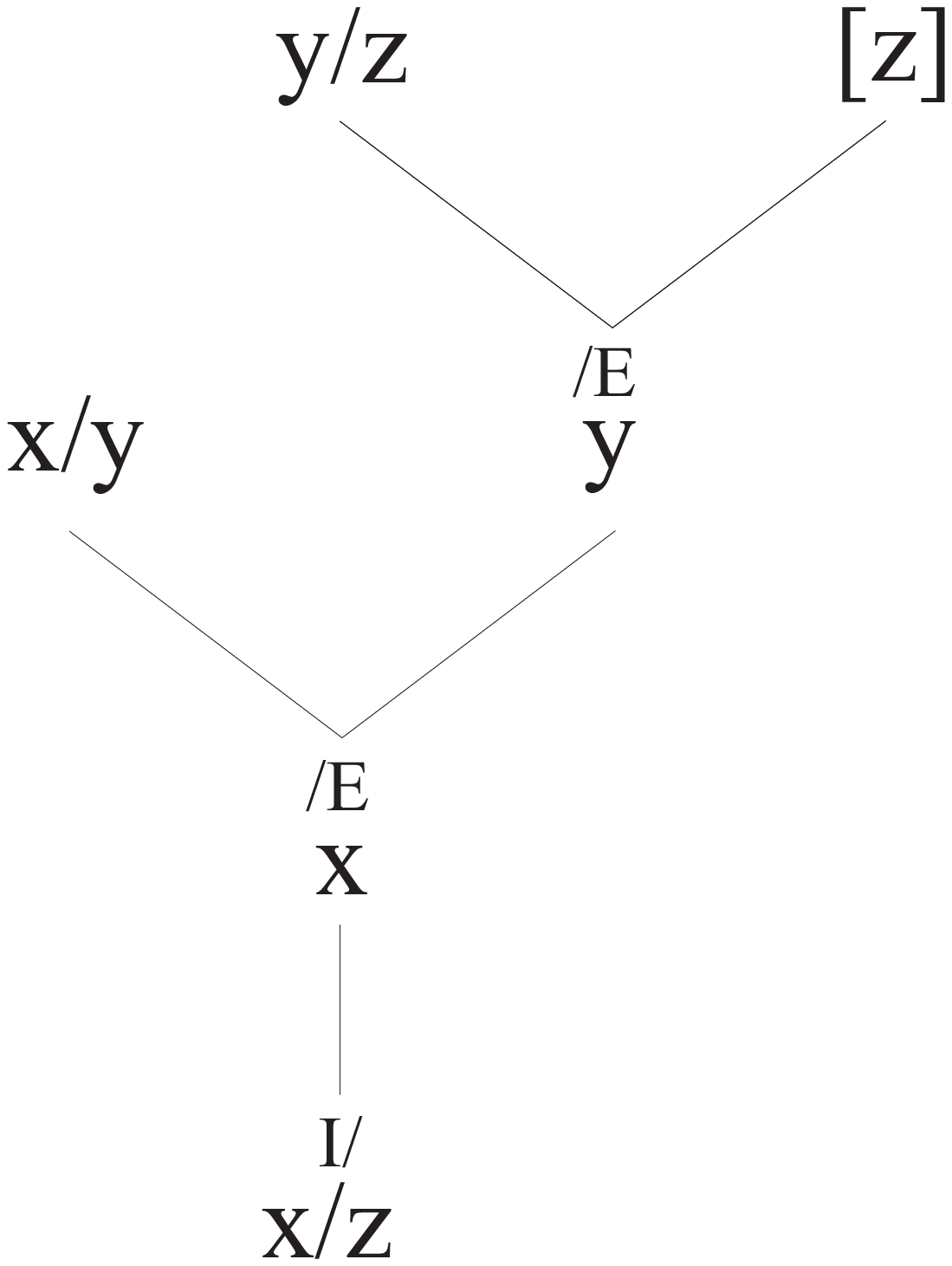}
        \end{center}
    \end{figure}

    \subsection{Lambek Grammars and Montague Semantics}

    \label{semsec}
    From a linguistic point of view, one of the main reasons of
    interest in Lambek grammars lies in the natural interface that
    proof-tree structures provide for Montague-like
    semantics. Just like Curry-Howard isomorphism shows that
    simply typed $\lambda$-terms can be seen as proofs in
    intuitionistic logics, and vice-versa, syntactical analysis
    of a sentence in a Lambek grammar is a proof in Lambek
    calculus, which is naturally embedded into intuitionistic
    logics. Indeed, if we read $B/A$ and $A\backslash B$ like the
    intuitionistic implication $A\rightarrow B$, every rule in
    Lambek calculus is a rule of intuitionistic logics.

    In order to fully appreciate this relation between syntax and
    semantics which is particularly strong for Lambek grammars, we
    define a morphism between syntactic types and semantic types:
    the latter are formulas of a minimal logics (where the only
    allowed connector is $\rightarrow$, that is, intuitionistic
    implication) built on the two types $e$ (entity) and $t$ (truth
    values).\newline

    \begin{tabular}{rcll}
        (Syntactic type)* &=& Semantic type &\\
        \hline
        $s^*$ &=& $t$  (a sentence is a proposition)\\
        $sn^*$ &=& $e$ (a nominal sintagma denotes an entity)\\
        $n^*$ &=& $e \rightarrow t$ (a noun is a subset of
        entities)
    \end{tabular}
    \ \newline\newline

    $(A\backslash B)^* = (B/A)^* = A^* \rightarrow B^*$ extends
    $(\_)^*$ to every types.\newline

    The lexicon associates also to every word $w$ a $\lambda$-term
    $\tau_k$ for every syntactic type $t_k \in {\cal L}(w)$, such that
    the type of $\tau_k$ is precisely $t_k^*$, the semantic type
    corresponding to that syntactic type. We introduce some
    constants for representing logical operations of
    quantification, conjunction etc:\newline

    \begin{center}
    \begin{tabular}{|c|c|}
        \hline
        Constant  & Type\\
        \hline
        $\exists$ & $(e\rightarrow t)\rightarrow t$\\
        $\forall$ & $(e\rightarrow t)\rightarrow t$\\
        $\wedge$  & $t\rightarrow(t\rightarrow t)$\\
        $\vee$    & $t\rightarrow(t\rightarrow t)$\\
        $\supset$ & $t\rightarrow(t\rightarrow t)$\\
        \hline
    \end{tabular}
    \end{center}

    Let the following be given:
    \begin{itemize}
        \item a syntactical analysis of $w_1 \ldots w_n$ in Lambek
        calculus, that is to say, a derivation ${\cal D}$ of $t_1,
        \ldots, t_n \vdash s$ and
        \item the semantics for every word $w_1, \ldots, w_n$, that
        is to say, $\lambda$-terms $\tau_i:t_i^*$,
    \end{itemize}
    then we get the semantics of the sentence by simply applying
    the following algorithm:
    \begin{itemize}
        \item Substitute in ${\cal D}$ every syntactic type with
        its corresponding semantic image; since intuitionistic
        logics is an extension of Lambek calculus, we get a
        derivation ${\cal D}^*$ into intuitionistic logic of
        $t_1^*, \ldots, t_n^* \vdash t = s^*$;
        \item this derivation in intuitionistic logic due to
        Curry-Howard isomorphism can be seen as a simply typed
        $\lambda$-term ${\cal D}^*_\lambda$, containing a free
        variable $x_i$ of type $t_i^*$ for every word $w_i$;
        \item in ${\cal D}^*_\lambda$ replace each variable $x_i$
        with $\lambda$-term $\tau_i$, equally typed with $t_i^*$;
        \item reduce the $\lambda$-term resulting at the end of the
        previous step, and we get the semantic representation of
        the analyzed sentence.
    \end{itemize}

    Let's consider the following example (taken from \cite{retore96}):
    \begin{center}
        \begin{tabular}{rl}
        {\bf word}      &       {\bf Syntactic type $t$}\\
                        &       {\bf Semantic type $t^*$}\\
                        &       {\bf Semantic representation: a $\lambda$-term of
                                type $t^*$}\\
        \hline
        some            &       $(s/(sn\backslash s))/n$\\
                        &       $(e\rightarrow t)\rightarrow((e\rightarrow t)\rightarrow t)$\\
                        &       $\lambda P: e\rightarrow t\ \lambda
                                Q:e\rightarrow t(\exists(\lambda x:
                                e(\wedge(P x)(Q x))))$\\
        \hline
        sentences       &       $n$\\
                        &       $e\rightarrow t$\\
                        &       $\lambda x:e(\mbox{\tt sentence }
                                x)$\\
        \hline
        talkabout      &       $sn\backslash(s/sn)$\\
                        &       $e\rightarrow(e\rightarrow t)$\\
                        &       $\lambda x:e\ \lambda y:e ((\mbox{\tt talkabout
                                }x)y)$\\
        \hline
        themselves      &       $((sn\backslash s)/sn)\backslash (sn\backslash
                                s)$\\
                        &       $(e\rightarrow(e\rightarrow t))\rightarrow(e\rightarrow
                                t)$\\
                        &       $\lambda P:e\rightarrow(e\rightarrow t) \lambda x:e((P
                                x)x)$\\
        \hline

        \end{tabular}
    \end{center}

    First of all, we'll prove that {\it Some sentences talk about
    themselves} is a well formed-sentence, that is, it belongs to
    the language generated by the lexicon at issue. This means
    building a natural deduction of:
    \begin{displaymath}
        (s/(sn\backslash s))/n, n, sn\backslash(s/sn),
        ((sn\backslash s)/sn)\backslash(sn\backslash s)\vdash s.
    \end{displaymath}
    If we indicate with $S, N, T, M$ the left-hand side of
    syntactic types we get
    \begin{displaymath}
        \begin{prooftree}
            \begin{prooftree}
                S \vdash (s/(sn\backslash s))/n\ \ \
                N \vdash n
                \justifies
                S, N \vdash s/(sn\backslash s)
                \using [/E]
            \end{prooftree}
            \ \ \ \
            \begin{prooftree}
                T \vdash (sn\backslash s)/sn\ \ \
                M \vdash ((sn\backslash s)/sn)\backslash(sn\backslash s)
                \justifies
                T, M \vdash sn\backslash s
                \using [\backslash E]
            \end{prooftree}
            \justifies
            S, N, T, M \vdash s
            \using [\backslash E]
        \end{prooftree}
    \end{displaymath}

    By applying the isomorphism between syntactic and semantic
    types, we get the following intuitionistic proof, where $S^*,
    N^*, T^*, M^*$ are the abbreviations for semantic types
    associated to $S, N, T, M$:
    \begin{displaymath}
        \begin{prooftree}
            \begin{prooftree}
                S^* \vdash (e\rightarrow t)\rightarrow(e\rightarrow t)\rightarrow t
                \ \
                N^* \vdash e\rightarrow t
                \justifies
                S^*, N^* \vdash (e\rightarrow t)\rightarrow t
                \using [\rightarrow E]
            \end{prooftree}
            \
            \begin{prooftree}
                T^* \vdash e\rightarrow e \rightarrow t
                \ \
                M^* \vdash (e\rightarrow e \rightarrow t)\rightarrow e
                \rightarrow t
                \justifies
                T^*, M^* \vdash e\rightarrow t
                \using [\rightarrow E]
            \end{prooftree}
            \justifies
            S^*, N^*, T^*, M^* \vdash t
            \using[\rightarrow E]
        \end{prooftree}
    \end{displaymath}

    The $\lambda$-term coding this proof is simply $((sn)(tm))$ of
    type $t$, where $s,n,t,m$ are variables of types respectively
    $S^*, N^*, T^*, M^*$.

    By replacing these variables with $\lambda$-terms of the same
    types associated by the lexicon to the words,  we get the
    following $\lambda$-term of type $t$:
    \begin{center}
        $((\lambda P\ \lambda Q\ (\exists\ (\lambda x(\wedge(P\
        x)(Q\ x)))))(\lambda x\ (\mbox{\tt sentence }x)))$\\
        $((\lambda P\ \lambda x\ ((P\ x)x))(\lambda x\ \lambda y\
        ((\mbox{\tt talkabout }x)y)))$
    \end{center}
    \begin{displaymath}
        \downarrow \beta
    \end{displaymath}
    \begin{displaymath}
        (\lambda Q\ (\exists (\lambda x(\wedge(\mbox{\tt sentence
        }x)(Q\ x)))))(\lambda x((\mbox{talkabout }x)x))
    \end{displaymath}
    \begin{displaymath}
        \downarrow \beta
    \end{displaymath}
    \begin{displaymath}
        (\exists(\lambda x(\wedge(\mbox{\tt sentence
        }x)((\mbox{\tt talkabout }x)x))))
    \end{displaymath}

    If we recall that the $x$ in this last term is of type $e$,
    the latter reduced term represents the following formula in
    predicate calculus:
    \begin{displaymath}
        \exists x:e (\mbox{sentence }(x) \wedge
        \mbox{talkabout}(x,x))
    \end{displaymath}
    which is the semantic representation of the previously
    analyzed sentence.

%% file: rigidlambek_report.tex
\section{Rigid Lambek Grammars}
    \label{rlgsection}
    In the present section we introduce the notion of {\it rigid
    Lambek grammar} (often referred to as RLG), whose learnability
    properties will be the subject of our inquiry in section
    \ref{learnlambeksection}.
    Basic notions and results presented here are almost trivial
    extensions of what has already been done for
    rigid CCGs (see \cite{kanazawa98}),
    since a specific a specific theory for rigid
    Lambek grammars is still missing.

    \subsection{Rigid and  k-Valued Lambek Grammars}
    A {\it rigid Lambek grammar} is a triple $G=\seq{\Sigma,s,F}$, where
    $\Sigma$ and $s$ are defined like in definition 
    \ref{LambekGrDef}, while $F: \Sigma \rightharpoonup
    Tp$ is a partial function that assigns to each symbol 
    of the alphabet {\it at most one type}.
    We can easily generalize the notion of rigid
    Lambek grammar to the notion of {\it k-valued} Lambek
    grammar\index{k-valued Lambek grammar} by a function $F$ that
    assigns to each symbol of the alphabet {\it at most k types}.
    Formally, $F: \Sigma \rightharpoonup \bigcup_{i=1}^{k}{Tp^k}$. \newline

    Let an alphabet $\Sigma$ be given. We call ${\cal G}_{rigid}$
    the class of rigid Lambek grammars
    over $\Sigma$, and ${\cal G}_{k-valued}$ the
    class of k-valued Lambek grammars over $\Sigma$.

    Let's define two classes of proof-tree structures:
    \begin{eqnarray*}
        {\cal PL}_{rigid}&=&\set{\pl(G)\ |\ G \in {\cal G}_{rigid}},\\
        {\cal PL}_{k-valued}&=&\set{\pl(G)\ |\ G \in {\cal
        G}_{k-valued}}.
    \end{eqnarray*}
    Members of ${\cal PL}_{rigid}$ are called {\it rigid (proof-tree)
    structure languages}, and members of ${\cal PL}_{k-valued}$
    are called {\it k-valued (proof-tree) structure
    languages}.\newline

    Let's define two classes of strings:
    \begin{eqnarray*}
        {\cal L}_{rigid} &=& \set{\naming(G)\ |\ G \in {\cal
            G}_{rigid}},\\
        {\cal L}_{k-valued} &=& \set{\naming(G)\
            |\ G \in {\cal G}_{k-valued}}.
    \end{eqnarray*}
    Members of ${\cal L}_{rigid}$ are called {\it rigid (string) languages},
    and members of ${\cal L}_{k-valued}$ are called
    {\it k-valued (string) languages}.\newline

    \begin{example}
        Let $\set{{\bf well, Francesca, dances}}\subseteq\Sigma$ and let
        $G_1, G_2$ be the following Lambek grammars:
        \begin{eqnarray*}
            G_1 : {\bf Francesca} &\mapsto& x,\\
                  {\bf dances} &\mapsto& x\backslash s,\ y,\\
                  {\bf well} &\mapsto& y\backslash (x\backslash s),\\
            G_2:  {\bf Francesca} &\mapsto& x,\\
                  {\bf dances} &\mapsto& x\backslash s,\\
                  {\bf well} &\mapsto& (x\backslash s)\backslash
                  (x\backslash s).
        \end{eqnarray*}
        Then $G_2$ is a rigid grammar, while $G_1$ is not. $G_1$
        is a 2-valued grammar.
    \end{example}

    \begin{definition}
        Any type $A$ can be written uniquely in the following
        form:
        \begin{displaymath}
            (\ldots((p|A_1)|A_2)|\ldots)|A_n
        \end{displaymath}
        where $B|C$ stands for either $B/C$ or $C\backslash B$ and
        $p \in Pr$. For $0\le i\le n$, we call the subtype
        $(\ldots(p|A_1)|\ldots)|A_i$ of $A$ a {\rm head subtype}
        of $A$. $p$ is the {\rm head} of $A$ and is denoted
        $head(A)$. $A_i$'s are called {\rm argument subtypes} of
        $A$. The number $n$ is called the {\rm arity} of $A$.
    \end{definition}

    The following propositions are almost trivial extensions to
    rigid Lambek grammars of analogous results proved by Kanazawa for CCGs
    in \cite{kanazawa98}.
    However, they deserve some attention since they can provide a
    first superficial insight about properties of RLGs.\newline

    First of all we prove a {\it hierarchy theorem} about strong
    generative capacity of k-valued Lambek grammars.
    \begin{proposition}
        \label{treehyer}
        Let ${\tt a} \in \Sigma$. For each $i \ge 1$, let $T_i$ be
        the following proof-tree structure:
        \begin{figure}[htbp]
            \center{\includegraphics[height=3cm]{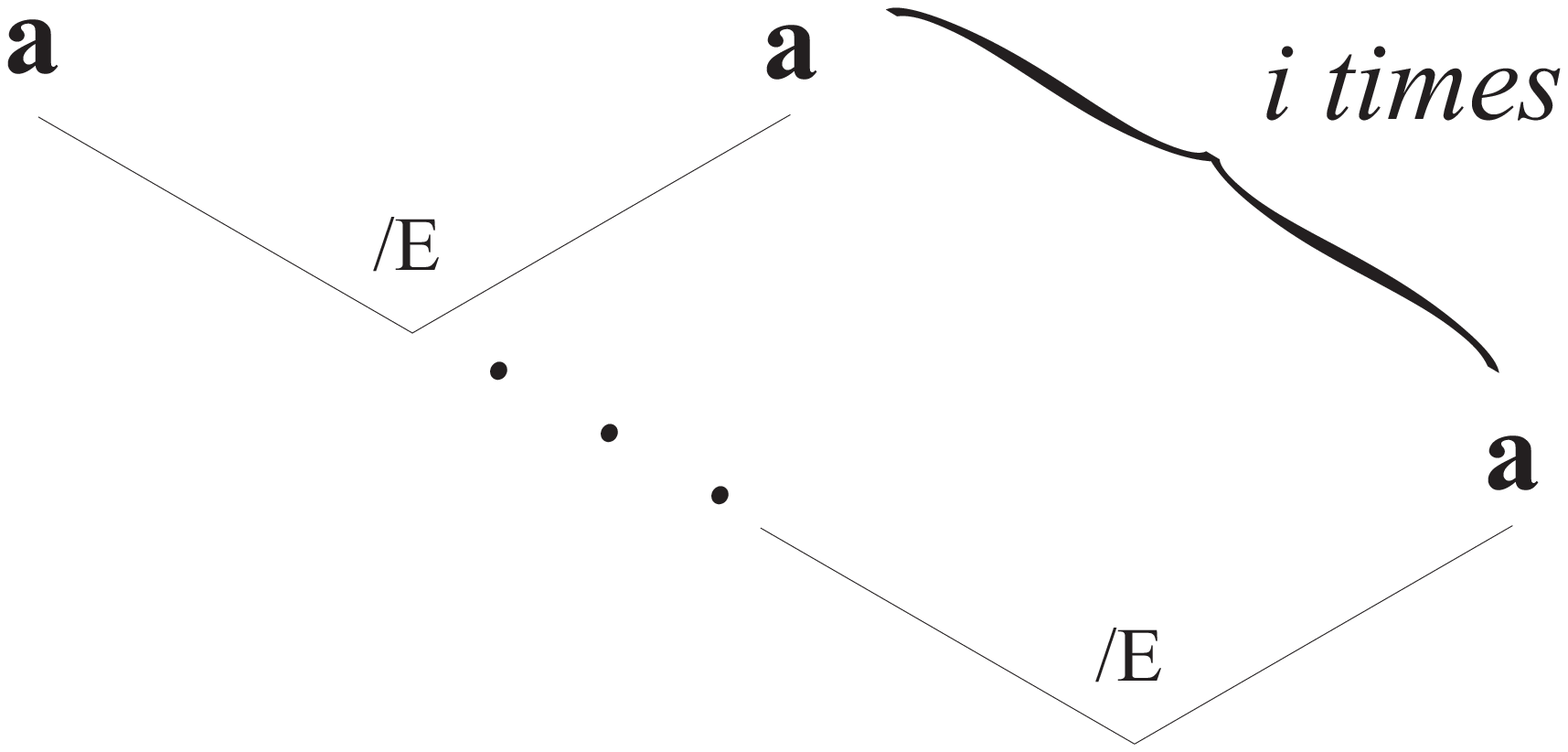}}
        \end{figure}\newline
        Then for each $k \ge 1$,
        \begin{displaymath}
            \set{T_1, \ldots, T_k} \in {\cal PL}_{k+1-valued}-{\cal
            PL}_{k-valued}.
        \end{displaymath}
        Thus, for each $k \in \naturals$, ${\cal
            PL}_{k-valued} \subset {\cal PL}_{k+1-valued}$.
    \end{proposition}
    {\it Proof.} (See \cite{kanazawa98}) Let $G_k$ be the
    following k+1-valued grammar:
    \begin{eqnarray*}
        G_k: a &\mapsto&x,\\
               &&s/x,\\
               &&(s/x)/x,\\
               &&\vdots\\
               &&(\ldots((s/\underbrace{x)/x)/\ldots)/x}_{k\ times}.
    \end{eqnarray*}
    Then one can easily verify that $\set{T_1, \ldots, T_k}
    \subset \pl(G_k)$.

    Let $G$ be a grammar such that $\set{T_1, \ldots, T_k}
    \subset \pl(G)$: we will show that $G$ is at least k+1-valued.

    Let ${\cal P}_i$ be a parse of $T_i$ in $G$ for $1\le i \le
    k$. Then the leftmost leaf of ${\cal P}_i$ is the ultimate
    functor of ${\cal P}_i$, and if we call $A_i$ the type
    labeling it, we can easily verify that the its arity must be
    exactly $i$. Thus, $i \not = j$ implies $A_i\not = A_j$.

    We show that there is at least one type $B$ such that $G:
    a\mapsto B$ and $B \not \in\set{A_1, \ldots, A_k}$. Since the
    relation ``is an argument subtype of'' is well-founded, there is
    at least one $i$ such that the argument subtypes of $A_i$ are
    not in $\set{A_1, \ldots, A_k}$. But in order to produce
    ${\cal P}_i$, any argument subtype of $A_i$ must be a type
    assigned to $a$ by $G$. Therefore $G$ must be at least
    k+1-valued.\newline

    The proof of proposition \ref{treehyer} shows
    \begin{corollary}
        There is no Lambek grammar G such that {\rm PL}$(G)=\Sigma^P$.
    \end{corollary}

    \begin{lemma}
        Let G be a rigid Lambek grammar. Then for each proof-tree
        structure T, there is at most one partial parse tree
        ${\cal P}$ such that T is the structure of ${\cal P}$.
    \end{lemma}
    {\it Proof.} By induction on the construction of
    $T$.\newline

    {\it Induction basis.} $T=c \in \Sigma$. Any partial parse tree
    ${\cal P}$ whose structure is $T$ is a height 0 tree whose
    only node is labeled by the symbol $c$ and a type $A$
    such that $G: c\mapsto A$. Since $G$ is rigid, there
    is at most one such type $A$. Then ${\cal P}$, if it
    exists, is unique.\newline

    {\it Induction step.} There are 4 cases to consider:
    \begin{enumerate}
        \item $T$ is the following proof-tree structure:
            \begin{figure}[htbp]
                \center{\includegraphics[height=2cm]{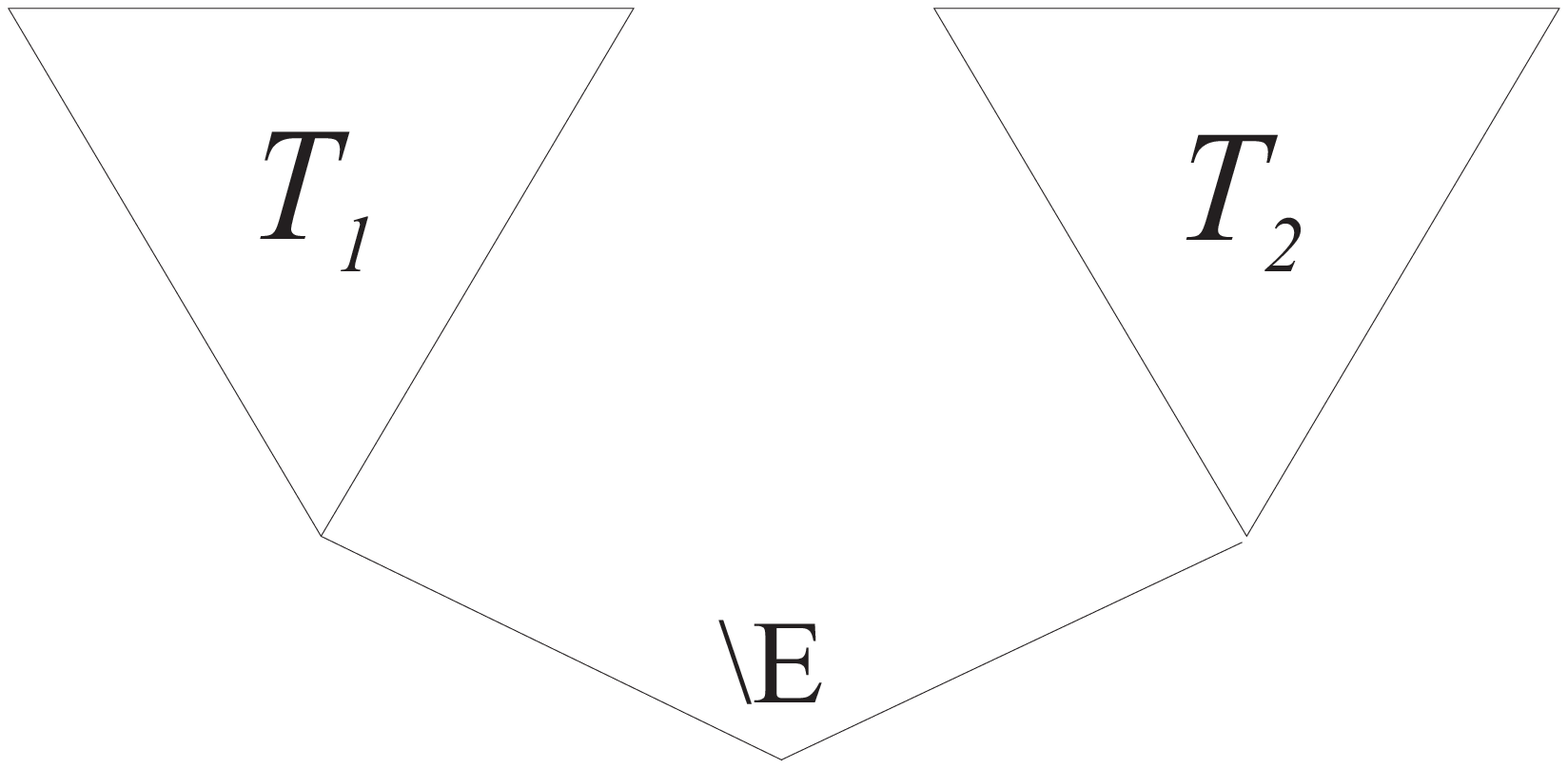}}
            \end{figure}\newline
            Then any partial parse tree of $G$ whose structure is $T$ has
            the form
            \begin{figure}[htbp]
                \center{\includegraphics[height=2.8cm]{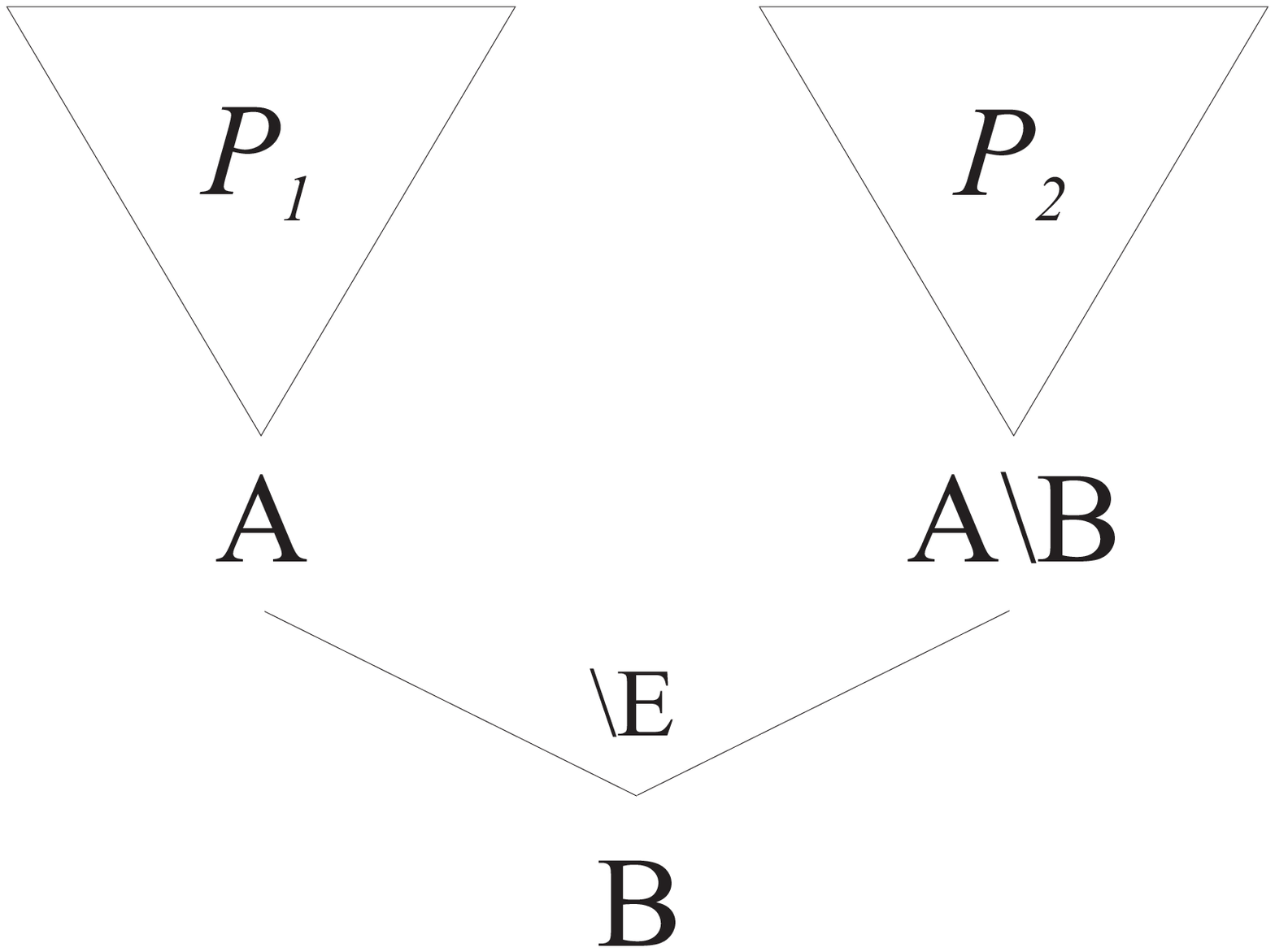}}
            \end{figure}
            where ${\cal P}_1$ and ${\cal P}_2$ are partial parse trees of
            $G$ whose structures are $T_1$ and $T_2$, respectively. By
            induction hypothesis, ${\cal P}_1$ and ${\cal P}_2$ are
            unique. This means that the type label $B$ is also uniquely
            determined, so ${\cal P}$ is also unique.\newline
        \item  Exactly like Case 1, with $/E$ in place of $\backslash E$.
        \item  $T$ is the following proof-tree structure:
            \begin{figure}[htbp]
                \center{\includegraphics[height=2.5cm]{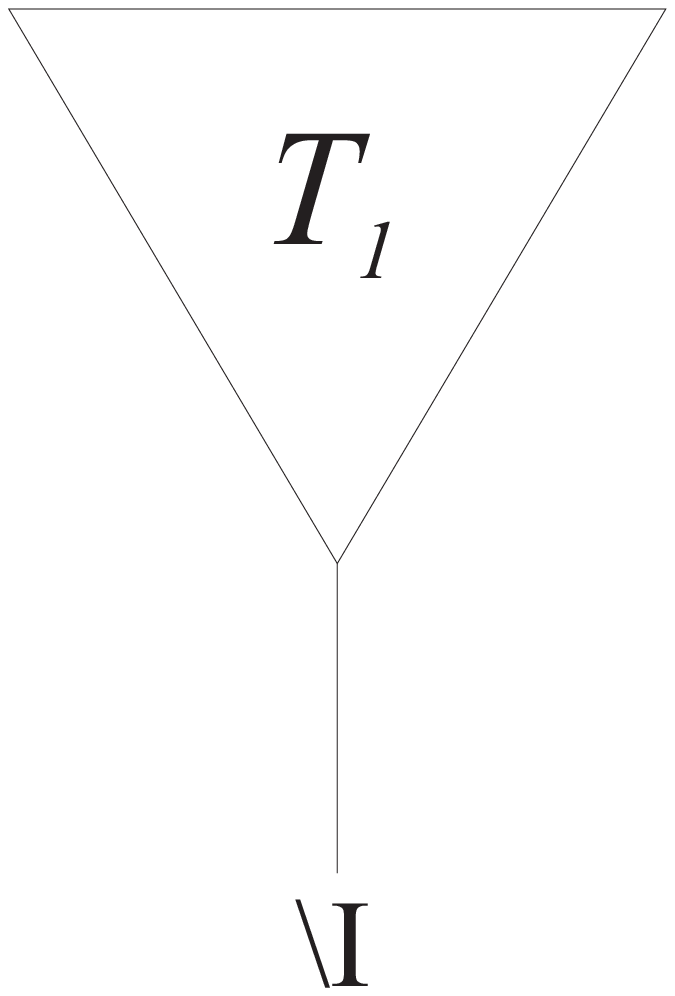}}
            \end{figure}\newline
            Then any partial parse tree of $G$ whose structure is $T$ has
            the form
            \begin{figure}[htbp]
                \center{\includegraphics[height=3.5cm]{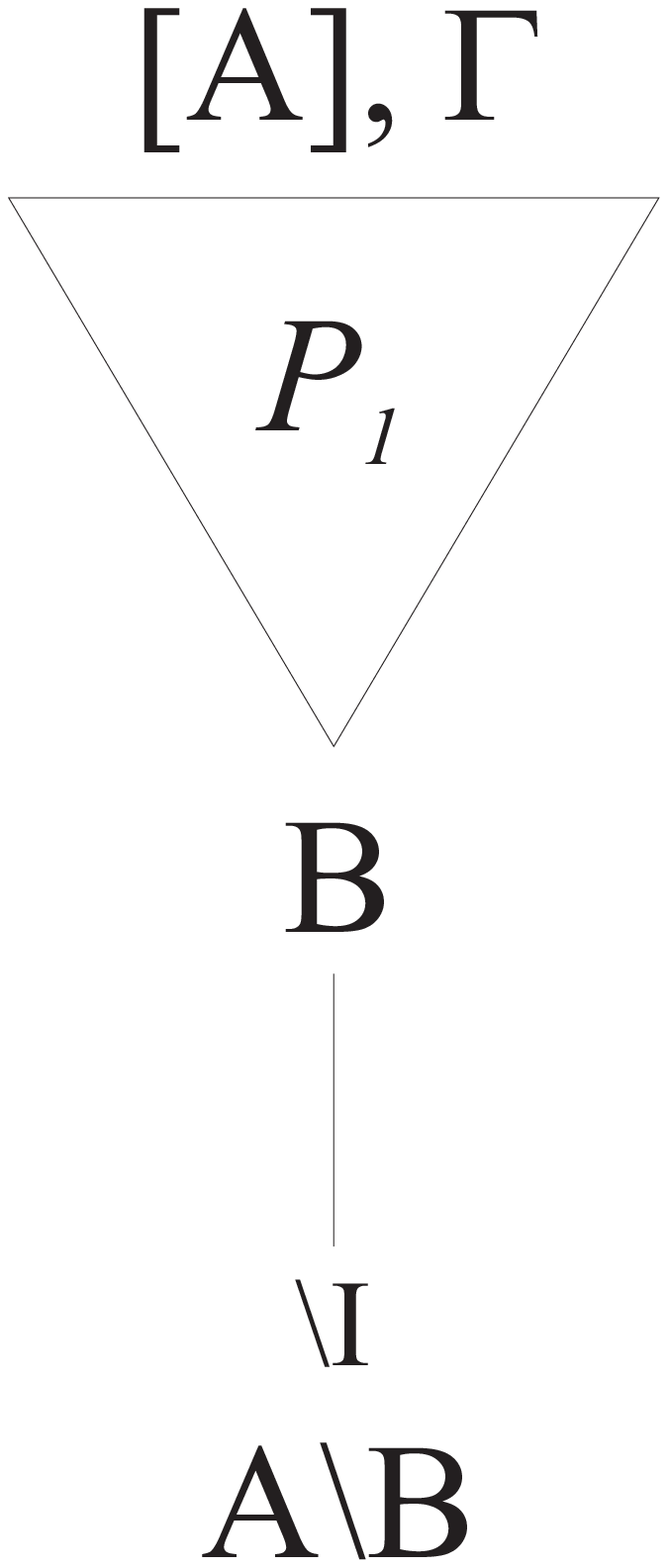}}
            \end{figure}
            where ${\cal P}_1$ is a partial parse tree of $G$ whose
            structure is $T_1$. By induction hypothesis, ${\cal P}_1$ is
            unique. This means the the type label $A\backslash B$ is
            uniquely determined, so ${\cal P}$ is also unique.
        \item Exactly like Case 3, with $/I$ in place of $\backslash I$.
    \end{enumerate}
    \begin{corollary}
        If G is a rigid Lambek grammar, each proof-tree structure
        $T \in {\rm PL}(G)$ has a unique parse.
    \end{corollary}

    Note that last corollary doesn't state that if $G$ is rigid,
    then each string $s \in L(G)$ has a unique parse: in general for each
    sentence there are infinitely many proof trees,
    as extensively shown in section \ref{ptreessec}.

    \begin{lemma}
        Let G be a rigid Lambek grammar. Then for each incomplete
        proof-tree structure T, there is at most one incomplete
        parse tree ${\cal P}$ of G such that T is the structure of
        ${\cal P}$.
    \end{lemma}
    {\it Proof.} See \cite{kanazawa98} trivially extended to
    Lambek grammars.


    \subsection{Most General Unifiers and $\sqcup$ Operator}
    \label{UnifSec}
    \label{latsec}
    Unification plays a crucial role in automated theorem proving
    in classical first-order logic and its extensions (see, for
    example, \cite{fitting96} for an exposition of its use in
    first-order logic). Since types are just a special kind of
    terms, the notion of unification applies straightforwardly to
    types.
    \begin{definition}
        Let $A$ and $B$ be types. A substitution $\sigma$ is a
        {\rm unifier} of $A$ and $B$ if $\sigma(A)=\sigma(B)$. A
        unifier $\sigma$ is a {\rm most general unifier} of $A$
        and $B$, if for any other unifier $\tau$ of $A$ and $B$,
        there exists a substitution $\eta$, such that
        $\tau=\sigma\circ\eta$, i.e. $\tau(C)=\eta(\sigma(C))$,
        for $C=A$ or $C=B$.
    \end{definition}
    A substitution $\sigma$ is said to {\it unify} a set {\bf A}
    of types if for all $A_1, A_2 \in {\bf A}$,
    $\sigma(A_1)=\sigma(A_2)$. We say that $\sigma$ unifies a
    family of sets of types, if $\sigma$ unifies each set in the
    family.

    A most general unifier is unique up to `renaming of
    variables'.
    \begin{example}
        Let ${\cal A}$ consist of the following sets:
        \begin{eqnarray*}
            A_1 &=& \set{x_1/x_2, x_3/x_4},\\
            A_2 &=& \set{x_5\backslash(x_3\backslash t},\\
            A_3 &=& \set{x_1\backslash t, x_5}.\\
        \end{eqnarray*}
        Then the most general unifier of ${\cal A}$ is:
        \begin{displaymath}
            \sigma = \set{x_3 \mapsto x_1, x_4 \mapsto x_2, x_5 \mapsto
            x_1\backslash t}.
        \end{displaymath}
    \end{example}

    There are many different efficient algorithms for unification,
    which decide whether a finite set of types has a unifier and,
    if it does, compute a most general unifier for it.
    For illustration purposes, we present here a non-deterministic
    version of an unification algorithm.

    Our algorithm  uses the notion of {\it disagreement pair}.
    The easiest way to define disagreement pair is to
    consider the types to be tree-like:
    \begin{definition}
        Let $A$ and $B$ be two types. A {\rm disagreement pair}
        for $A$ and $B$ is a pair of subterms of $A$ and $B$,
        $A^\prime, B^\prime$, such that $A^\prime\not = B^\prime$
        and the path from the root of $A$ to the root of
        $A^\prime$ is equal to the path from the root of $B$ to
        the root of $B^\prime$.
    \end{definition}
     The following, non-deterministic version of the unification
     algorithm is taken from \cite{fitting96}:\newline

     {\bf \sc Unification Algorithm}.
     \begin{itemize}
        \item{{\bf input:} two types $A$ and $B$;}
        \item{{\bf output:} a most general unifier $\sigma$ of $A, B$,
        if it exists, or a correct statement that $A$ and $B$ are not
        unifiable.}
     \end{itemize}
     \begin{quote}
        Let $\sigma:=\epsilon$\newline
        While $\sigma(A)\not=\sigma(B)$ do\newline
        {\bf begin}\newline
        choose a disagreement pair $A^\prime, B^\prime$ for
        $\sigma(A), \sigma(B)$;\newline
        if neither $A^\prime$ nor $B^\prime$ is a variable, then
        FAIL;\newline
        let $x$ be whichever of $A^\prime, B^\prime$ is a variable (if both are, choose one)\\
        and let $C$ be the other one of $A^\prime, B^\prime$\newline
        if $x$ occurs in $C$, then FAIL;\newline
        let $\sigma:=\sigma\circ\set{x\mapsto C}$;\newline
        {\bf end}
    \end{quote}

    The previous algorithm present one of many efficient
    algorithms for unification, so we the following is a
    well-defined notion:

    \begin{definition}
        We define a computable partial function $mgu$
        that maps a finite family ${\cal A}$ of finite sets of
        types to a most general unifier $mgu({\cal A})$, if ${\cal
        A}$ is unifiable.
    \end{definition}
    The set ${\cal G}_{rigid}$ of all rigid Lambek grammars is
    partially ordered by $\sqsubseteq$.
    \begin{definition}
        Let ${\cal G} \subseteq {\cal G}_{rigid}$, and let $G \in
        {\cal G}$.Then $G$ is called an {\rm upper bound} of ${\cal G}$
            if for every $G^\prime \in {\cal G}$, $G^\prime \sqsubseteq G$.
    \end{definition}            

    We introduce here a new operator among rigid grammars that
    will be used to prove an interesting property for our learning
    algorithm at the end of the fifth chapter.
    \begin{definition}
        \label{sqdef}
        Let $G_1$ and $G_2$ be rigid Lambek grammars. We can
        assume that $G_1$ and $G_2$ have no common variables (if
        they do, we can always choose a suitable alphabetic
        variant of one of them such that $Var(G_1)\cap Var(G_2) =
        \emptyset$). Let
        \begin{displaymath}
            {\cal A}=\set{\set{A\ |\ G_1 \cup G_2:c \mapsto A}\ |\ c\in dom(G_1\cup G_2)}
        \end{displaymath}
        and let
        \begin{displaymath}
            \sigma = mgu ({\cal A}).
        \end{displaymath}
        Note that $G_1 \cup G_2$ is a 2-valued grammar. Then we
        define $G_1\sqcup G_2$ as follows:
        \begin{displaymath}
            G_1 \sqcup G_2 = \sigma[G_1 \cup G_2].
        \end{displaymath}
        If ${\cal A}$ is not unifiable, then $G_1 \sqcup G_2$ is
        undefined.
    \end{definition}

    \begin{example}
        Let $G_1$ and $G_2$ be the following rigid Lambek
        grammars:
        \begin{eqnarray*}
            G_1: a &\mapsto& s/x,\\
                 b &\mapsto& x,\\
            G_2: b &\mapsto& y\backslash s,\\
                 c &\mapsto& y.
        \end{eqnarray*}
        Then
        \begin{eqnarray*}
            G_1\sqcup G_2: a &\mapsto& s/(y\backslash s),\\
                           b &\mapsto& y\backslash s,\\
                           c &\mapsto& y.\\
        \end{eqnarray*}
    \end{example}

    Obviously, from definition \ref{sqdef}, we have
    \begin{lemma}
        \label{rigidlemma}
        If $G_1 \sqcup G_2$ exists, then $G_1 \sqsubseteq G_1
        \sqcup G_2$ and $G_2 \sqsubseteq G_1 \sqcup G_2$.
    \end{lemma}

    \begin{proposition}[Kanazawa, 1998]
        \label{rigidprop}
        Let $G_1, G_2 \in {\cal G}_{rigid}$. If $\set{G_1, G_2}$
        has un upper bound, then $G_1 \sqcup G_2$ exists and it's
        the least upper bound of $\set{G_1, G_2}$.
    \end{proposition}
    {\it Proof.} (See \cite{kanazawa98}).

%% file: learnlambek_report.tex
\section{Learning Rigid Lambek Grammars from Structures}
\label{learnlambeksection}

In the present chapter we will explore a model of learning for
Rigid Lambek Grammars based on {\it positive structured data}. In
addition to the standard model where sentences are presented to
the learner as flat sequences of words, in this somewhat enriched
model, strings come with additional information about their ``deep
structure''. Following the approach sketched in section \ref{ptreessec}, 
largely indebted with Tiede's study on proof trees in Lambek calculus as
grammatical structures for Lambek grammars (see \cite{tiede99}),
in our model each sentence comes to the learner with a structure
in the form of a {\it proof tree structure} 
as extensively described in section \ref{ptreessec}.

Formally, given a finite alphabet $\Sigma$, we will present a
learning algorithm for the grammar system $\seq{{\cal G}_{rigid},
\Sigma^P, \pl}$: that is to say, samples to which the learner is
exposed to are proof-tree structures over the alphabet $\Sigma$,
and guesses are made about the set of rigid Lambek grammars that
can generate such a set of structures.

We follow the advice of Kanazawa (see \cite{kanazawa98}) who
underlines how such an approach, which turns out to be quite
logically independent from an approach based on flat strings of
words, seems to make the task of learning easier but doesn't
trivialize it. If, on one hand, in the process of learning from
structures the learner is provided with more information, on the
other hand the criterion for successful learning is stricter. 
It is not sufficient that the string language of $G$
contains exactly the {\it yields} of the structures in the input
sequence, the
learning function is required to converge to a grammar $G$ that
generates all the grammatical {\it structures} which appear in the input
sequence. We could say that the learning function must converge to
a grammar that is both weakly and strongly equivalent to the
grammar that generated the input samples.

Clearly, from a psycholinguistic point of view, both learning from
flat strings and from proof tree structures are quite unrealistic
models of first language acquisition by human beings. In the first
case, experimental evidences (see \cite{pinker94}) show that
children can't acquire a language simply by passively listening to
flat strings of words. First of all, we can think that prosody (or
punctuation, in written text) can provide ``structural'' information
to the children on the syntactic bracketing of the sentences she is
exposed to (although they do not always coincide) and it is known
that prosody is needed to learn a language for a child.
Furthermore, another interesting evidence of the fact that a child
needs something more to learn her mother tongue is given by the
fact that no children can improve their grammatical skills during
the early stages of their language acquisition process by watching
TV: it seems very likely they need ``richer data'' than simple
sentences uttered by an adult. Some researchers (see
\cite{tellier99}) hypothesize this additional information comes to
the children as the semantic content of the first sentences she is
exposed to, whose she could have a first, primitive grasp through
first sensory-motor experiences.

On the other hand, it is also highly unlikely that a child can have
access to something like a proof tree structure of the sentence
she is exposed to. Our belief is that a good formal model for the
process of learning should rely on something ``halfway'' between
flat strings of words and highly structured and complete
information coming from the proof tree structure of the sentence.
However, since, as we've already seen in section \ref{semsec},
proof tree structures provide a very natural support for a
Montague-like semantics, we think that our model for learning a
rigid Lambek grammar from structured data represents a first,
simple but meaningful approximation of a more plausible model of
learning.

In any case, even though in most of real-world applications only
unstructured data are available, we are often interested not only
in the sentences that a grammar derives, but also in derivation
strings that grammar assigns to sentences. That is, we generally
want a grammar that makes structural sense.

\subsection{Grammatical Inference as Unification}
We set our inquiry over the learnability for rigid Lambek grammars
in the more general logical framework of the Theory of
Unification. We will stick to the approach described in
\cite{nicolas99} based on the attempt to reduce the process of
inferring a categorial grammar to the problem of unifying a set of terms.
This approach establishes a fruitful connection between Inductive
Logic Programming techniques and the field of Grammatical
Inference, a connection that has already been proved successful in
devising efficient algorithms to infer k-valued CCGs from positive
structured data (see \cite{kanazawa98}). Our aim is to exploit as
much as possible what has already been done in this direction by
exploring the possibility of adapting existing algorithms for CCGs
to rigid Lambek grammars.

    \subsection{Argument Nodes and Typing Algorithm}
    \label{argumentsec}
    Our learning algorithm is based on a process of
    labeling for the nodes of a set of proof tree structures. We
    introduce here the notion of {\it argument node} for a normal form proof tree.
    We will be a bit sloppy in defining such a notion, and
    sometimes we will use the same notation to indicate a node and
    the type it's labeled by, when this doesn't engender
    confusion, and much will be left to the graphical''
    interpretation of trees and their nodes. However, we can
    always think of a node as a De Bruijn-like object (see
    \cite{debruijn72}) without substantially affecting the meaning of what will be
    proved.

    \begin{definition}
        \label{argdef}
        Let ${\cal P}$ be a normal form partial parse tree. Let's define
        inductively the set $Arg({\cal P})$ of {\it argument
        nodes} of ${\cal P}$. There are three cases to consider:
        \begin{itemize}
            \item ${\cal P}$ is a single node labeled by a type
            $x$, which is the only member of $Arg({\cal P})$.
            \item ${\cal P}$ looks like one of the following
            \begin{figure}[htbp]
                \begin{center}
                    \includegraphics[height=2.5cm]{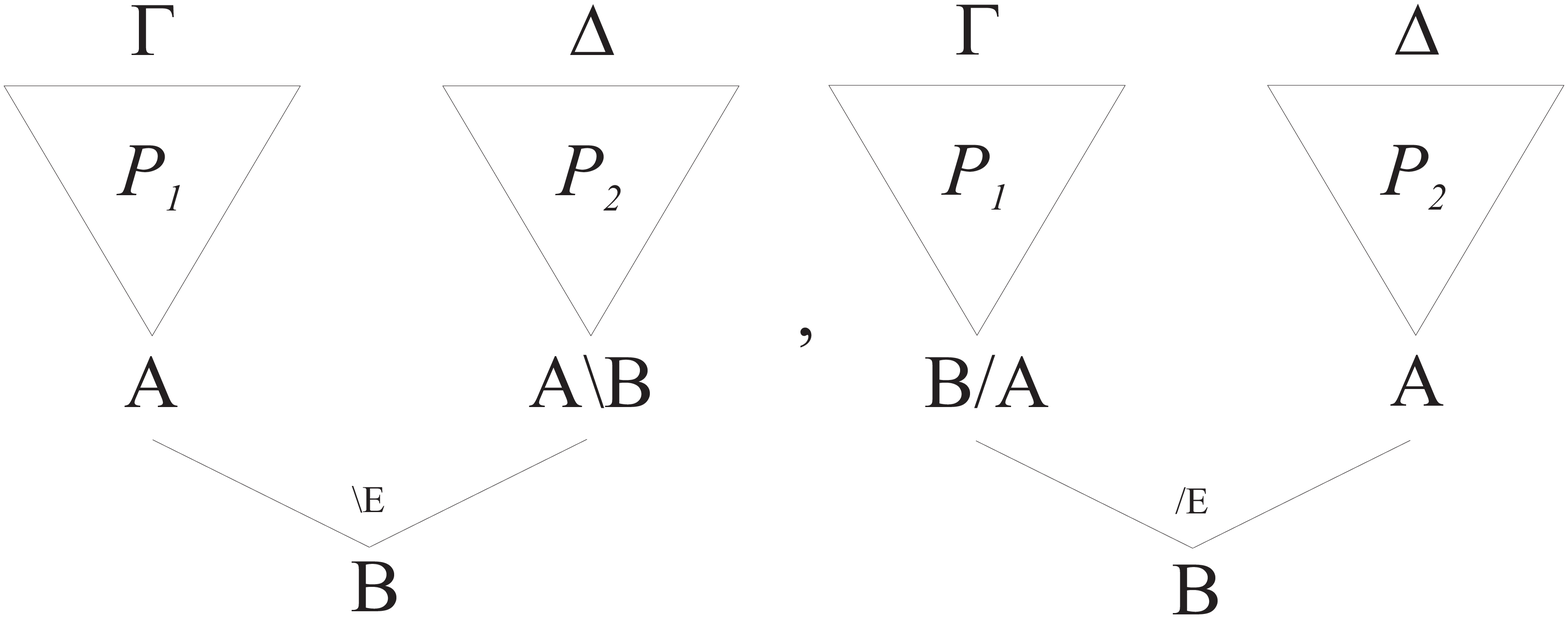}
                \end{center}
            \end{figure}\newpage
            then in the first case
            \begin{displaymath}
                Arg({\cal P}) = \set{Root({\cal P})} \cup Arg({\cal P}_1) \cup Arg({\cal
                P}_2) - \set{Root({\cal P}_2)},
            \end{displaymath}
            and in the second case
            \begin{displaymath}
                Arg({\cal P}) = \set{Root({\cal P})}\cup Arg({\cal P}_1) \cup Arg({\cal
                P}_2) - \set{Root({\cal P}_1)}.
            \end{displaymath}
            \item ${\cal P}$ looks like one of the following
            \begin{figure}[htbp]
                \begin{center}
                    \includegraphics[height=4cm]{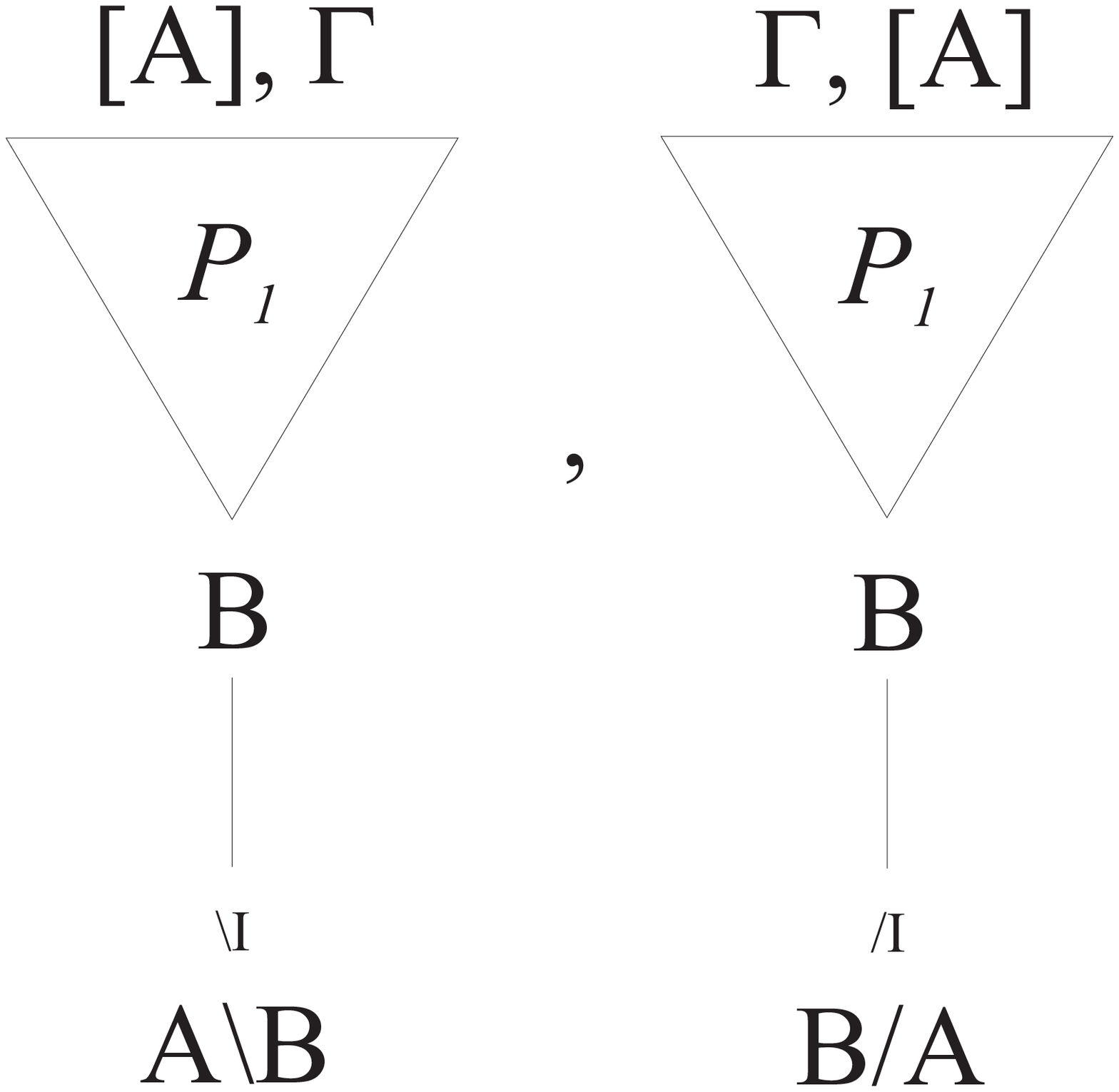}
                \end{center}
            \end{figure}\newline
            then $Arg({\cal P}) = Arg({\cal P}_1)$.
        \end{itemize}
    \end{definition}


    The following proposition justifies our interest for argument nodes for a normal
    form proof tree structure:

    \begin{proposition}
        \label{labelprop}
        Let $t$ be a well formed normal form proof tree structure.
        If each argument node is labeled, then any other node in
        $t$ can be labeled with one and only one type.
    \end{proposition}
    {\it Proof.} We prove that, once argument nodes are labeled,
    any other node can be labeled, by providing a typing
    algorithm; uniqueness of typing follows from the rules applied.\newline\newline
    By induction on the height $h$ of $t$:\newline

    {\it Induction Basis}. There are two cases to consider:
    \begin{enumerate}
        \item $h=0$. Trivially, by definition \ref{argdef}, 
	$t$ is a single argument node, the result of
        the application of a single axiom rule $[ID]$ and by definition
        it's already typed.
        \item $h=1$. Then $t$ must be the result of a single application of
        a $[/E]$ or $[\backslash E]$ rule. By hypothesis and definition
        \ref{argdef}, its two argument nodes are labeled with,
        say, $x_1$ and $x_2$, and the remaining node must be
        labeled according to one of the following rules:\newpage
        \begin{figure}[htbp]
            \begin{center}
                \includegraphics[height=2cm]{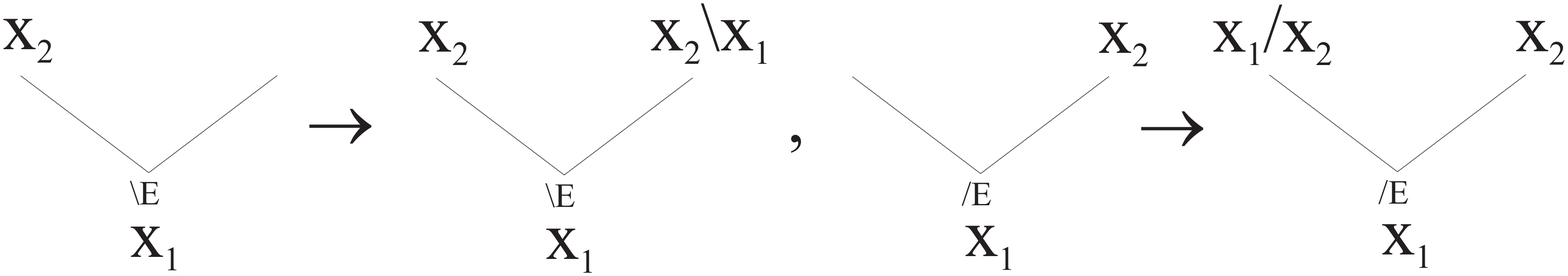}
            \end{center}
        \end{figure}
    \end{enumerate}

    {\it Induction Step}. Let $t$ be a normal form proof tree
    structure of height $h > 1$. There are 3 cases to consider:
    \begin{enumerate}
        \item $t \equiv \backslash E(t_1, t_2)$.
        Since, by hypothesis, each node in
        $Arg(t) = \set{Root(t)} \cup Arg(t_1) \cup Arg(t_2) - \set{Root(t_2)}$,
        is labeled, then also $Root(t)$ is labeled with, say, $x_2$.
        For the same reason, any node of $Arg(t_1)$ is labeled, too, and so, by
        induction hypothesis, $t_1$ is fully (and uniquely) labeled.
        In particular its root is labeled with, say, $x_1$.
        Since $t$ is well formed, $t_2$ cannot be the result of
        the application of a $[/I]$ rule, and since $t$ is normal,
        $t_2$ cannot be the result of the application of a
        $[\backslash I]$ rule, so its root node is an argument node of its, too.
        By hypothesis, each node in
        $Arg(t_2) - \set{Root(t_2)}$ has a type, so we can apply the following rule:
        \begin{figure}[htbp]
            \begin{center}
                \includegraphics[height=3cm]{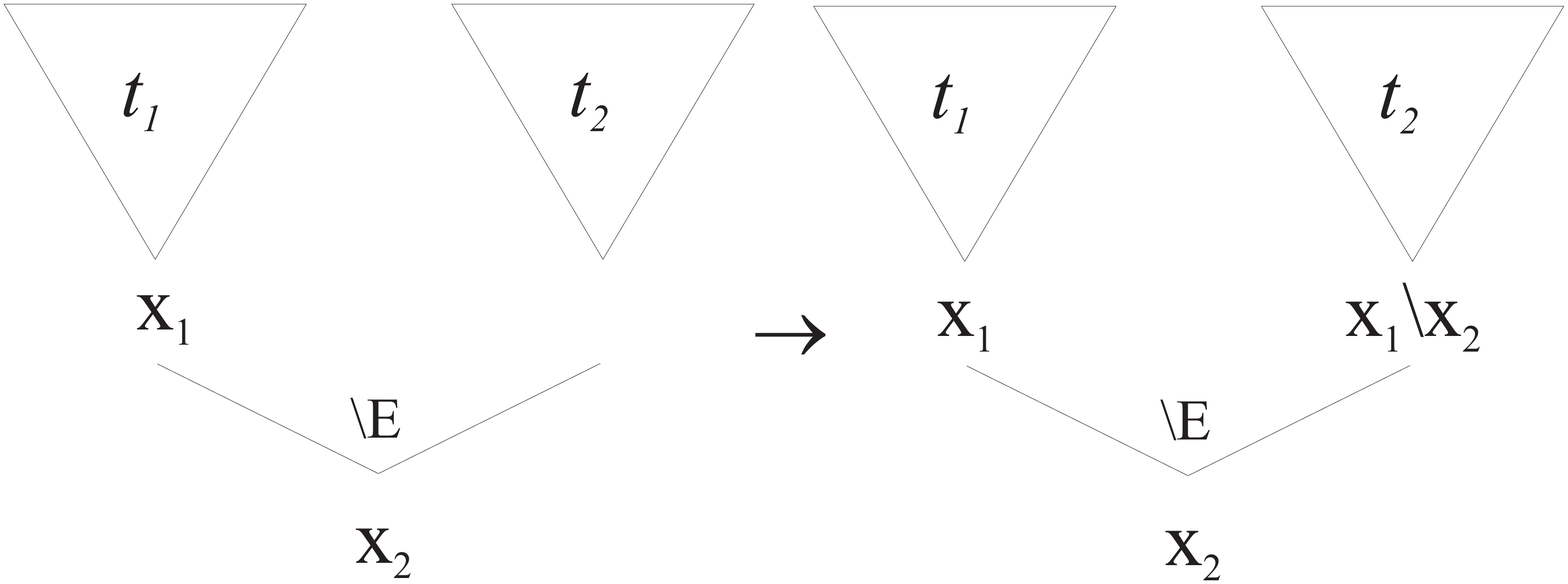}
            \end{center}
        \end{figure}\newline
        and $t_2$ has all of its argument nodes (uniquely) labeled. So, by
        induction hypothesis, its fully and uniquely labeled, and
        so is $t$.
        \item $t \equiv /E(t_1, t_2)$. Analogous to case 1.
        \item $t \equiv \backslash I(t_1)$ or $t \equiv /I(t_1)$.
        By definition, $Arg(t)=Arg(t_1)$, then by hypothesis, any
        argument node in $t_1$ is labeled. Then, by induction
        hypothesis, $t_1$ is fully (and uniquely) labeled, and since $t$ is
        well-formed, there must be at least two undischarged
        leaves in $t_1$. So $t$ can be fully labeled according, respectively,
        to the following rules:\newpage
        \begin{figure}[htbp]
            \begin{center}
                \includegraphics[height=4cm]{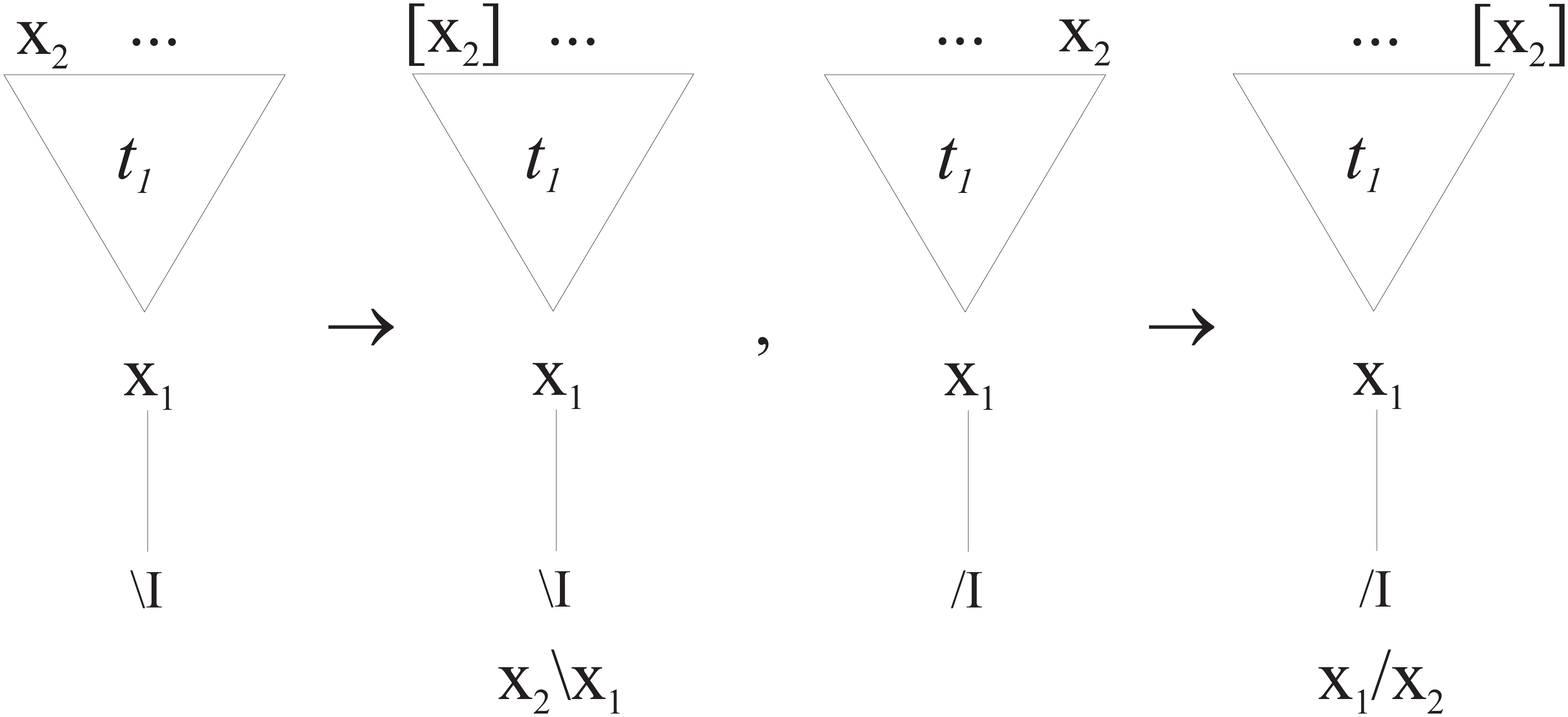}
            \end{center}
        \end{figure}
        where $x_2$ labels, respectively, the leftmost and the rightmost
        undischarged leaf.
     \end{enumerate}

     The proof of the previous proposition has implicitly defined
     an algorithm for labeling in the most general way the nodes
     of a normal form proof tree structure.
     \begin{definition}
        A {\rm principal parse} of a proof tree structure $t$ is a
        partial parse tree ${\cal T}$ of $t$, such that for any other
        partial parse tree ${\cal T}^\prime$ of $t$, there exists
        a substitution $\sigma$ such that, if a node of $t$ is
        labeled by type $A$ in ${\cal T}$, it's labeled by
        $\sigma(A)$ in ${\cal T}^\prime$.
     \end{definition}

     From the proof of proposition \ref{labelprop} it's easy to devise
     an algorithm to get a principal parse for any well formed
     normal form proof tree structure.\newline

     {\sc Principal Parse Algorithm}
     \begin{itemize}
        \item {\bf Input:} a well formed normal form proof tree structure $t$;
        \item {\bf Output:} a principal parse ${\cal T}$ of $t$ in
        a Lambek grammar $G$.
     \end{itemize}

     {\bf Step 1.} Label with distinct variables each argument
     node in $t$;\newline

     {\bf Step 2.} Compute the types for the remaining nodes
     according to the rules described in the proof of proposition
     \ref{labelprop}.\newline

     Obviously, this algorithm always terminates. If ${\cal T}$ is
     the resulting parse, we can easily prove it's principal. If ${\cal T}^\prime$
     is another parse for $t$, let's define a substitution
     $\sigma$ in the following way: for each variable $x \in
     Var(G)$, find the (unique, for construction) node in ${\cal
     T}$ labeled by $x$, and let $\sigma(x)$ be the type labeling the
     same node in ${\cal T}^\prime$. By induction on $A \in Tp(G)$
     (where $Tp(G)$ is the set of all subtypes appearing in a Lambek
     Grammar $G$), we prove that
     \begin{quote}
        if $A$ labels a node of ${\cal T}$, $\sigma(A)$
        labels the corresponding node of~${\cal T}^\prime$.
     \end{quote}

     {\it Induction Basis.} If $A \in Var$, this holds by
     definition.\newline

     {\it Induction Step.} Let $A=B\backslash C$ labels a node of
     ${\cal T}$. Then the relevant part of ${\cal T}$ must look
     like one of the following cases:\newpage

     \begin{itemize}
        \item First case:
                \begin{figure}[htbp]
                    \begin{center}
                        \includegraphics[height=2.7cm]{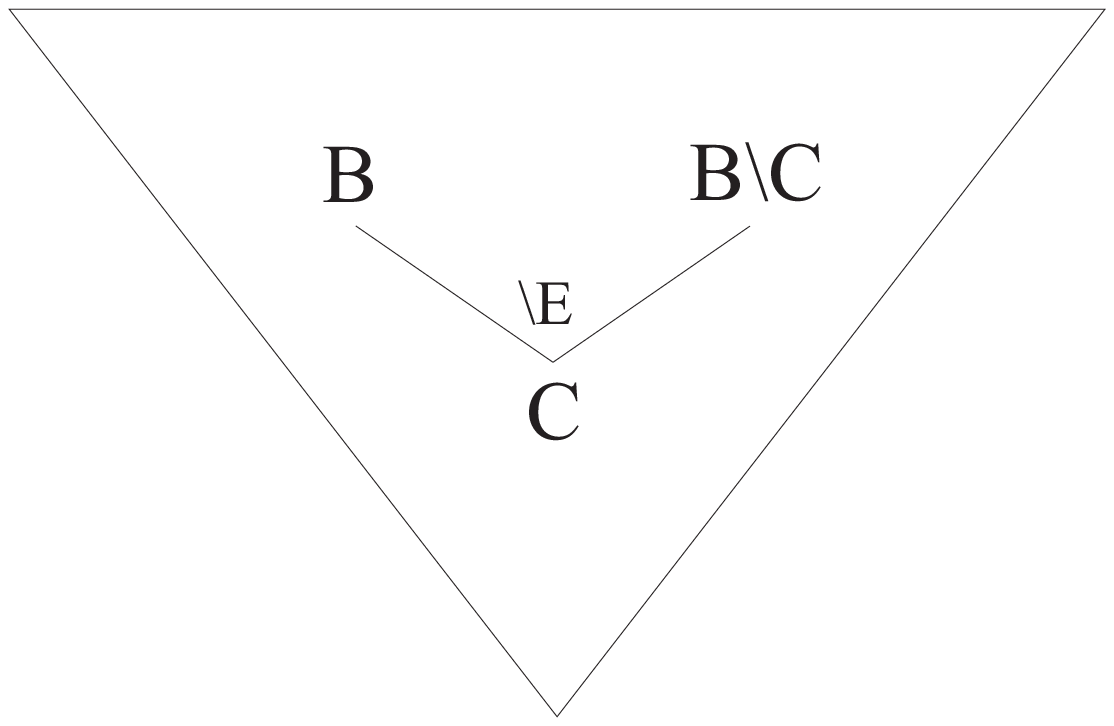}
                    \end{center}
              \end{figure}\newline
              By induction hypothesis, the corresponding part of
              ${\cal T}^\prime$ looks like:
              \begin{figure}[htbp]
                    \begin{center}
                        \includegraphics[height=2.7cm]{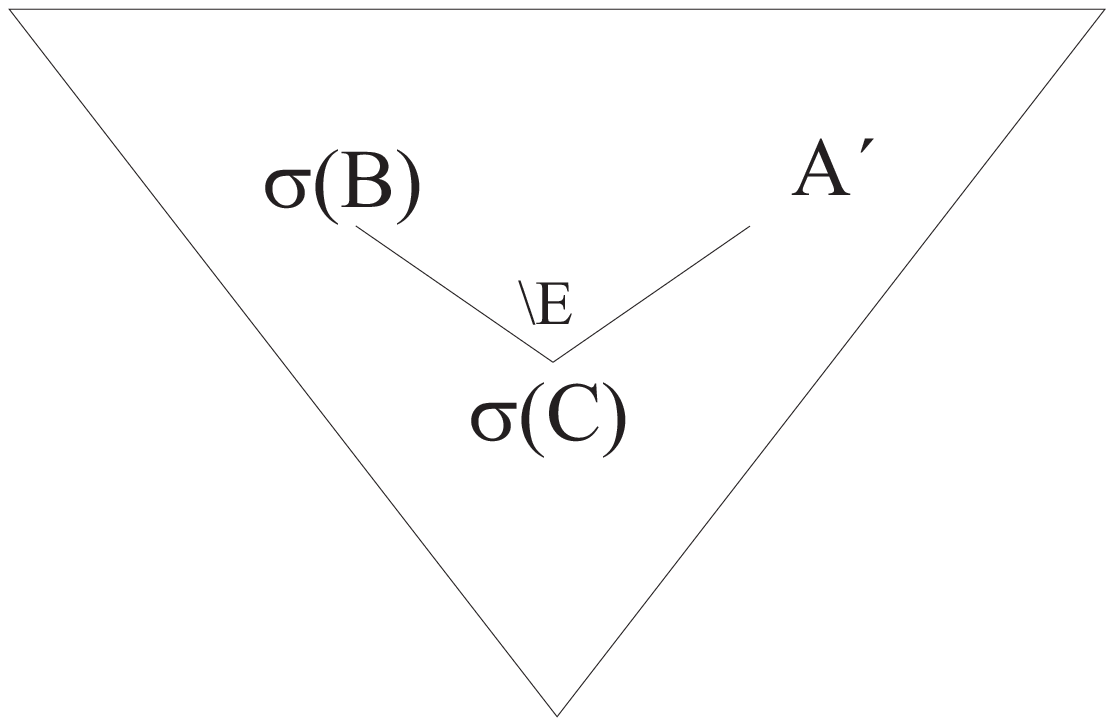}
                    \end{center}
              \end{figure}\newline
              Then $A^\prime = \sigma(B)\backslash \sigma(C) =
              \sigma(B\backslash C)= \sigma(A)$.
         \item Second case:
            \begin{figure}[htbp]
                \begin{center}
                    \includegraphics[height=2.7cm]{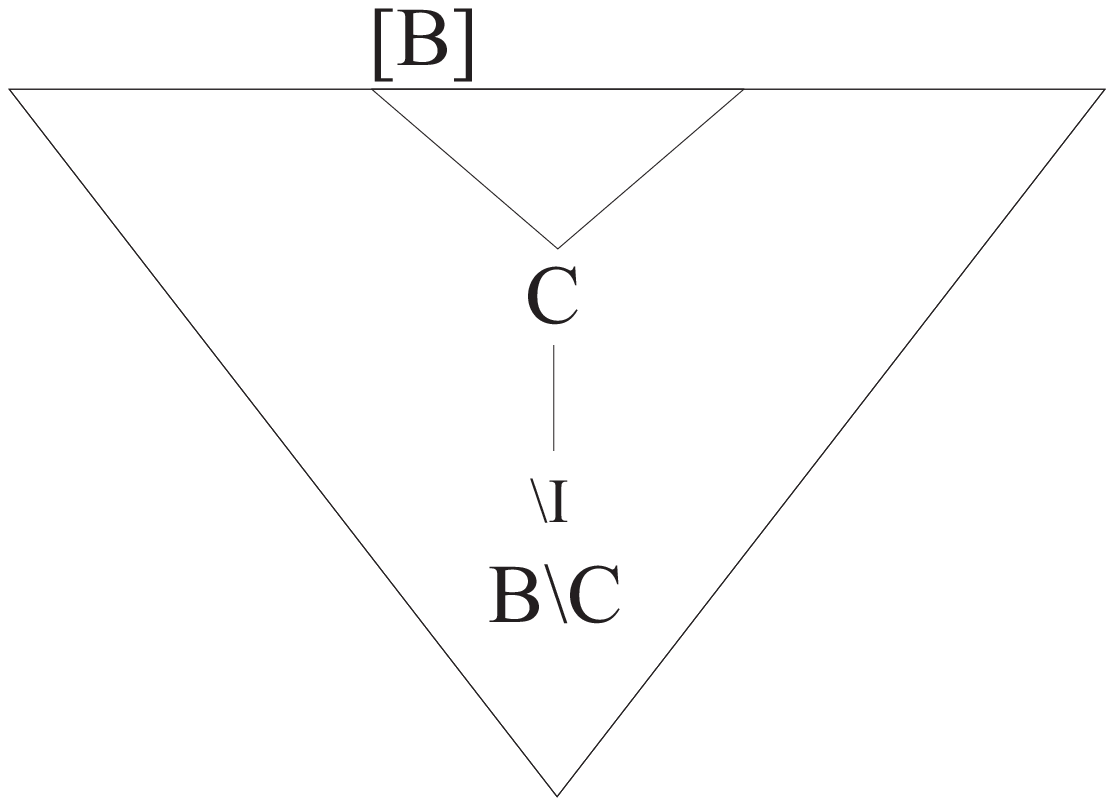}
                \end{center}
            \end{figure}\newline
            By induction hypothesis, the corresponding part of
            ${\cal T}^\prime$ looks like:
            \begin{figure}[htbp]
                    \begin{center}
                        \includegraphics[height=2.7cm]{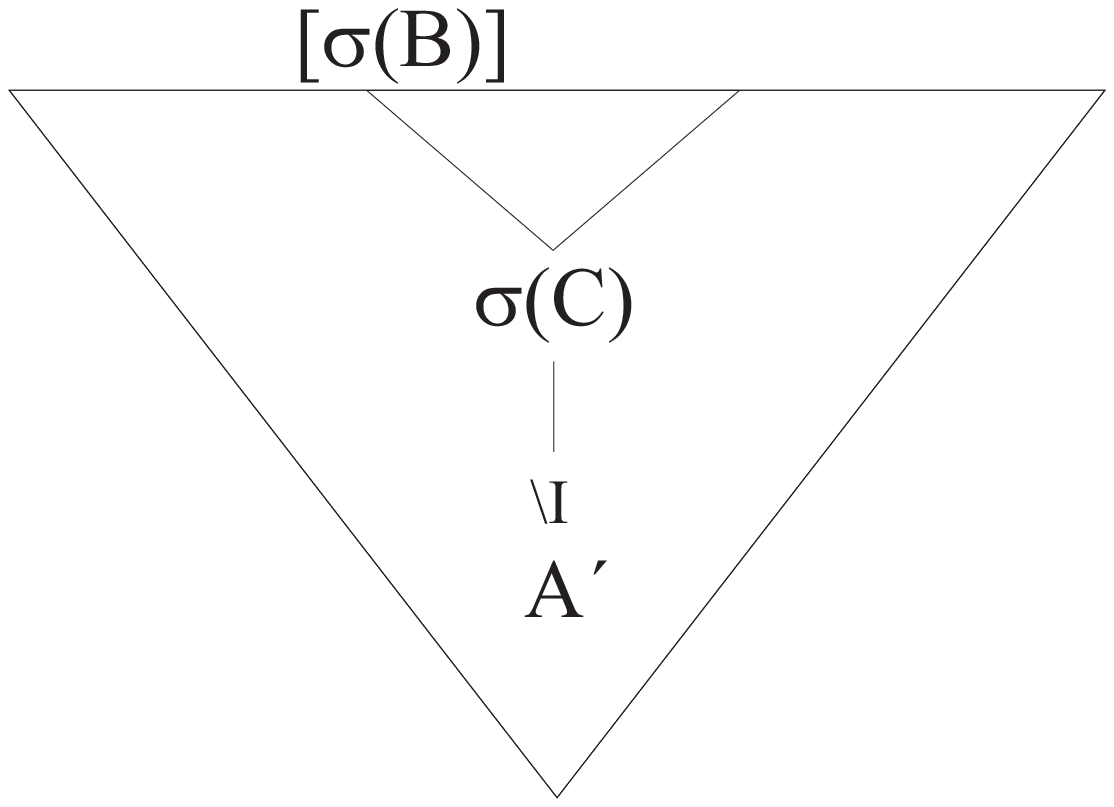}
                    \end{center}
              \end{figure}\newline
            Then $A^\prime = \sigma(B)\backslash \sigma(C) =
              \sigma(B\backslash C)= \sigma(A)$.
     \end{itemize}

     The case $A=C/B$ is entirely similar, thus completing the
     induction.

     It follows that if a node of ${\cal T}$ is labeled by $A$,
     then the corresponding node of ${\cal T}^\prime$ is labeled
     by $\sigma(A)$. That is to say, with a small abuse of
     notation, ${\cal T^\prime} = \sigma({\cal T})$.\newline

    \subsection{RLG Algorithm}
    \label{algorithmsec}
    Our algorithm (called RLG from Rigid Lambek Grammar)
    takes as its input a finite set $D$ of proof tree
    structures over a finite alphabet $\Sigma$ and returns a
    rigid Lambek grammar $G$ over the same alphabet whose
    structure language contains (properly) $D$, if it exists;
    a correct statement that there's no such a rigid Lambek grammar
    otherwise.

    Our algorithm is based on the type algorithm described in
    section \ref{argumentsec} and on the unification algorithm described
    in section \ref{UnifSec}.\newline

    {\bf \sc RLG Algorithm}.
    \begin{itemize}
        \item{{\bf input:} a finite set $D$ of proof tree structures.}
        \item{{\bf output:} a rigid Lambek grammar $G$ such that
            $D \subset \pl(G)$, if there is one.}
    \end{itemize}
    We illustrate the algorithm using the following example:
    \begin{figure}[htbp]
            \begin{center}
                    \includegraphics[height=5.5cm]{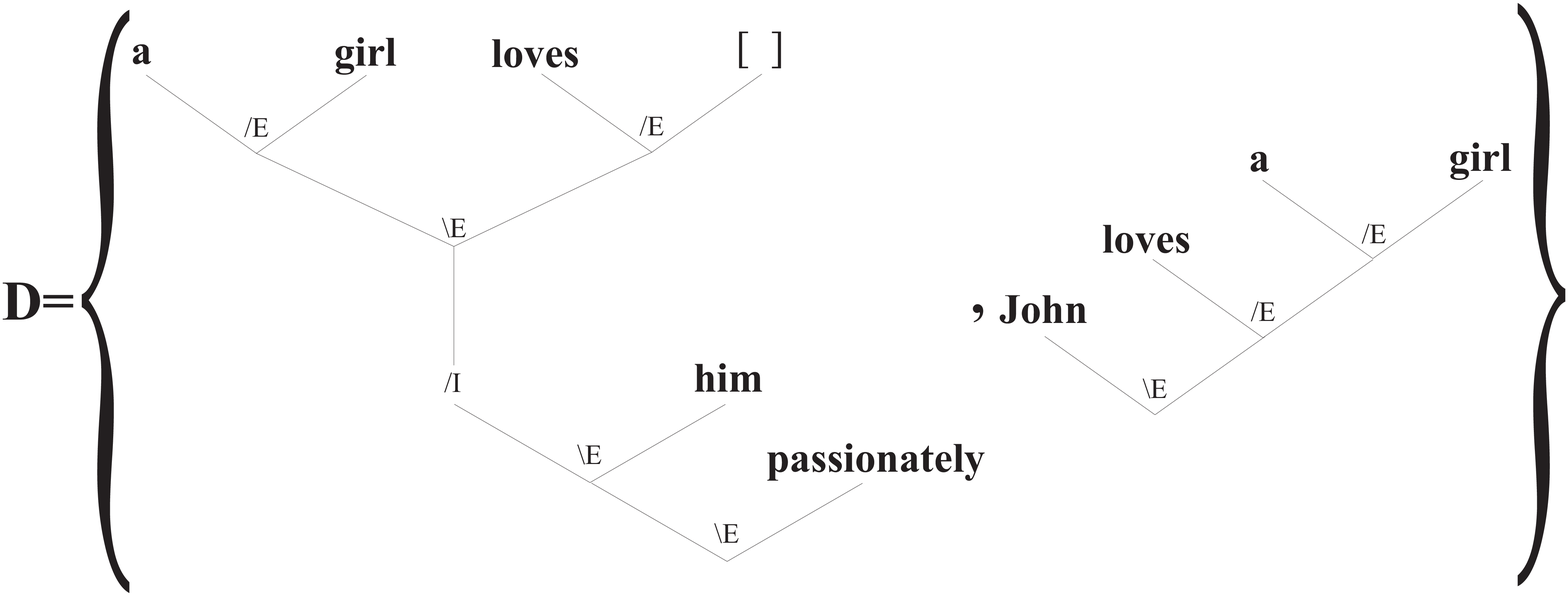}
            \end{center}
    \end{figure}\newpage

    {\bf Step 1.} Normalize all the proof tree structures in $D$,
    if they are not normal, according to the rules described in
    section \ref{rulesec}.\newline

    {\bf Step 2.} Assign a type to each node of the structure in $D$
    as follows:
    \begin{enumerate}
        \item Assign $s$ to each root node.
        \item Assign distinct variables to the argument nodes.
            \begin{figure}[htbp]
                \begin{center}
                    \includegraphics[height=5.7cm]{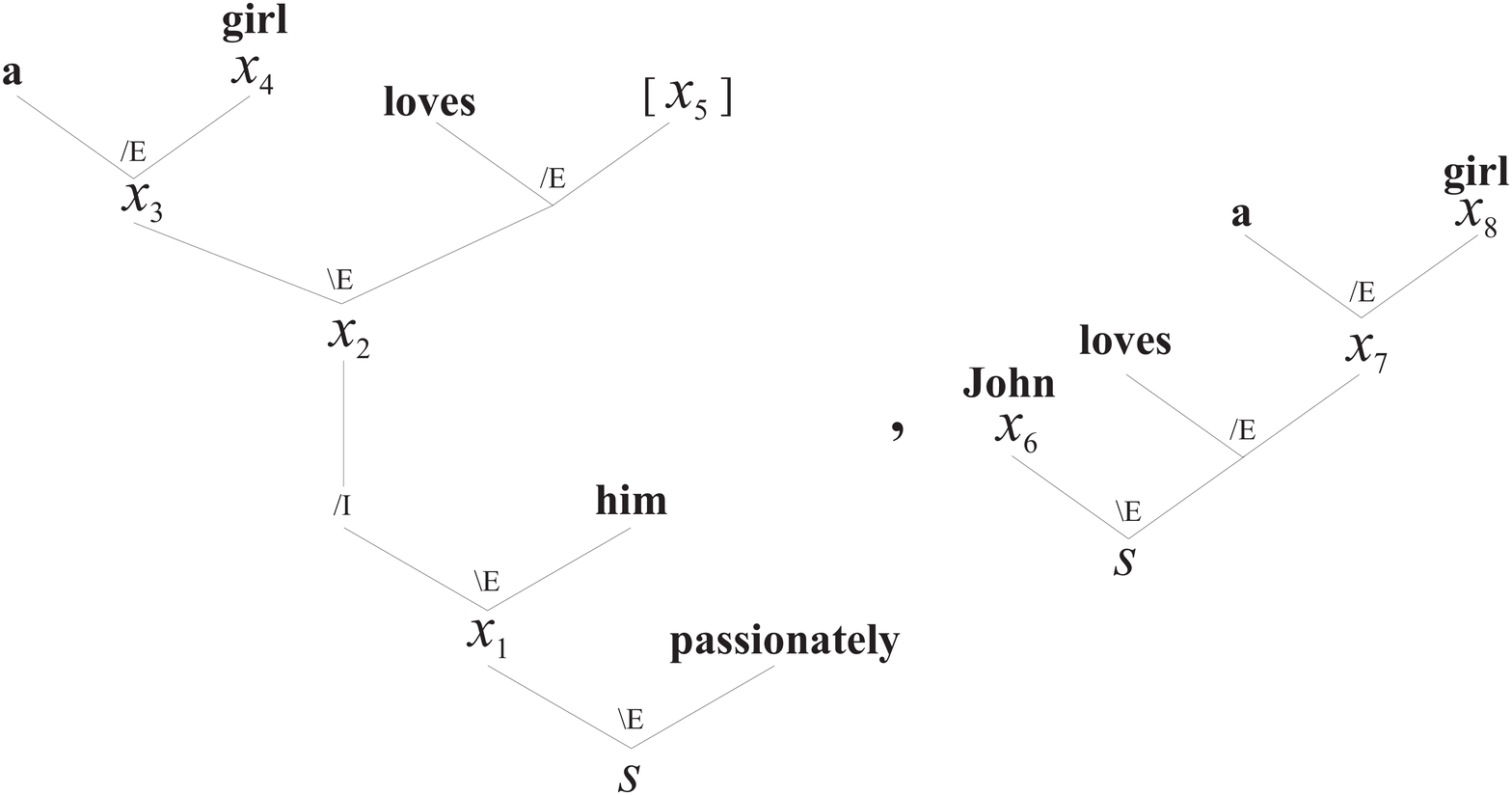}
                \end{center}
            \end{figure}
        \item Compute types for the remaining nodes according to the rules
        described in proposition \ref{labelprop}.
            \begin{figure}[htbp]
                \begin{center}
                    \includegraphics[height=5.7cm]{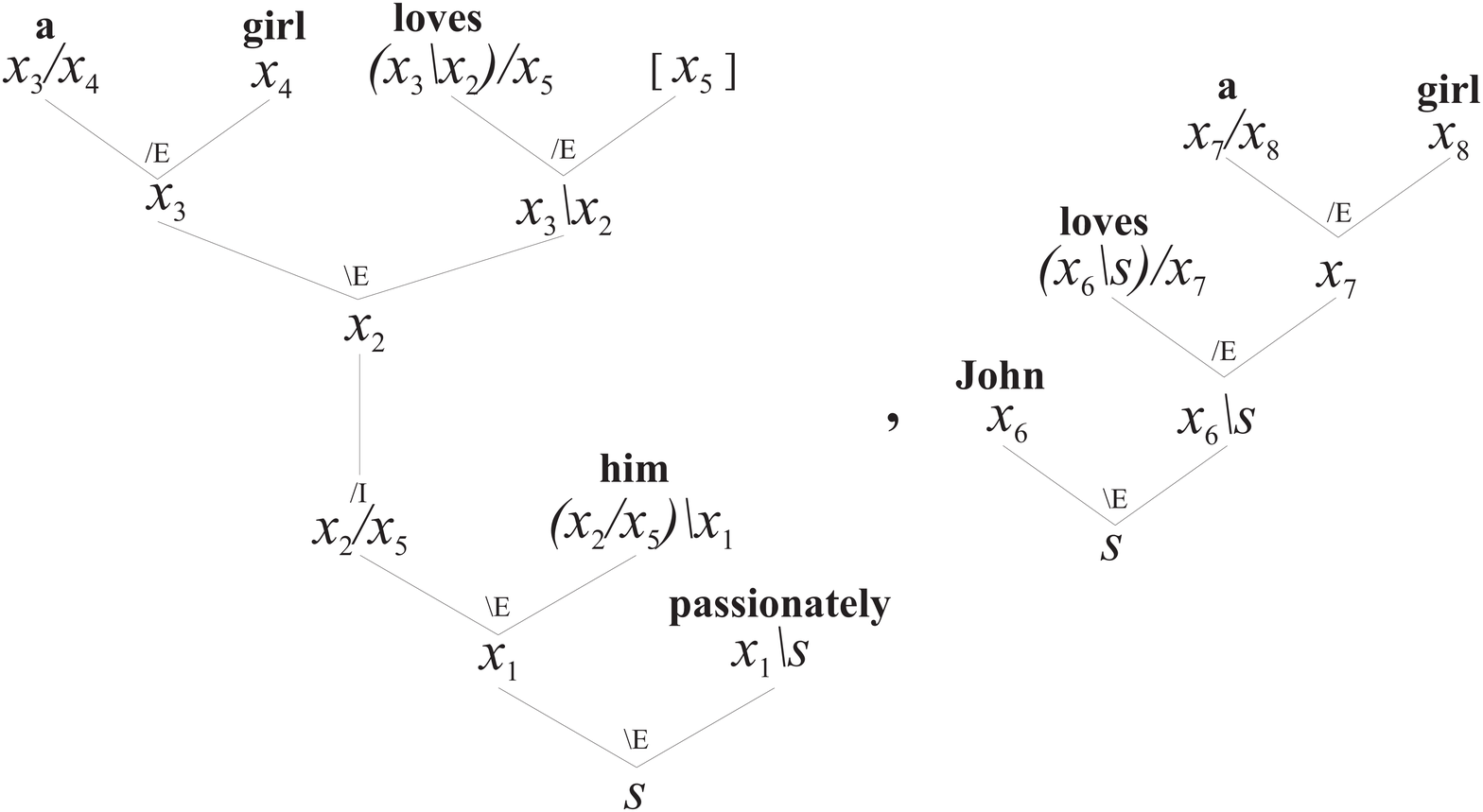}
                \end{center}
            \end{figure}
    \end{enumerate}\newpage

    {\bf Step 3.} Collect the types assigned to the leaf nodes into a
    grammar $GF(D)$ called the {\it general form} induced by $D$. In
    general, $GF(D): c \mapsto A$ if and only if the previous step
    assigns $A$ to a leaf node labeled by symbol $c$.
    \begin{eqnarray*}
        GF(D):  {\bf passionately} &\mapsto& x_1\backslash s\\
                {\bf him} &\mapsto& (x_2/x_5)\backslash x_1\\
                {\bf a} &\mapsto& x_3/x_4, x_7/x_8\\
                {\bf girl} &\mapsto& x_4, x_8\\
                {\bf loves} &\mapsto& (x_3\backslash x_2)/x_5,
                    (x_6\backslash s)/x_7\\
                {\bf John} &\mapsto& x_6
    \end{eqnarray*}

    {\bf Step 4.} Unify the types assigned to the same symbol. 
    Let ${\cal A}=\set{\set{A\ |\ GF(D): c \mapsto A}\ |\ c \in dom(GF(D))}$,
    and compute $\sigma = mgu({\cal A})$. The algorithm fails if
    unification fails.
    \begin{displaymath}
        \sigma = \set{x_7\mapsto x_3, x_8\mapsto x_4, x_6 
	\mapsto x_3, x_2 \mapsto s, x_5 \mapsto x_3}
    \end{displaymath}
    \newline

    {\bf Step 5.} Let $RLG(D)= \sigma[GF(D)]$.
    \begin{eqnarray*}
        RLG(D):  {\bf passionately} &\mapsto& x_1\backslash s\\
                {\bf him} &\mapsto& (s/x_3)\backslash x_1\\
                {\bf a} &\mapsto& x_3/x_4\\
                {\bf girl} &\mapsto& x_4\\
                {\bf loves} &\mapsto& (x_3\backslash s)/x_3\\
                {\bf John} &\mapsto& x_3
    \end{eqnarray*}

    Our algorithm is based on the ``principal parse algorithm''
    described in the previous section, which has been proved to be
    correct and terminate, and the unification algorithm described
    in section \ref{UnifSec}. The result is, intuitively,  the
    most general rigid Lambek Grammar which can generate all the
    proof tree structures appearing in the input sequence.

    \subsection{Properties of RLG}
    \label{algorithm}
    In the present section we prove some properties of the RLG
    algorithm that will be helpful to study its behaviour in the
    limit.

    The following lemma is almost trivial but it will play an important
    role in the convergence proof for the RLG algorithm. It simply states
    that the tree language of the grammar inferred just after 
    the labeling of the structures properly contains the sample structures. 

    \begin{lemma}
        \label{LemmaBonato1}
        Let $D$ be the input set of proof tree structures for the
        RLG algorithm. Then the set of the proof tree structures
        generated by the `general form' grammar contains properly
        $D$. That is, $D \subset \pl(GF(D))$.
    \end{lemma}
    {\it Proof.} Let $D = \set{T_1, \ldots, T_n}$. The labeling of
    the nodes of the structures in $D$ that precedes the
    construction of $GF(D)$ in fact forms a parse tree ${\cal
    P}_i$ of $GF(D)$ for each structure $T_i$ in $D$. This shows
    $D \subseteq PL(GF(D))$. The proper inclusion follows
    trivially from the fact that $D$ is by hypothesis a finite
    set, while $\pl(G)$, the set of proof tree structures generated by a Lambek grammar
    $G$, is always infinite.

    \begin{lemma}
        \label{lastminutelemma}
        Each variable $x \in Var(GF(D))$ labels a unique node in a
        unique parse tree of~$D$.
    \end{lemma}
    {\it Proof}. Obviously, by construction, if $x \in Var(GF(D))$, then there
    must be an $i \in \naturals$ such that $x$ labels one of the
    nodes of a parse tree ${\cal P}_i$. Since, by construction, for
    each $i \not = j$ the sets of variables that label ${\cal
    P}_i$ are disjoint, $x$ appears in one and only one ${\cal
    P}_i$. Besides, since variables are assigned only during the first
    phase of the type-assignment process of our algorithm, again
    by construction each variable labels only one node in the
    deduction tree.\newline

    The following lemma makes explicit the relation between the
    grammar inferred just after the labeling of the structures in the
    algorithm RLG, and the structure language of the rigid grammar
    we are trying to infer. 
    \begin{lemma}
        \label{LemmaBonato2}
        Let $D$ be a finite set of proof tree structures. Then, for
        any Lambek grammar $G$, the following are equivalent:
        \renewcommand{\labelenumi}{{\rm (\roman{enumi})}}
        \begin{enumerate}
            \item $D \subseteq \pl(G)$
            \item There is a substitution $\sigma$ such that
            $\sigma[GF(D)] \subseteq G$.
        \end{enumerate}
    \end{lemma}
    {\it Proof.} (ii)$\Rightarrow$(i). Suppose there is a substitution
    $\sigma$ such that $\sigma[GF(D)] \subseteq G$. Then, from
    proposition \ref{bonatoprop}, we have that $\pl(GF(D))\subseteq \pl(G)$.
    This, together with lemma \ref{LemmaBonato1} proves (i).

    (i) $\Rightarrow$(ii). Let $D=\set{T_1, \ldots, T_n}$ and let
    ${\cal P}_i$ be $GF(D)$'s parse of $T_i$ for $1 \le i \le n$.
    Assume $D \subseteq \pl(G)$. Then $G$ has a parse ${\cal Q}_i$
    of each $T_i$. Define a substitution $\sigma$ as follows: for
    each variable $x \in Var(GF(D))$, find a (unique, due to lemma \ref{lastminutelemma})
    ${\cal P}_i$ that contains a (unique, again due to lemma \ref{lastminutelemma})
    node labeled by $x$, and let $\sigma(x)$ be the type labeling the corresponding node of
    ${\cal Q}_i$. We show that
    \begin{quote}
        if $A$ labels a node of some ${\cal P}_i$, then
        $\sigma(A)$ labels the corresponding node of ${\cal Q}_i$.
    \end{quote}

    {\it Proof.} By induction on $A\in Tp(GF(D)) = \set{T\ 
	|\ T\ \mbox{is a subtype of some}\ B\in range(GF(D))}$):\newline

    {\it Induction basis}. If $A \in Var$, this holds by
    definition. If $A = t$, then any node labeled by $A$ in
    $\set{{\cal P}_1, \ldots, {\cal P}_n}$ is the root node of some
    ${\cal P}_i$. Since ${\cal Q}_i$ is a parse tree of $G$, the
    root node of ${\cal Q}_i$ must be labeled by $t$.\newline

    {\it Induction step}. Let $A=B\backslash C$ labels a node of
    ${\cal P}_i$. Then the relevant part of ${\cal P}_i$ must look
    like one of the two following cases:\newpage

    \begin{itemize}
        \item First case

            \begin{figure}[htbp]
                \begin{center}
                    \includegraphics[height=3.5cm]{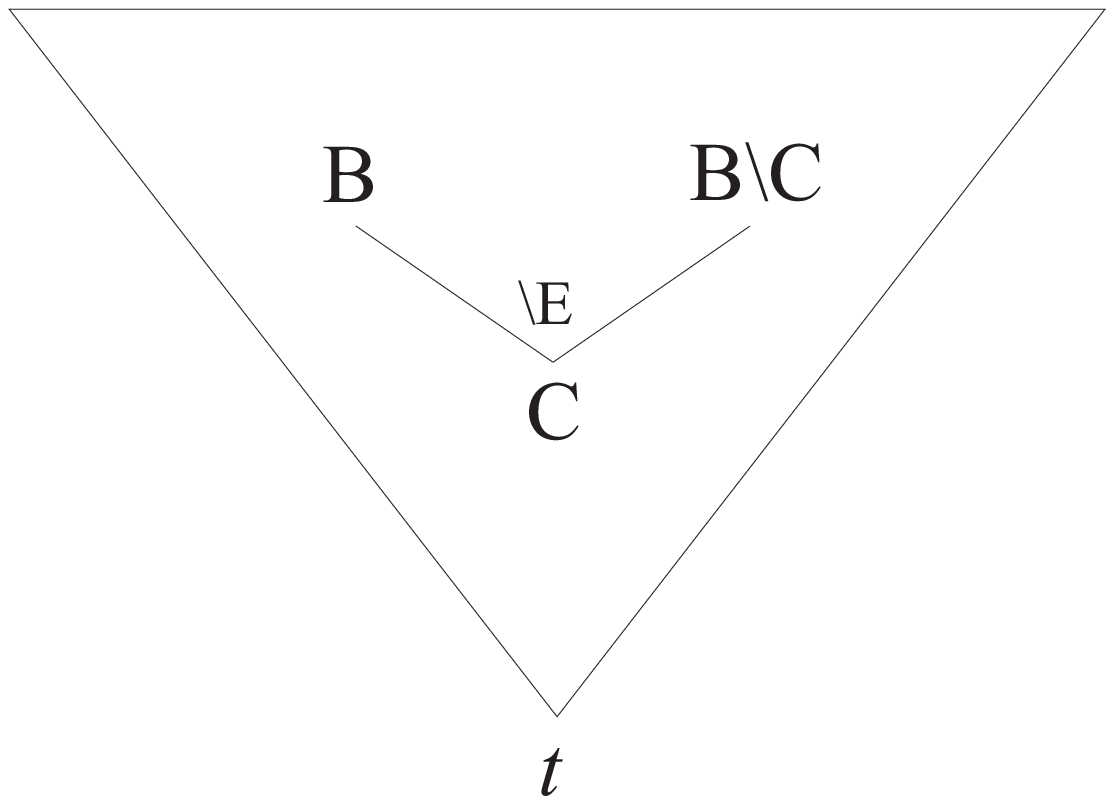}
                \end{center}
            \end{figure}
            By induction hypothesis, the corresponding part of
            ${\cal Q}_i$ looks like:
            \begin{figure}[htbp]
            \begin{center}
                    \includegraphics[height=3.5cm]{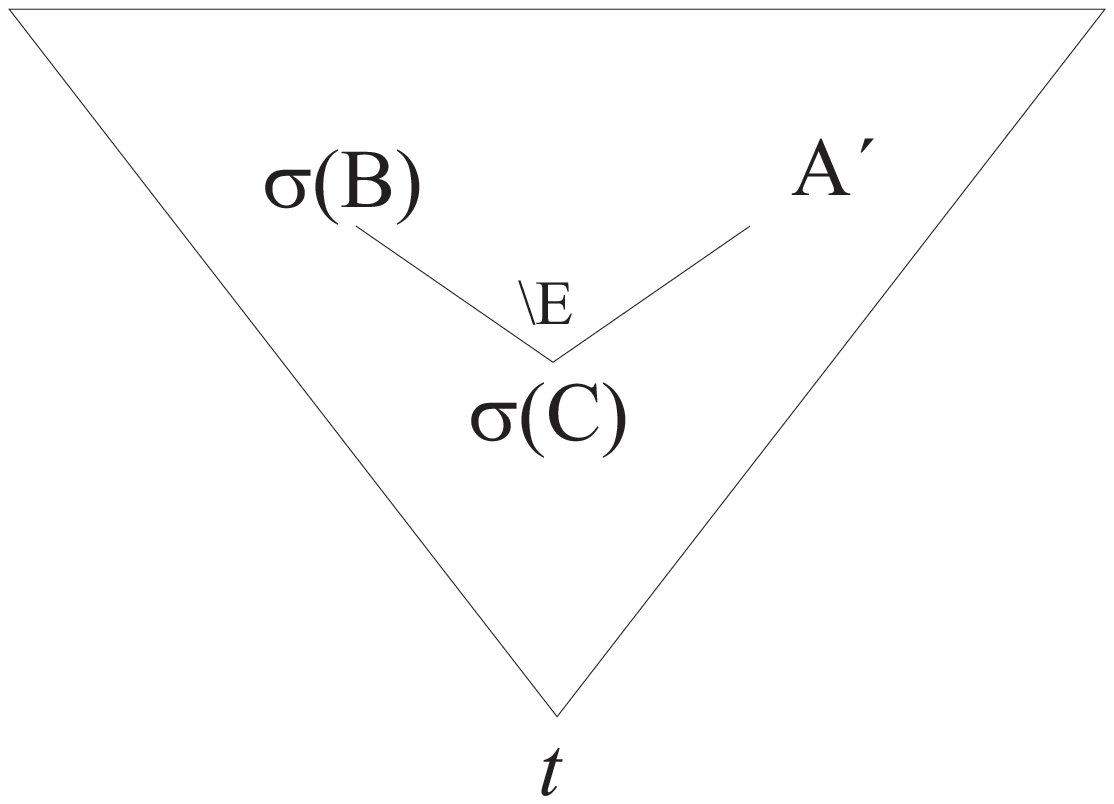}
            \end{center}
    \end{figure}\newline

    Then $A^\prime = \sigma(B)\backslash \sigma(C) = \sigma(B\backslash C) =
    \sigma(A)$.\newline

    \item Second case

    \begin{figure}[htbp]
           \begin{center}
                    \includegraphics[height=4cm]{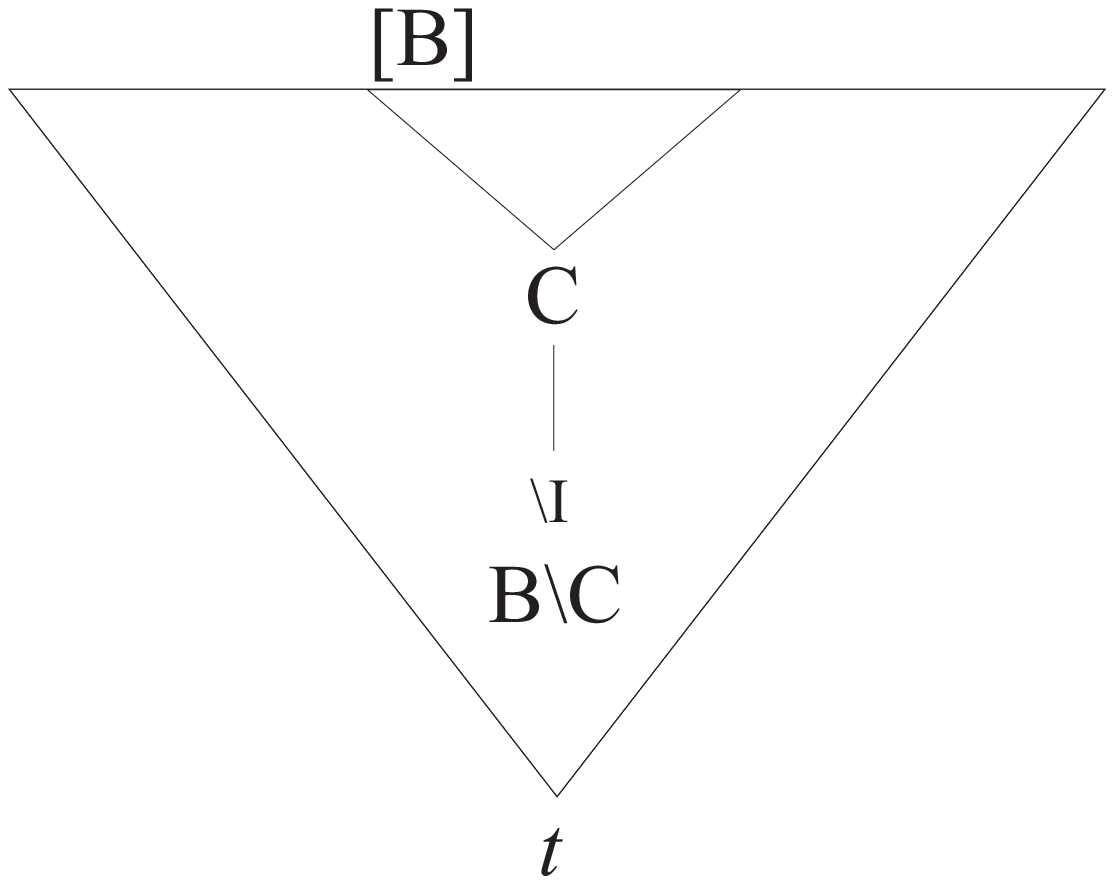}
           \end{center}
    \end{figure}\newpage
    By induction hypothesis, the corresponding part of
    ${\cal Q}_i$ looks like:
    \begin{figure}[htbp]
           \begin{center}
                  \includegraphics[height=4cm]{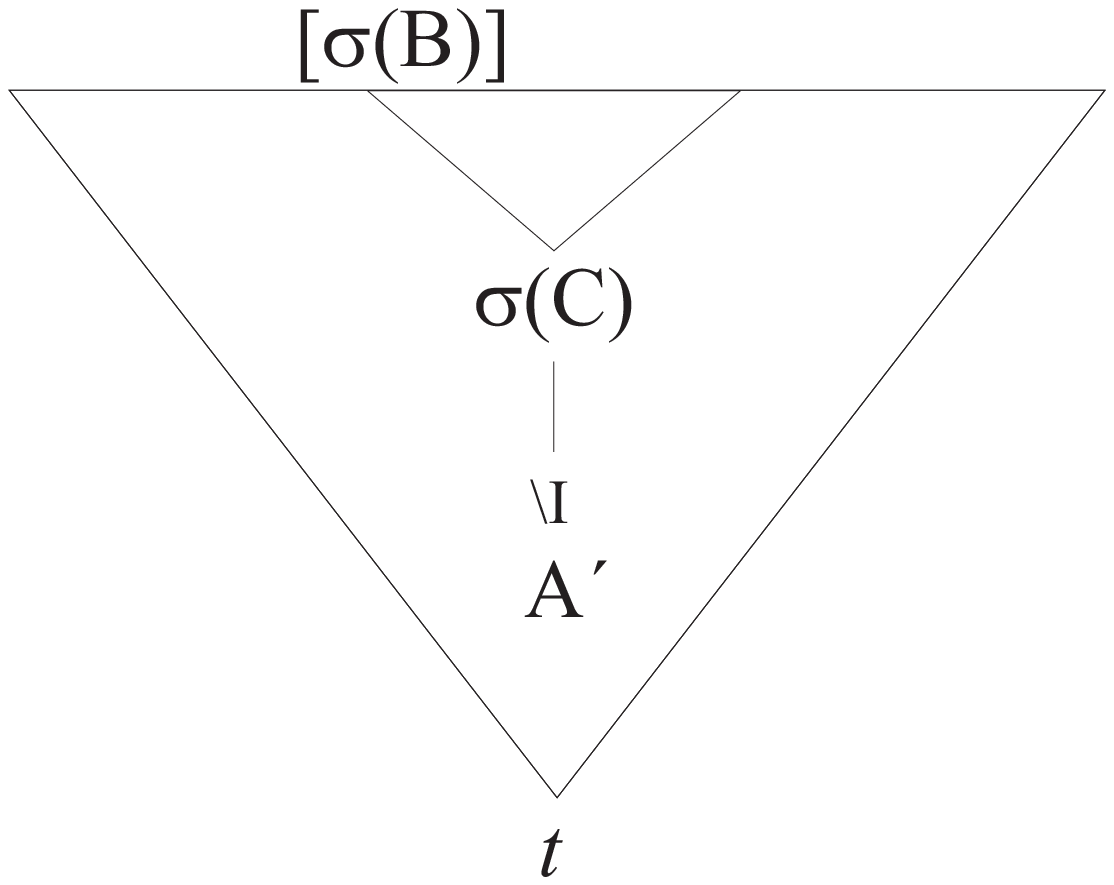}
           \end{center}
    \end{figure}

    Then again $A^\prime = \sigma(B)\backslash \sigma(C) = \sigma(B\backslash C) =
    \sigma(A)$.
    \end{itemize}
    The case $A = C/B$ is entirely similar, thus completing the
    induction.
    It follows that if $GF(D): c \mapsto A$, then $G: c \mapsto
    \sigma(A)$. Therefore, $\sigma[GF(D)] \subseteq~G$.\newline

    The following proposition establishes an ``if and only if'' relation
    between the inclusion of our set of positive samples $D$ in a tree
    language generated by a rigid grammar $G$ and the successful termination
    of the RLG algorithm when it has $D$ as its input set. Even more, we have
    that the rigid grammar inferred by the algorithm is not ``larger'' than
    the rigid grammar~$G$.

    \begin{proposition}
        \label{PropBonato3}
        Let $D$ be a finite set of proof tree structures. Then, for
        any rigid grammar $G$, the following are equivalent:
        \renewcommand{\labelenumi}{{\rm (\roman{enumi})}}
        \begin{enumerate}
            \item $D \subseteq \pl(G)$;
            \item $RLG(D)$ exists and $RLG(D)\sqsubseteq G$
                (equivalently, there is a substitution $\tau$ such
                that $\tau[RLG(D)]\subseteq G$).
        \end{enumerate}
    \end{proposition}
    {\it Proof}. (ii) $\Rightarrow$ (i) follows from lemma
    \ref{LemmaBonato2} and the fact that $RLG(D)$ is a substitution
    instance of $GF(D)$.

    (i) $\Rightarrow$ (ii). Assume that $G$ is a rigid grammar
    such that $D \subseteq PL(G)$. By lemma \ref{LemmaBonato2}
    there is a substitution $\sigma$ such that
    $\sigma[GF(D)] \subseteq G$. Since $G$ is a rigid grammar,
    $\sigma[GF(D)]$ is also a rigid grammar. Then $\sigma$ unifies
    the family ${\cal A} = \set{\set{A\ |\ GF(D): c \mapsto A}\ |\ c \in
    dom(GF(D))}$. This means that $RLG(D)$ exists and
    $RLG(D)=\sigma_0[GF(D)]$, where $\sigma_0=mgu({\cal A})$. Then
    there is a substitution $\tau$ such that $\sigma = \tau \circ
    \sigma_0$. Therefore, $\tau[RLG(D)]=\tau[\sigma_0[GF(D)]] =
    (\tau \circ \sigma_0)[GF(D)] = \sigma[GF(D)]$. By assumption,
    $\sigma[GF(D)]\subseteq G$, so $\tau[RLG(D)]\subseteq
    G$.\newline

    \begin{corollary}
        \label{CorBonato}
        Let $D_1$ and $D_2$ be two finite sets of proof tree
        structures such that $D_1 \subseteq D_2$. If $RLG(D_2)$
        exists, $RLG(D_1)$ also exists and $RLG(D_1)\sqsubseteq
        RLG(D_2)$ and $\pl(RLG(D_1))\subseteq \pl(RLG(D_2))$.
    \end{corollary}
    {\it Proof}. Immediate from proposition \ref{PropBonato3},
    noting that if $D_1 \subseteq D_2$, then $\set{G \in {\cal
    G}_{rigid}\ |\ D_1 \subseteq \pl(G)} \supseteq \set{G \in {\cal
    G}_{rigid}\ |\ D_2 \subseteq \pl(G)}$.\newline

    \begin{definition}
        Let $\varphi_{RLG}$ be the learning function for the
        grammar system $\seq{G_{rigid}, \Sigma^P, \pl}$ defined
        as follows:\footnote{Recall that the symbol $\simeq$ means
        that either both sides are defined and are equal, or else
        both sides are undefined}
        \begin{displaymath}
            \varphi_{RLG}(\seq{T_0, \ldots, T_n}) \simeq
            RLG(\set{T_0, \ldots, T_n}).
        \end{displaymath}
    \end{definition}
    \ \newline

    Thanks to previous propositions and lemmas we are able to prove
    the convergence for the RLG algorithm: 
    \begin{theorem}
        $\varphi_{RLG}$ learns ${\cal G}_{rigid}$ from structures.
    \end{theorem}
    {\it Proof}. We prove that $\varphi_{RLG}$ learns the class of
    rigid Lambek grammars from proof tree structures.

    Let $G$ be any rigid Lambek grammar and let $\seq{T_i}_{i \in
    \naturals}$ be an infinite sequence enumerating $\pl(G)$. For
    each $i \in \naturals$, $\set{T_0, \ldots, T_i} \subseteq
    \pl(G)$, so by proposition \ref{PropBonato3} $\varphi_{RLG}
    (\seq{T_0, \ldots, T_i}) = RLG(\set{T_0, \ldots, T_i})$
    is defined and
    \begin{displaymath}
        \varphi_{RLG}(\seq{T_0, \ldots, T_i}) \sqsubseteq
        \varphi_{RLG}(\seq{T_0, \ldots, T_{i+1}}),
    \end{displaymath}
    by corollary \ref{CorBonato}, and
    \begin{displaymath}
        \varphi_{RLG}(\seq{T_0, \ldots, T_i})
        \sqsubseteq G.
    \end{displaymath}
    Since, by corollary \ref{sizecorollary},
    there are only finitely many Lambek grammars
    $G^{\prime\prime} \sqsubseteq G$, $\varphi_{RLG}$ must
    converge on $\langle T_i\rangle_{i \in \naturals}$ to some
    $G^\prime$. Then $\pl(G)=\{T_i \ |\ i \in \naturals\} \subseteq
    \pl(G^\prime)$. Since $G^\prime \sqsubseteq G$, by proposition
    \ref{bonatoprop}, $\pl(G^\prime)\subseteq \pl(G)$. Therefore,
    $\pl(G^\prime)=\pl(G)$.\newline

    When RLG is applied successively to a sequence of increasing
    set of proof tree structures $D_0 \subset D_1 \subset D_2
    \subset \cdots$, it is more efficient to make use of the
    previous value RLG$(D_{i-1})$ to compute the current value
    RLG$(D_i)$.

    \begin{definition}
        If $G$ is a rigid Lambek grammar and $D$ is a finite set
        of proof tree structures, then let
        \begin{displaymath}
            RLG^{(2)}(G, D) \simeq G \sqcup RLG(D).
        \end{displaymath}
    \end{definition}

    \begin{lemma}
        \label{lemmabonato3}
        If $D_1$ and $D_2$ are two finite sets of proof tree
        structures,
        \begin{displaymath}
            RLG^{(2)}(RLG(D_1), D_2) \simeq RLG(D_1 \cup D_2).
        \end{displaymath}
    \end{lemma}
    {\it Proof.} (See \cite{kanazawa98}). Suppose that $RLG^{(2)}(RLG(D_1),
    D_2)$ is defined. By lemma \ref{rigidlemma}, $RLG(D_1)
    \sqsubseteq RLG^{(2)}(RLG(D_1), D_2)$ and $RLG(D_2)=RLG^{(2)}(RLG(D_1),
    D_2)$. This implies that $D_1 \cup D_2 \subseteq \pl(RLG^{(2)}(RLG(D_1),
    D_2))$, so by proposition \ref{PropBonato3}, $RLG(D_1 \cup
    D_2)$ exists and $RLG(D_1 \cup D_2)\sqsubseteq RLG^{(2)}(RLG(D_1),
    D_2)$.

    Suppose now that $RLG(D_1 \cup D_2)$ is defined. By corollary
    \ref{CorBonato}, $RLG(D_1)$ and $RLG(D_2)$ exist and $RLG(D_1)
    \sqsubseteq RLG(D_1 \cup D_2)$ and $RLG(D_2) \sqsubseteq
    RLG(D_1 \cup D_2)$. Then $RLG(D_1 \cup D_2)$ is an upper bound
    of $\set{RLG(D_1), RLG(D_2)}$. By proposition \ref{rigidprop},
    $RLG(D_1) \sqcup RLG(D_2) = RLG^{(2)}(RLG(D_1), D_2)$ exists
    and $RLG^{(2)}(RLG(D_1), D_2)\sqsubseteq RLG(D_1 \cup D_2)$.

    Thus it has been proved that if one of $RLG^{(2)}(RLG(D_1),
    D_2)$ and $RLG(D_1 \cup D_2)$ is defined the other is defined
    and they are equal.

    \begin{proposition}
        $\varphi_{RLG}$ has the following properties:
        \renewcommand{\labelenumi}{{\rm (\roman{enumi})}}
        \begin{enumerate}
            \item $\varphi_{RLG}$ learns ${\cal G}_{rigid}$
            prudently.
            \item $\varphi_{RLG}$ is responsive and consistent on
            ${\cal G}_{rigid}$.
            \item $\varphi_{RLG}$ is set-driven.
            \item $\varphi_{RLG}$ is conservative.
            \item $\varphi_{RLG}$ is monotone increasing.
            \item $\varphi_{RLG}$ is incremental.
        \end{enumerate}
    \end{proposition}
    {\it Proof.}
        \renewcommand{\labelenumi}{{\rm (\roman{enumi})}}
        \begin{enumerate}
            \item Since $range(\varphi_{RLG}) \subseteq {\cal
            G}_{rigid}$, $\varphi_{RLG}$ learns ${\cal G}_{rigid}$
            prudently.
            \item If $D \subseteq L$ for some $L\in {\cal
            PL}_{rigid}$, then by proposition \ref{PropBonato3}
            $RLG(D)$ exists and by lemma \ref{LemmaBonato2}
            $D\subseteq \pl(RLG(D))$. This means that
            $\varphi_{RLG}$ is responsive and consistent on ${\cal
            G}_{rigid}$.
            \item $\varphi_{RLG}$ is set-driven by definition.
            \item Let $T\in \pl(RLG(D))$. Then $D \cup \set{T}
            \subset \pl(RLG(D))$. By proposition
            \ref{PropBonato3}, $RLG(D \cup \set{T})$ exists and
            $RLG(D\cup \set{T}) \sqsubseteq RLG(D)$. By
            corollary \ref{CorBonato} we have also
            $RLG(D)\sqsubseteq RLG(D \cup \set{T})$. This shows
            that $\varphi_{RLG}$ is conservative.
            \item Trivial from corollary \ref{CorBonato}.
            \item Define a computable function $\psi: {\cal G}_{rigid} \times
            \Sigma^P \rightarrow {\cal G}_{rigid}$ as follows:
            \begin{displaymath}
                \psi(G, T) \simeq \left \{
                                       \begin{array}{ll}
                                            RLG^{(2)}(G, \set{T})
                                            & \mbox{if } G\in {\cal
                                            G}_{rigid}\mbox{ and }
                                            RLG(\set{T})\mbox{ is
                                            defined,}\\
                                            \mbox{undefined} &
                                            \mbox{otherwise.}
                                       \end{array}
                                \right.
            \end{displaymath}
            Then by lemma \ref{lemmabonato3}, $\varphi_{RLG}(\seq{T_0, \ldots,
            T_{i+1}})\simeq\psi(\varphi_{RLG}(\seq{T_0, \ldots, T_i}),
            T_{i+1})$.
        \end{enumerate}

%% file: conclusion_report.tex
\section{Conclusion and Further Research}

This work aims at making a further step in the direction of bridging 
the gap which still separates any formal/computational theory of
learning from a meaningful formal linguistic theory.

We have introduced the basic notions of Formal Learnability Theory
as first formulated by E.M. Gold in 1967, and of Lambek Grammars, which
appeared for the first time in an article of 1958.

The former, which is one of the first completely formal descriptions
of the process of grammatical inference, after an initial skepticism about
its effective applicability, is at present to object of a renewed interest
due to some meaningful and promising learnabiluty results.

Even the latter, long neglected by the linguistic community, is experiencing
a strong renewed interest as a consequence of recent linguistics achievements
which point at formal grammars completely lexicalized, as Lambek grammars are.
Even if they're still far from being the ultimate formal device for the
formalization of human linguistics competence, they're universally looked at
as a promising tool for further developments of computational linguistics.

In the present work we've drawn the attention to a particular class 
of Lambek grammars called {\it rigid Lambek grammars}, and we've proved 
that they are learnable in Gold's framework from a structured input. We've 
used most recent results by Hans-Joerg Tiede for formally define our
notion of {\it structure} for a sentence: he has recently proved that 
the proof tree language generated by a Lambek grammar strictly contains
the tree language generated by context-free grammars. His notion of a
proof as the grammatical structure  of a sentence in a categorial grammar
is also useful in providing a natural support to a Montagovian semantics 
for that sentence. Therefore, our choice for a structured input for 
our learning algorithm in the form of proof tree structures is not 
gratuitous, but it's coherent with the mainstream of (psycho-)linguistics
theories about first language learning which stress the importance
of providing the learner with informatioannly and semantically rich 
input in the process of her language acquisition.

We believe it to be a partial but meaningful result, which once
more shows how versatile and powerful can be this learning theory,
once neglected because it was widely held that it couldn't but
account for the learnability of most trivial classes of grammars.\newline

Much is left to be done along many directions. First of all, there's 
still no real theory of rigid, or k-valued, Lambek grammars: we still
know very few formal properties of such grammars which seem to have
an undisputable linguistic interest. We still lack, for example,
a hierarchy theorem for languages generated by k-valued Lambek grammars.

Another important point which is still unanswered lies in the decidibility
for $PL(G_1)\subseteq PL(G_2)$ for $G_1, G_2$ Lambek grammars, that is deciding
whether the tree language generated by a grammar is contained in the
tree language generated by another one, for any two grammars. Such a question
is decidable for the non-associative variant of Lambek grammars. Proving this 
question decidable would allow as to very esaily devise a learning algorithm
for k-valued Lambek grammars.\newline

Our learnability result is in our opinion a first step toward a more convincing
and linguistically plausible model of learning for k-valued Lambek grammars from
less and less structurally rich input. Needless to say, learning from such an
informationally rich input like proof-tree structures are hardly has any
linguistic plausibility. On the other hand the deep connections between
proof tree structures for a sentence in Lambek grammars and its ``Montague-like''
semantics seems to address to a more convincing model for learning based
both on syntactic and semantic information.